\documentclass[letterpaper]{article} 
\usepackage{aaai2026}  
\usepackage{times}  
\usepackage{helvet}  
\usepackage{courier}  
\usepackage[hyphens]{url}  
\usepackage{graphicx} 
\usepackage{natbib}  
\usepackage{caption} 
\usepackage{algorithm}
\usepackage{algorithmic}

\usepackage{pifont}
\usepackage{makecell}
\usepackage{subcaption}
\usepackage{float}
\usepackage{footnote}
\usepackage{enumitem}
\usepackage{bm}
\usepackage{arydshln}
\usepackage{booktabs}
\usepackage{multicol}
\usepackage{multirow}
\usepackage{color}
\usepackage{xcolor}     
\usepackage{colortbl}
\usepackage{soul}
\usepackage{bbding}
\usepackage{makecell}
\usepackage{mathtools}
\usepackage{imakeidx}
\usepackage{amssymb}
\usepackage{amsmath}
\usepackage{threeparttable}
\usepackage{hyperref}

\usepackage{newfloat}
\usepackage{listings}
\DeclareCaptionStyle{ruled}{labelfont=normalfont,labelsep=colon,strut=off} 
\floatstyle{ruled}
\newfloat{listing}{tb}{lst}{}
\floatname{listing}{Listing}
\title{MedReasoner: Reinforcement Learning Drives Reasoning Grounding from Clinical Thought to Pixel-Level Precision}
\author{
    Zhonghao Yan\textsuperscript{\rm 1}\equalcontrib,
    Muxi Diao\textsuperscript{\rm 1}\textsuperscript{\rm 2}\equalcontrib,
    Yuxuan Yang\textsuperscript{\rm 1},
    Ruoyan Jing\textsuperscript{\rm 1},
    Jiayuan Xu\textsuperscript{\rm 1},
    Kaizhou Zhang\textsuperscript{\rm 1},\\
    Lele Yang\textsuperscript{\rm 1},
    Yanxi Liu\textsuperscript{\rm 3},
    Kongming Liang\textsuperscript{\rm 1}\thanks{Corresponding Author},
    Zhanyu Ma\textsuperscript{\rm 1}
}
\affiliations{
    \textsuperscript{\rm 1}Beijing University of Posts and Telecommunications\\
    \textsuperscript{\rm 2}Zhongguancun Academy\\
    \textsuperscript{\rm 3}Beijing Information Science and Technology University\\
    \texttt{\{zhonghao.yan, dmx, liangkongming\}@bupt.edu.cn}\\
    \href{https://pris-cv.github.io/MedReasoner/}{\texttt{https://pris-cv.github.io/MedReasoner.github.io/}}
}

\usepackage{bibentry}

\begin{document}

\maketitle

\begin{abstract}

Accurately grounding regions of interest (ROIs) is critical for diagnosis and treatment planning in medical imaging. While multimodal large language models (MLLMs) combine visual perception with natural language, current medical-grounding pipelines still rely on supervised fine-tuning with explicit spatial hints, making them ill-equipped to handle the implicit queries common in clinical practice.
This work makes three core contributions. We first define \textbf{Unified Medical Reasoning Grounding (UMRG)}, a novel vision–language task that demands clinical reasoning and pixel-level grounding. Second, we release \textbf{U-MRG-14K}, a dataset of 14K samples featuring pixel-level masks alongside implicit clinical queries and reasoning traces, spanning 10 modalities, 15 super-categories, and 108 specific categories. Finally, we introduce \textbf{MedReasoner}, a modular framework that distinctly separates reasoning from segmentation: an MLLM reasoner is optimized with reinforcement learning, while a frozen segmentation expert converts spatial prompts into masks, with alignment achieved through format and accuracy rewards.
MedReasoner achieves state-of-the-art performance on U-MRG-14K and demonstrates strong generalization to unseen clinical queries, underscoring the significant promise of reinforcement learning for interpretable medical grounding.

\end{abstract}

\begin{figure*}[t]
    \centering
    \includegraphics[width=1\linewidth]{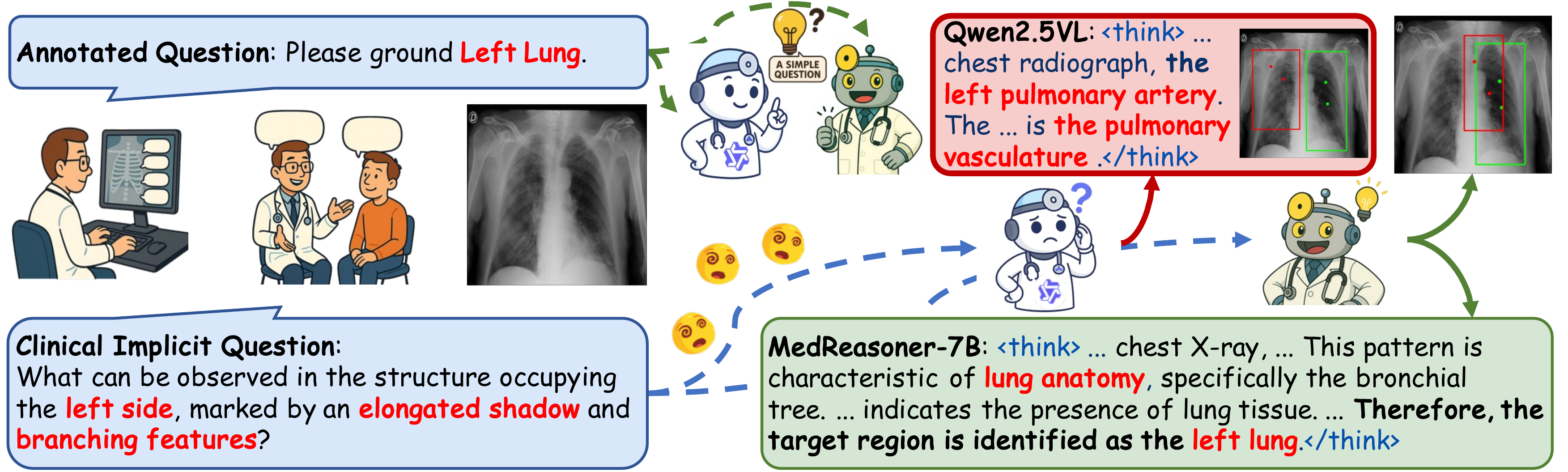}
    \caption{Comparison of annotated question and implicit clinical question. The ground-truth bounding box is green, and models' predicted box is red. \textbf{MedReasoner} precisely identifies the target with the reasoning trace and achieves accurate grounding.}
    \label{fig:teaser}
\end{figure*}

\section{Introduction}
Medical imaging plays a central role in modern healthcare, where clinicians routinely examine regions of interest (ROIs) within these images to assess the health of organs and tissues~\cite{cheng2023sammed2d,lin2024stable,yan2025pgp}. Consequently, precise object detection and image segmentation (often called \textbf{grounding}) are essential for tasks such as disease diagnosis and treatment planning~\cite{chen2021transunet,ma2024segment}.
To further enhance diagnostic efficiency and interpretability, medical Multimodal Large Language Models (MLLMs) have recently emerged~\cite{li2023llava,chen2024huatuogpt,xu2025lingshu}. These models integrate visual perception with language interaction, allowing them to accept free-form language queries, generate high-quality responses, and even identify queried ROIs.

Despite these significant advances, a crucial limitation persists: \textbf{MLLM outputs remain at the image level}. To translate reasoning into visual outputs, every language reference must be grounded to a spatial location. However, while expert models~\cite{cheng2023sammed2d,yue2024surgicalsam} achieve high grounding accuracy, they rely on precise spatial prompts (e.g., \textit{bounding boxes} and \textit{points}). Such detailed annotations are rarely provided by clinicians in real workflows (see Fig.~\ref{fig:teaser} for an example query).
Recent MLLMs attempt to move beyond handcrafted prompts by coupling rich visual components~\cite{da2024segment,huang2025medseg}. However, existing medical grounding pipelines are still trained in a fully supervised manner on explicitly phrased referring expressions (e.g., “\textit{segment the left lung}”)~\cite{liu2023clip,koleilat2024medclip}. Collecting such finely annotated data is costly and, more importantly, misaligned with real clinical queries, which are often \textbf{implicit} (e.g., “\textit{What can be inferred from the irregular shadow?}”). Although some models can name anatomical structures, they often fail to ground them (see Fig.~\ref{fig:teaser}).
Therefore, we need models with reasoning that can turn implicit clinical phrases into explicit spatial targets for grounding in clinical scenarios.

Existing medical visual–question answering (VQA) datasets~\cite{lau2018dataset,he2020pathvqa,liu2021slake} evaluate semantic understanding with image-level question–answer pairs but lack spatial labels. Conversely, large-scale segmentation datasets~\cite{ye2023sa,zhao2024biomedparse,li2024abdomenatlas} provide pixel-accurate masks yet omit language annotations. Neither class of dataset addresses the implicit queries that arise in real clinical practice. \textbf{We have no principled way to measure whether a framework can translate implicit clinical queries into precise spatial grounding}.
Here, we are particularly interested in two research questions that must be addressed before implicit clinical queries can be grounded reliably:
\begin{itemize}
    \item \textit{\textbf{RQ1:} How can we create data that mirrors clinicians’ implicit query patterns while still providing the pixel-level annotations needed for training and evaluation? }
    \item \textit{\textbf{RQ2:} How can we enable models to interpret implicit clinical queries and accurately ground the corresponding image regions without handcrafted spatial prompts?}
\end{itemize}

Guided by the research questions above, we formally introduce the \textbf{Unified Medical Reasoning Grounding (UMRG)} task. UMRG demands a framework that integrates linguistic reasoning with spatial grounding. To succeed, a framework need: (1) interpret the implicit query, (2) reason over visual cues and anatomical priors to infer the latent target, and
(3) generate the accurate pixel-level grounding of that ROIs.
This three-stage process mirrors how clinicians inspect images, reflect, and mark ROIs. Full task specifications are given in Section~\ref{sec:task}.

In response to \textbf{\textit{RQ1}}, we propose \textbf{U-MRG-14K}, a rigorously curated dataset of 14K high-quality samples tailored to the UMRG task. U-MRG-14K is constructed from three open-source datasets. To generate semantically rich and clinically meaningful supervision, we employ GPT-4o~\cite{gpt4o} as a simulator of clinician behavior. And we design a three-stage prompting pipeline that yields high-quality QA pairs, including implicit queries, chain-of-thought (CoT) reasoning traces, and final grounding steps for each target region. Further construction details appear in Section~\ref{sec:dataset}.

In response to \textbf{\textit{RQ2}}, we present \textbf{MedReasoner}, a reinforcement learning (RL) framework for medical reasoning and grounding. MedReasoner is decoupled into two plug-and-play components: \textbf{Clinical Reasoning Module (CRM),} any MLLM that reasons over implicit queries and generates lightweight spatial prompts (a bounding box plus two key points); \textbf{Anatomical Segmentation Module (ASM),} any model that accepts these prompts and returns a pixel-level mask. Because CRM and ASM exchange minimal geometric cues, they can be upgraded without retraining the other.
Most existing grounding pipelines rely on supervised fine-tuning (SFT) with special tokens~\cite{lai2024lisa,tong2025medisee}. This approach suffers from: \textbf{(1) annotation hunger,} it requires large, heavily annotated datasets and CoT traces are especially costly; and \textbf{(2) phrase overfitting,} it encourages models to echo explicit referring phrases and fails to develop genuine reasoning ability.
MedReasoner solve these weaknesses through a rule-based RL training scheme that optimizes only the CRM. In each step, the CRM produces a \textit{think} trace and an \textit{answer} containing spatial prompts, and the frozen ASM renders a mask. Rewards for output format and spatial accuracy drive exploration, gradually aligning reasoning with precise grounding and achieving state-of-the-art performance on U-MRG-14K.
As shown in Fig.~\ref{fig:teaser}, the RL-driven MedReasoner yields sharper grounding and more coherent reasoning than an instruction-tuned baseline, demonstrating its superiority on implicit-query grounding.

To summarize, our contributions are as follows:
\begin{itemize}
    \item We formulate the \textbf{UMRG} task and propose \textbf{U-MRG-14K}. U-MRG-14K pairs implicit clinical queries with pixel-level masks and includes CoT traces to improve the interpretability of grounding.
    \item We present \textbf{MedReasoner}, an RL-driven, plug-and-play framework in which the CRM and the ASM are fully decoupled. This design enables easy substitution and extension to future models and clinical modalities.
    \item We demonstrate through extensive empirical evaluations the effectiveness of our proven MedReasoner framework. We will release the code, and dataset for future research.
\end{itemize}

\section{Related Work}

\subsection{MLLMs for Medical Image Analysis}
Recent advancements in MLLMs have significantly enhanced their capabilities for medical image analysis, with contributions from visual-language alignment techniques~\cite{zhu2025internvl3, wang2024healthg, Qwen2.5-VL, guo2025deepseek} . These progressions have been further extended to various medical applications, including the integration of visual expert modules into pre-trained language models~\cite{li2023llava, sellergren2025medgemma}, and the unification of medical understanding and generation through heterogeneous knowledge adaptation and general foundation models~\cite{chen2024huatuogpt, lin2025healthgpt, xu2025lingshu}. 
However, significant gaps remain in their handling of clinical complexities and crucial clinical grounding tasks, which have seen limited exploration.

\subsection{Visual Grounding with Medical Reasoning}
Recent MLLMs have demonstrated powerful reasoning capabilities~\cite{openai2024o1,guo2025deepseek,liu2025seg,zhu2025internvl3,Qwen2.5-VL}. For visual grounding in general-purpose images, these models often leverage segmentation tools like SAM~\cite{kirillov2023segment}, with methods ranging from training new tokens~\cite{lai2024lisa,ren2024pixellm} to prompting for geometric outputs~\cite{chen2024sam4mllm,uesato2022solving}. However, direct application in medical scenarios is challenging due to opaque reasoning and noisy data. While some specialized works have attempted to address this~\cite{huang2025medseg,trinh2025prs,luo2024vividmed,li2025enhancing}, they often struggle with the natural language found in clinical practice. Inspired by Seg-Zero~\cite{uesato2022solving}, we employ reinforcement learning to generate an explicit CoT~\cite{wei2022chain}. This approach enhances medical visual grounding performance while offering a transparent reasoning process, thereby increasing trust in clinical applications.

\begin{figure*}[t]
    \centering
    \includegraphics[width=1\linewidth]{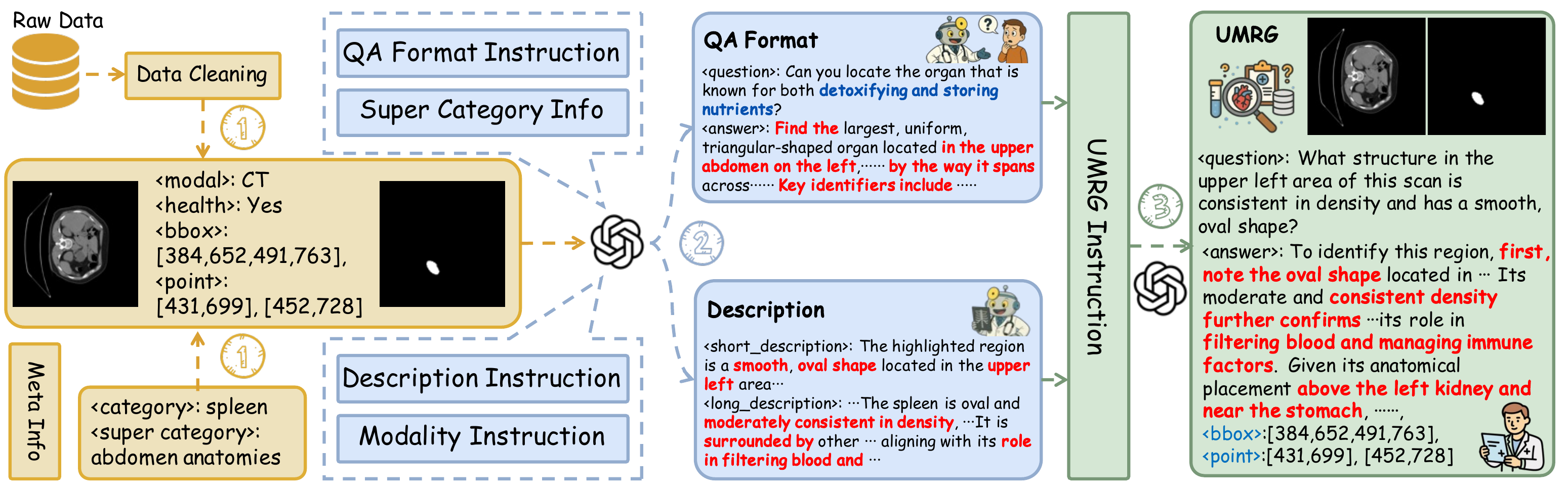}
    \caption{Overview of the \textbf{U-MRG-14K} construction pipeline: (1) Data cleaning and metadata organization manually, (2) Description and QA format generation via GPT-4o, (3) QA pair generation with GPT-4o and human verification.}
    \label{fig:dataset}
\end{figure*}

\section{U-MRG-14K Dataset} \label{sec:dataset}

\subsection{Data Generation} \label{sec:data_generation}
Most existing medical imaging datasets treat visual-grounding and VQA as separate tasks. As a result, some models support natural-language interaction without pixel-level analysis, whereas the accuracy of mainstream segmentation models hinges on the precision of supplied visual prompts. MoCoVQA~\cite{huang2025towards} attempts to unify the two tasks, yet its questions use explicit phrasing that fails to reflect the ambiguity common in routine clinical practice.

To address this gap, we construct \textbf{U-MRG-14K}, a medical grounding dataset centered on implicit referential expressions. U-MRG-14K is generated with GPT-4o~\cite{gpt4o} through carefully designed prompts. As shown in Fig.~\ref{fig:dataset}, its generation process has three stages.

\paragraph{Stage 1: Dataset Preprocessing.} We collect 14K image–mask pairs from SA-Med2D-20M~\cite{ye2023sa}, BiomedParse~\cite{zhao2024biomedparse}, and IMIS-Bench~\cite{cheng2025interactive}. We then standardize and complete the \textit{super-category} labels (coarse anatomical regions) and \textit{category} labels (specific organs or lesions) from the source datasets, producing a consistent and reliable taxonomy. The dataset comprises 15 super-categories and 108 categories. Table~\ref{tab:dataset_compare} provides a systematic comparison showing the advantages of U-MRG-14K over existing datasets.

\paragraph{Stage 2: Descriptions \& QA Formats Generation.} To facilitate the creation of high-quality QA pairs, we perform two preparatory steps. First, for each image, we generate two complementary descriptions: (i) a \textbf{short description} capturing the visual appearance of the region in plain and intuitive language, and (ii) a \textbf{long description} providing a medically precise interpretation of the target area. Second, we use GPT-4o to design a set of QA formats for each super-category. The \textit{questions} mimic realistic clinical queries with vague or implicit references, while the \textit{answers} provide a step-by-step, clinical reasoning path for correct grounding. On average, we create about 20 formats per super-category, with the exact number manually adjusted for class diversity. 

\paragraph{Stage 3: QA Pairs Construction.} Using the per-image descriptions and super-category QA formats from Stage 2, we prompt GPT-4o to synthesize instance-level QA pairs. Each answer contains an explicit, step-by-step reasoning trace guiding the model from an under-specified query to the correct spatial grounding, thereby enhancing interpretability and enabling manual verification. Prompts are iteratively refined, and all generated QA pairs undergo manual screening to remove factual inconsistencies or misaligned reasoning. U-MRG-14K is the first medical-image grounding dataset that includes both pixel-level annotations and complete CoT reasoning traces, providing a valuable resource for reasoning-based grounding and implicit-query QA tasks. 

All generated QA pairs were manually reviewed to eliminate duplicates and factual inconsistencies. Further details and examples of the procedure are provided in Appendix~\ref{apd:dataset}.

\begin{table}[t]
    \centering
    \footnotesize
    \begin{tabular*}{\linewidth}{@{\extracolsep{\fill}}lccccc}
    \toprule
    Dataset & \# Prompts & QAs & Sup. & Cat. & CoT \\
    \midrule
    SA-Med2D & 20M & \ding{55} & - & 219 & \ding{55} \\
    BioMedParse & 1.1M & \ding{55} & 3 & 82 & \ding{55} \\
    IMED & 361M & \ding{55} & 6 & 204 & \ding{55} \\
    MoCoVQA & 100K & \ding{51} & - & - & \ding{55} \\
    \midrule
    \textbf{U-MRG-14K} & 14K & \ding{51} & 15 & 108 & \ding{51} \\
    \bottomrule
    \end{tabular*}
    \caption{
    Comparison of U-MRG-14K with existing medical vision–language datasets. \textbf{Sup.} and \textbf{Cat.} denote the numbers of \textit{super-categories} and fine-grained \textit{categories}, respectively. U-MRG-14K supplies customized QA templates for each category, and is the only dataset that includes CoT annotations for reasoning-aware evaluation.
    }
    \label{tab:dataset_compare}
\end{table}

\subsection{Dataset statistics}
U-MRG-14K contains 14K image-mask pairs from ten imaging modalities (e.g., \textit{CT}, \textit{MRI}). The dataset is organized into 15 super-categories (e.g., \textit{abdomen}, \textit{lung}) covering frequent anatomical regions and pathology-oriented classes, providing broad clinical coverage. Within these, 108 fine-grained categories denote specific structures, reflecting hierarchical structure of anatomy. For instance, \textit{left lung} and \textit{right lung} are separate categories nested under the \textit{lung} super-category. Beyond pixel-level masks, every sample includes a CoT reasoning trace. These annotations make the reasoning process transparent and verifiable, allowing researchers to inspect the model's decision path.

We implemented a stringent three-phase manual quality control process for U-MRG-14K. Initially, 5 graduate-level students with medical imaging knowledge performed data deduplication and fact-checking, with each entry reviewed by 3 individuals. Subsequently, we conducted a detailed review, focusing on eliminating clinically illogical CoT data. Finally, 2 experienced radiologists with over 5 years of practice performed a random audit of 10\% of the data to quantify overall quality. Comprehensive data statistics are in Appendix~\ref{apd:statistics}.

\section{MedReasoner}

\subsection{Task Definition} \label{sec:task}

Given a medical image $\mathcal{I}$ and a clinical query $\mathcal{Q}$ with implicit referring expressions, the model $\mathbf{G}$ outputs a bounding box $\mathcal{B}$, two semantic key points $\mathcal{P}_1$ and $\mathcal{P}_2$, and a pixel-level segmentation mask $\mathcal{M}$. The process can be formulated as:
\begin{equation}
\{\mathcal{T},\mathcal{B},\mathcal{P}_1,\mathcal{P}_2,\mathcal{M}\}=\mathbf{G}(\mathcal{I},\mathcal{Q}).
\label{eq:umrg_task}
\end{equation}
where $\mathcal{T}$ is an optional CoT trace that records the model’s intermediate reasoning, analogous to how a clinicians infers the target from implicit linguistic cues.

\begin{figure*}[t]
    \centering
    \includegraphics[width=1\linewidth]{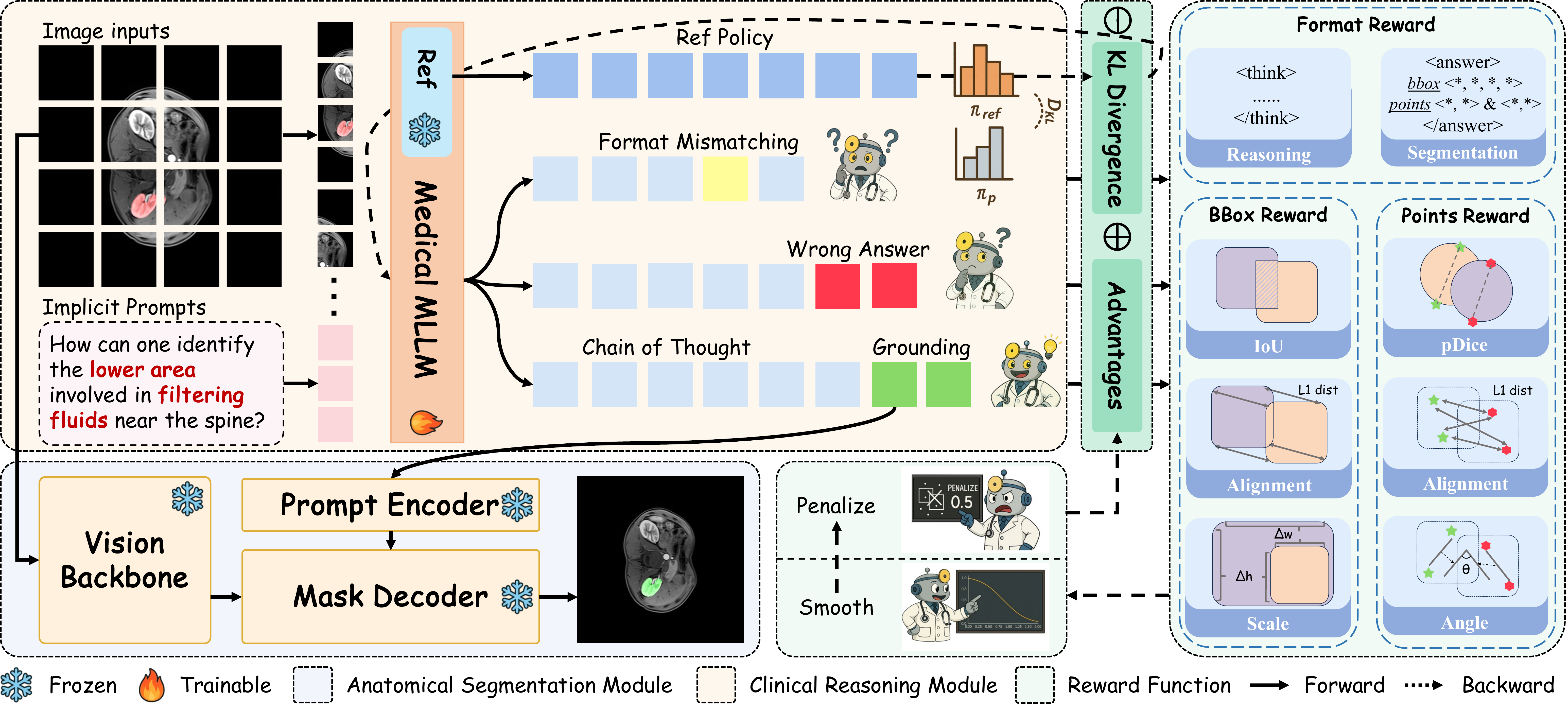}
    \caption{Overview of the \textbf{MedReasoner} framework. MedReasoner transforms implicit clinical prompts into pixel-level grounding via a two-stage process. The \textbf{CRM} first generates intermediate reasoning and grounding outputs (\textit{CoT}, \textit{bounding box}, and \textit{key points}). Then, the \textbf{ASM} converts the grounded outputs into final segmentation masks.}
    \label{fig:model}
\end{figure*}

\subsection{Model Architecture}

Enabling native pixel-level segmentation in an MLLM usually requires custom \texttt{[MASK]} tokens, multi-head decoders, and large collections of manual mask annotations~\cite{pi2024perceptiongpt,lai2024lisa,tong2025medisee}. However, MedSAM2~\cite{ma2025medsam2} already yields modality-agnostic masks out of the box.
As shown in Fig.~\ref{fig:model}, \textbf{MedReasoner} decouples language reasoning from pixel-level grounding, thereby (1) mitigating phrase over-fitting to enable authentic reasoning, and (2) treating MedSAM-family models as plug-and-play components controllable by language.

\paragraph{Clinical Reasoning Module.} 
We employ Lingshu~\cite{xu2025lingshu} as our \textbf{Clinical Reasoning Module (CRM)} $\mathbf{F}_{reason}$. Given $(\mathcal{I},\mathcal{Q})$, CRM outputs a structured tuple $\texttt{<think>}\cdots\texttt{<think>} \texttt{<answer>}\mathcal{B},\mathcal{P}_{1},\mathcal{P}_{2}\texttt{<answer>}$. A bounding box $\mathcal{B}$ is often inadequate in medical images: boxes may enclose multiple organs or lesions, and their corners lack semantics for SAM-style prompters. We therefore add two key points $\mathcal{P}_1,\mathcal{P}_2$ on visually distinctive regions. These enrich spatial cues at low annotation cost.
To learn reliable cues without compromising linguistic competence, we train $\mathbf{F}_{\text{reason}}$ with \textbf{Group Relative Policy Optimization (GRPO)}~\cite{shao2024deepseekmath}, using: (1) \textbf{format rewards} enforcing the output schema, and (2) \textbf{accuracy rewards} measuring spatial correctness.

\paragraph{Anatomical Segmentation Module.} 
We instantiate the Anatomical Segmentation Module (ASM) with a frozen MedSAM2~\cite{ma2025medsam2}, denoted as $\mathbf{F}_{\text{seg}}$. The tuple $(\mathcal{B},\mathcal{P}_1,\mathcal{P}_2)$ produced by the CRM is fed to $\mathbf{F}_{\text{seg}}$, which transforms these coarse prompts into a high-resolution mask $\mathcal{M}$ without any task-specific fine-tuning. 
Freezing $\mathbf{F}_{\text{seg}}$ preserves MedSAM2’s strong zero-shot delineation ability, while allowing $\mathbf{F}_{\text{reason}}$ to concentrate on language understanding and spatial reasoning.  

\subsection{Reward Functions} \label{sec:reward_functions}
Reward functions in RL guide a model toward the behaviors we desire. For UMRG, we introduce three rewards that first prompt the model to reason about the implicit target and then to predict the bounding box and key points.

\paragraph{Reasoning Formats Reward.}
This reward evaluates the structural validity of the model's output, focusing on the formatting of the reasoning and answer components. It assigns $\mathbb{R}_{\text{think}}$ to assess whether the model produces a well-structured \texttt{<think>} block, and $\mathbb{R}_{\text{answer}}$ to verify whether the \texttt{<answer>} block is a valid JSON object containing the required fields: \texttt{bbox}, \texttt{points\_1}, and \texttt{points\_2}. These rewards do not evaluate the correctness or quality of the reasoning content itself, but rather the presence and structural completeness of the expected formats. Both $\mathbb{R}_{\text{think}}$ and $\mathbb{R}_{\text{answer}}$ are assigned discrete values in the range $[0, 1]$.

\paragraph{Grounding Box Reward.}
This reward evaluates the quality of the predicted bounding box $\mathcal{B}_p = [x_1^p, y_1^p, x_2^p, y_2^p]$ against the ground-truth box $\mathcal{B}_g = [x_1^g, y_1^g, x_2^g, y_2^g]$, where all coordinates are normalized to $[0, 1]$. First, the \textbf{IoU reward} measures the spatial overlap between two boxes:
\begin{equation}
    \mathbb{R}_{\mathrm{iou}} = \frac{\mathrm{Area}(\mathcal{B}_p \cap \mathcal{B}_g)}{\mathrm{Area}(\mathcal{B}_p \cup \mathcal{B}_g)}.
\end{equation}
Second, the \textbf{Alignment reward} computes the average L1 distance between corresponding corner coordinates, normalized by the diagonal of $\mathcal{B}_g$:
\begin{equation}
    \mathbb{R}_{\mathrm{align}} = \frac{1}{4} \sum_{i=1}^{4} \left| \mathcal{B}_p^{(i)} - \mathcal{B}_g^{(i)} \right|.
\end{equation}
Third, the \textbf{Scale reward} captures shape consistency in terms of area and aspect ratio. Specifically, we compute the logarithmic difference in box area and aspect ratio, and define:
\begin{equation}
    \mathbb{R}_{\mathrm{scale}} = \sqrt{(\Delta \log A)^2 + (\Delta \log R)^2},
\end{equation}
where $A$ denotes box area and $R$ the aspect ratio. Smaller values indicate better structural alignment.

\paragraph{Grounding Points Reward.}
This reward evaluates the quality of the predicted key point pair $\mathcal{P}_p = \{\mathbf{p}_1^p, \mathbf{p}_2^p\}$ by comparing it with the ground-truth pair $\mathcal{P}_g = \{\mathbf{p}_1^g, \mathbf{p}_2^g\}$, where each point $\mathbf{p} = (x, y)$ is normalized to $[0, 1]$.
First, the \textbf{pDice reward} estimates the spatial overlap between circles formed by each point pair. Each uses the two points as diameter endpoints, and the Dice score is computed as:
\begin{equation}
    \mathbb{R}_{\mathrm{pdice}} = \frac{2 \cdot \mathrm{Area}(O_p \cap O_g)}{\mathrm{Area}(O_p) + \mathrm{Area}(O_g)},
\end{equation}
where $O_p$ and $O_g$ are the circles constructed from $\mathcal{P}_p$ and $\mathcal{P}_g$, respectively.
Second, the \textbf{Alignment reward} computes the mean absolute error between corresponding points:
\begin{equation}
    \mathbb{R}_{\mathrm{align}} = \frac{1}{2} \sum_{i=1}^{2} \left( |x_i^p - x_i^g| + |y_i^p - y_i^g| \right).
\end{equation}
Third, the \textbf{Angle reward} measures the cosine similarity between the predicted and ground-truth direction vectors, capturing angular consistency:
\begin{equation}
    \mathbb{R}_{\mathrm{angle}} = \left| \cos\left( \theta \right) \right| = \left| \frac{\langle \mathbf{v}_p, \mathbf{v}_g \rangle}{\| \mathbf{v}_p \|_2 \cdot \| \mathbf{v}_g \|_2} \right|,
\end{equation}
where $\mathbf{v}_p = \mathbf{p}_2^p - \mathbf{p}_1^p$ and $\mathbf{v}_g = \mathbf{p}_2^g - \mathbf{p}_1^g$. Further details of reward functions are provided in Appendix~\ref{apd:reward_function}.

\paragraph{Smoothing and Penalization.}
To enhance training stability and differentiate prediction quality, we apply smoothing functions to all reward components. For the $\mathbb{R}_{\mathrm{iou}}$, $\mathbb{R}_{\mathrm{pdice}}$ and $\mathbb{R}_{\mathrm{angle}}$ rewards, we use logarithmic smoothing:
\begin{equation}
    \mathcal{S}_{\log}(r; k) = \frac{\log(k r + 1)}{\log(k + 1)},
\end{equation}
where $r \in [0, 1]$ is the raw reward and $k$ is a scaling factor (default $k = 3$).  
For the $\mathbb{R}_{\mathrm{align}}$ and $\mathbb{R}_{\mathrm{scale}}$ rewards, we use exponential smoothing:
\begin{equation}
    \mathcal{S}_{\exp}(d; k, c) = \frac{1}{1 + e^{k(d - c)}},
\end{equation}
where $d \in [0, 2]$ is the normalized distance, and $c$ is the target center (default $c = 1$).

After smoothing, we apply a penalization function $\mathcal{N}(\cdot)$ to softly down-weight unreliable predictions. For each reward, two validity scores are computed to reflect the spatial plausibility of the output. The final reward is adjusted as:
\begin{equation}
    \mathcal{N}(r; v_1, v_2) = \lambda r + (1 - \lambda) r \cdot \frac{v_1 + v_2}{2},
\end{equation}
where $r$ is the smoothed reward, $v\_1$ and $v\_2$ are the two validity scores, and $\lambda = 0.7$ by default. More details of smoothing and penalization are provided in Appendix~\ref{apd:smoothing_penalization}.

\section{Experiments}

\begin{table*}[t]
    \centering
    \footnotesize
    \begin{tabular*}{\textwidth}{@{\extracolsep{\fill}}lrrrrrrrrrrrrr}
    \toprule
    \multirow{2}{*}{Method} & \multirow{2}{*}{IoU$\uparrow$} & \multirow{2}{*}{pDice$\uparrow$} & \multirow{2}{*}{Dice$\uparrow$} & \multicolumn{10}{c}{Super-Categories (IoU$\uparrow$)} \\
    \cmidrule(lr){5-14}
    & & & & Abd. & Brain & Eye & Heart & Hist. & Lung & Ves. & Neo. & N-Neo. & Inf. \\
    \midrule
    \multicolumn{14}{c}{\textbf{General MLLMs}} \\
    \midrule
    GPT-4o & 2.65 & 1.12 & 4.72 & 0.92 & 0.91 & 3.29 & 0.36 & 2.8 & 11.70 & 1.83 & 1.01 & 4.16 & 6.37 \\
    Gemini-2.5-flash & 7.86 & 3.24 & 14.29 & 3.99 & 5.69 & 6.39 & 7.77 & 6.63 & 16.37 & 9.08 & 7.15 & 13.91 & 11.4 \\
    Qwen2.5VL-7B & 12.61 & 7.14 & 22.73 & 6.84 & \underline{23.97} & 29.35 & 8.37 & 9.22 & 20.79 & 20.46 & 8.00 & 24.97 & 19.4 \\
    InternVL3-8B & 5.70 & 2.46 & 9.23 & 3.72 & 6.54 & 2.02 & 3.67 & 5.56 & 14.44 & 7.88 & 3.78 & 8.71 & 9.00 \\
    Qwen2.5-VL-72B & \underline{18.32} & \underline{12.39} & \underline{29.71} & \underline{13.60} & 20.06 & 38.3 & \underline{15.51} & 8.74 & \underline{35.25} & 20.64 & \underline{20.69} & \underline{30.19} & 16.92 \\
    InternVL3-78B & 4.02 & 1.55 & 7.23 & 2.04 & 2.95 & 2.33 & 2.12 & 6.12 & 12.21 & 4.19 & 1.33 & 8.19 & 5.62 \\
    \midrule
    \multicolumn{14}{c}{\textbf{Medical-Specific MLLMs}} \\
    \midrule
    MedR1-2B & 8.18 & 3.60 & 14.73 & 3.53 & 12.55 & 1.10 & 3.53 & 8.14 & 25.58 & 8.81 & 4.39 & 13.57 & 17.35 \\
    MiniInternVL-4B & 2.88 & 0.85 & 4.76 & 1.88 & 2.67 & 0.68 & 1.60 & 3.45 & 7.99 & 3.59 & 1.56 & 3.76 & 6.59 \\
    MedGamma-4B & 5.39 & 1.90 & 8.90 & 4.23 & 6.92 & 1.28 & 3.41 & 4.78 & 17.22 & 6.92 & 3.17 & 3.90 & 10.04 \\
    HuatuoGPT-7B & 10.13 & 5.23 & 19.76 & 5.88 & 18.16 & 3.88 & 6.63 & 9.56 & 22.94 & 15.58 & 8.25 & 16.12 & 15.87 \\
    Lingshu-7B & 8.19 & 3.73 & 16.48 & 4.03 & 15.72 & 6.97 & 6.27 & 8.06 & 19.77 & 8.63 & 6.34 & 13.31 & 11.99 \\
    Chiron-o1-8B & 6.40 & 2.46 & 10.05 & 3.82 & 6.90 & 4.29 & 4.20 & 5.99 & 12.86 & 9.50 & 5.53 & 11.31 & 10.86 \\
    \midrule
    \multicolumn{14}{c}{\textbf{Grounding-Specific MLLMs}} \\
    \midrule
    VLMR1-REC-3B & 13.96 & - & 22.19 & 8.64 & 21.81 & 25.09 & 8.19 & \underline{10.69} & 29.77 & \underline{21.35} & 8.76 & 26.59 & 21.41 \\
    SegZero-7B & 16.14 & 5.23 & 26.05 & 11.66 & 23.37 & \underline{40.23} & 13.12 & 9.35 & 22.18 & 20.68 & 12.58 & 29.46 & \underline{21.93} \\
    SAM4MLLM-8B & 7.94 & - & 16.49 & 6.30 & 14.69 & 5.09 & 5.81 & 7.46 & 12.61 & 11.99 & 6.24 & 11.96 & 12.40 \\
    \textbf{MedReasoner-7B} & \textbf{32.42} & \textbf{26.55} & \textbf{37.78} & \textbf{30.27} & \textbf{32.81} & \textbf{51.50} & \textbf{34.72} & \textbf{11.66} & \textbf{50.75} & \textbf{29.91} & \textbf{33.58} & \textbf{37.19} & \textbf{30.48} \\
    \bottomrule
    \end{tabular*}
    \caption{
    Results on the \textbf{U-MRG-14K} test set under the \textbf{MedReasoner} paradigm. Each candidate uses one medical MLLM as the \textbf{CRM} to output a bounding box and two key points; the \textbf{ASM} is fixed to \textit{MedSAM2}. \textbf{Bold} numbers denote the best score in each column, and \underline{underlined} numbers denote the second best.
}
\label{tab:main_results}
\end{table*}

\subsection{Experimental Settings}
\paragraph{Models.}
We conduct a comprehensive comparison across a wide range of models.
For general MLLMs, we utilized GPT-4o~\cite{gpt4o}, Gemini-2.5-flash~\cite{gemini2.5_flash}, Qwen2.5VL-7B/72B~\cite{Qwen2.5-VL} and InternVL3-8B/78B~\cite{zhu2025internvl3}.
For medical-specific MLLMs, we selected MedR1-2B~\cite{lai2025med}, MiniInternVL-4B~\cite{gao2024mini}, MedGamma-4B~\cite{sellergren2025medgemma}, HuatuoGPT-7B-Qwen2.5VL~\cite{chen2024huatuogpt}, Lingshu-7B~\cite{xu2025lingshu}, and Chiron-o1-8B~\cite{sun2025enhancing}.
For segmentation models, we chose MedSAM~\cite{ma2024segment}, SAM-Med2D~\cite{cheng2023sammed2d} and MedSAM2 \cite{ma2025medsam2}.
For grounding-specific models, we included SAM4MLLM~\cite{chen2024sam4mllm}, VLMR1-REC-3B~\cite{shen2025vlm} and SegZero-7B~\cite{liu2025seg}.

\paragraph{Datasets.}
We train MedReasoner on U-MRG-14K, using the data preparation strategy mentioned in Section \ref{sec:data_generation}. We randomly hold out 2.5K samples as a test set, and use the remaining data for training. All quantitative results reported in this paper are obtained on the test set.

\paragraph{Implementation Details.}
We adopt Lingshu-7B with the Soft reward function as our default CRM and default ASM to MedSAM2 (see details in Appendix~\ref{apd:implementation_details}).

\paragraph{Evaluation Metrics.}
We compute three evaluation metrics: \textbf{IoU}, \textbf{pDice}, and \textbf{Dice} to assess model performance. 
\textbf{IoU} measures the bounding box localization accuracy predicted by MLLMs.
\textbf{pDice} quantifies keypoint pair semantic alignment by evaluating the overlap of circles formed by predicted endpoints (formally defined in Section~\ref{sec:reward_functions}). 
\textbf{Dice} assesses segmentation quality based on masks generated by downstream models conditioned on MLLM outputs.

\subsection{Medical Reasoning Grounding Results}
For fair comparison, we evaluated models under the MedReasoner paradigm, using a single MLLM as CRM to return bounding box and key point, with MedSAM2 fixed as ASM. All MLLMs are driven by the same user prompt (full prompt in Appendix~\ref{apd:user_prompts}).
As shown in Table~\ref{tab:main_results}, MedReasoner-7B achieved superior overall performance, significantly leading the second-best Qwen2.5VL-72B by 14.10 in IoU, 14.16 in pDice, and 8.07 in Dice. This highlights its precise spatial prompting capability. While General MLLMs, such as GPT-4o (IoU 2.65), and Medical-Specific MLLMs, like HuatuoGPT-7B (IoU 10.13), demonstrated cross-modal understanding or domain benefits, they consistently lacked the fine-grained precision required for UMRG. In contrast, MedReasoner-7B established a substantial lead among Grounding-Specific MLLMs, surpassing SegZero-7B (IoU 16.14) by over 16 IoU points, validating our RL-driven grounding strategy for accurate regional prompt translation. This superiority extended across super-categories, with MedReasoner-7B leading in most (e.g., Lung's IoU was 50.75, Eye's IoU was 51.50), though all models, including ours, faced challenges in complex categories like Histology (MedReasoner-7B's IoU was 11.66).

\begin{table}[t]
    \centering
    \begin{tabular*}{\linewidth}{@{\extracolsep{\fill}}lrrrr}
    \toprule
    Method & IoU$\uparrow$ & pDice$\uparrow$ & Dice$\uparrow$ & \# Ref.$\downarrow$ \\
    \midrule
    Lingshu & 8.19 & 3.73 & 16.51 & 2 \\
    Lingshu w/ SFT & 9.15 & 2.88 & 15.22 & 2 \\
    \midrule
    Lingshu w/ RL(Base) & 15.85 & 8.29 & 28.79 & 0 \\
    Lingshu w/ RL(Hard) & \underline{31.69} & \underline{24.36} & \underline{33.51} & 0 \\
    Lingshu w/ RL(Soft) & \textbf{32.42} & \textbf{26.55} & \textbf{37.78} & 0 \\
    \bottomrule
    \end{tabular*}
    \caption{
    Comparison of the \textbf{SFT} baseline with three RL variants: \textbf{Base}, \textbf{Hard}, and \textbf{Soft} on U-MRG-14K. \textbf{\# Ref.} denotes refusals to ground answers with the reasoning prompt.
    }
    \label{tab:ablation_reward}
\end{table}

\begin{table}[t]
    \centering
    \begin{tabular*}{\linewidth}{@{\extracolsep{\fill}}l|rrr}
    \toprule
    \multirow{2}{*}{Method} & \multicolumn{3}{c}{Dice$\uparrow$} \\
    \cmidrule(lr){2-4}
    & w/ Points & w/ BBox & BBox \& Points \\
    \midrule
    MedSAM & 5.67 & 28.39 & 19.00 \\
    SAM-Med2D & \underline{33.23} & \underline{35.03} & \underline{36.48} \\
    MedSAM2 & \textbf{34.86} & \textbf{37.15} & \textbf{37.78} \\
    \bottomrule
    \end{tabular*}
    \caption{Dice scores for three segmentation backbones under three prompt types: \textbf{key points only}, \textbf{bounding box only}, and the \textbf{combined bounding box \& points}.
    }
    \label{tab:ablation_seg}
\end{table}

\begin{figure*}[t]
    \centering
    \includegraphics[width=1\linewidth]{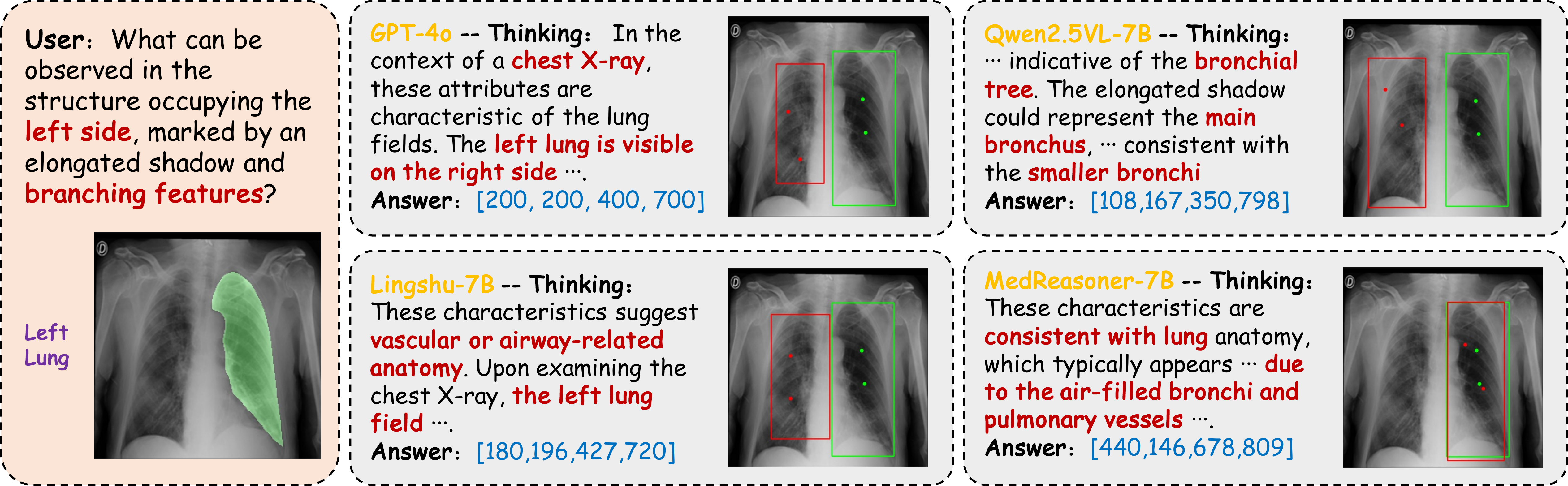}
    \caption{Open-source (\textbf{Qwen2.5VL-7B}), closed-source (\textbf{GPT-4o}), and medically post-trained (\textbf{Lingshu-7B}) models reason and specify referents within CoT processes, a characteristic tied to their respective training methodologies. Our \textbf{MedReasoner-7B} integrates grounding information during training, thereby aligning pixel-level grounding with semantic-level understanding.}
    \label{fig:qualitative}
\end{figure*}

\subsection{Ablation Study}

We conduct ablation studies to verify our proposed design's effectiveness. All experiments are trained on U-MRG-14K and use the same user prompt during inference. 

\paragraph{Effect of Reward Types.}
This ablation study investigated how reward design influences RL training for CRM. Our SFT baseline had a low IoU of 9.15 and 2 query refusals (as Table~\ref{tab:ablation_reward} shows). RL fine-tuning drastically improved performance, eliminating all refusals. We evaluated three reward variants: \textbf{Base} (a hard-threshold scheme~\cite{liu2025seg}), \textbf{Hard} (our full reward), and \textbf{Soft} (IoU and pDice only). While Base removed refusals, its IoU of 15.85 was considerably lower. Our Hard reward significantly outperformed Base, increasing IoU by 15.84 points. The Soft reward variant achieved the best overall IoU of 32.42, surpassing Hard by 0.73 points, suggesting that less strict alignment fosters better exploration and more accurate grounding solutions.

\paragraph{Effect of Segmentation Backbones.}
This ablation assesses the ASM. Table~\ref{tab:ablation_seg} reports results for three medical SAM variants: MedSAM, SAM-Med2D, and MedSAM2. To investigate prompt influence, we evaluated three input formats per backbone: points only, bounding box only, and the combined bounding box and points. The combination consistently yielded the best Dice (37.78), with MedSAM2 achieving the highest performance across all configurations.

\paragraph{Effect of Reasoning Strategies.}
This ablation tests whether prompting the model to reason before grounding helps when answering implicit queries. We designed two user prompts (see Appendix~\ref{apd:user_prompts}): \textbf{Direct} asks the CRM to output the spatial prompt immediately, whereas \textbf{Reasoning} instructs it to first generate a brief CoT.
As Table~\ref{tab:ablation_reason} shows, the Reasoning prompt significantly reduces refusal rates compared to the Direct prompt. While base Qwen2.5VL and Lingshu exhibit a slight performance drop due to their limited inherent reasoning capabilities, this behavior is expected. However, after CRM training within the MedReasoner framework, the Reasoning strategy clearly outperforms the Direct one. This confirms that an explicit reasoning phase is valuable for implicit-query grounding.

\begin{table}[t]
    \centering
    \begin{tabular*}{\linewidth}{@{\extracolsep{\fill}}lrrrr}
    \toprule
    Method & Reason & IoU$\uparrow$ & pDice$\uparrow$ & \# Ref.$\downarrow$ \\
    \midrule
    Qwen2.5VL-7B & \ding{55} & 14.57 & 8.14 & 13 \\
    Qwen2.5VL-7B & \ding{51} & 12.61 & 7.14 & 0 \\
    Lingshu-7B & \ding{55} & 9.35 & 2.40 & 4 \\
    Lingshu-7B & \ding{51} & 8.19 & 3.73 & 2 \\
    MedReasoner-7B & \ding{55} & \underline{30.29} & \underline{25.82} & 12 \\
    MedReasoner-7B & \ding{51} & \textbf{32.42} & \textbf{26.55} & 0 \\
    \bottomrule
    \end{tabular*}
    \caption{
    Impact of adding an explicit reasoning step vs. a direct prompt for three CRMs. \textbf{Reason} indicates whether the model is prompted to reason first (\ding{51}) or respond directly (\ding{55}).
    }
    \label{tab:ablation_reason}
\end{table}

\subsection{Qualitative Results}
Figure~\ref{fig:qualitative} compares four MLLMs' predictions on a chest X-ray query requiring implicit reasoning.
\textbf{GPT-4o} produces a coherent CoT and an accurate image-level answer, but its spatial output is wrong: the bounding box is misplaced and coordinates are rounded, indicating limited fine-grained grounding. \textbf{Qwen2.5VL-7B} fails at the reasoning stage, resulting in an incorrect diagnosis and an irrelevant box. \textbf{Lingshu-7B} correctly identifies the \textit{left lung} but misplaces the box, demonstrating that it alone doesn't guarantee accurate localization. Only \textbf{MedReasoner-7B} precisely identifies and pinpoints the target; its box tightly encloses the bronchial tree of the left lung, with key points aligning to it.
These observations highlight the necessity of explicit RL-based grounding. It preserves the reasoning quality of large models while enforcing the spatial precision crucial for UMRG. Additional qualitative results are in Appendix~\ref{apd:qualitative_results}.

\section{Conclusion}

We present the \textbf{UMRG} task, which challenges models to transform implicit clinical queries into precise pixel-level grounding. To support this, we introduce \textbf{U-MRG-14K}, a large-scale dataset featuring rich annotations and reasoning traces. To solve UMRG, we propose \textbf{MedReasoner}, a modular framework that decouples reasoning from segmentation and leverages RL to align linguistic reasoning with spatial precision. Extensive experiments demonstrate that MedReasoner consistently outperforms existing models in accuracy. We believe this framework offers a promising step toward trustworthy and generalizable medical grounding systems.

\bibliography{aaai2026}

@inproceedings{sheng2025hybridflow,
  title={Hybridflow: A flexible and efficient rlhf framework},
  author={Sheng, Guangming and Zhang, Chi and Ye, Zilingfeng and Wu, Xibin and Zhang, Wang and Zhang, Ru and Peng, Yanghua and Lin, Haibin and Wu, Chuan},
  booktitle={Proceedings of the Twentieth European Conference on Computer Systems},
  pages={1279--1297},
  year={2025}
}

@inproceedings{huang2025towards,
  title={Towards a multimodal large language model with pixel-level insight for biomedicine},
  author={Huang, Xiaoshuang and Shen, Lingdong and Liu, Jia and Shang, Fangxin and Li, Hongxiang and Huang, Haifeng and Yang, Yehui},
  booktitle={Proceedings of the AAAI Conference on Artificial Intelligence},
  volume={39},
  pages={3779--3787},
  year={2025}
}

@inproceedings{yan2025pgp,
  title={PGP-SAM: Prototype-Guided Prompt Learning for Efficient Few-Shot Medical Image Segmentation},
  author={Yan, Zhonghao and Yin, Zijin and Lin, Tianyu and Zeng, Xiangzhu and Liang, Kongming and Ma, Zhanyu},
  booktitle={2025 IEEE 22nd International Symposium on Biomedical Imaging (ISBI)},
  pages={1--5},
  year={2025},
  organization={IEEE}
}

@inproceedings{chen2024internvl,
  title={Internvl: Scaling up vision foundation models and aligning for generic visual-linguistic tasks},
  author={Chen, Zhe and Wu, Jiannan and Wang, Wenhai and Su, Weijie and Chen, Guo and Xing, Sen and Zhong, Muyan and Zhang, Qinglong and Zhu, Xizhou and Lu, Lewei and others},
  booktitle={Proceedings of the IEEE/CVF Conference on Computer Vision and Pattern Recognition},
  pages={24185--24198},
  year={2024}
}

@inproceedings{chen2024sam4mllm,
  title={Sam4mllm: Enhance multi-modal large language model for referring expression segmentation},
  author={Chen, Yi-Chia and Li, Wei-Hua and Sun, Cheng and Wang, Yu-Chiang Frank and Chen, Chu-Song},
  booktitle={European Conference on Computer Vision},
  pages={323--340},
  year={2024},
  organization={Springer}
}

@inproceedings{ren2024pixellm,
  title={Pixellm: Pixel reasoning with large multimodal model},
  author={Ren, Zhongwei and Huang, Zhicheng and Wei, Yunchao and Zhao, Yao and Fu, Dongmei and Feng, Jiashi and Jin, Xiaojie},
  booktitle={Proceedings of the IEEE/CVF Conference on Computer Vision and Pattern Recognition},
  pages={26374--26383},
  year={2024}
}

@inproceedings{yue2024surgicalsam,
  title={SurgicalSAM: Efficient class promptable surgical instrument segmentation},
  author={Yue, Wenxi and Zhang, Jing and Hu, Kun and Xia, Yong and Luo, Jiebo and Wang, Zhiyong},
  booktitle={Proceedings of the AAAI Conference on Artificial Intelligence},
  volume={38},
  pages={6890--6898},
  year={2024}
}

@inproceedings{pi2024perceptiongpt,
  title={Perceptiongpt: Effectively fusing visual perception into llm},
  author={Pi, Renjie and Yao, Lewei and Gao, Jiahui and Zhang, Jipeng and Zhang, Tong},
  booktitle={Proceedings of the IEEE/CVF conference on computer vision and pattern recognition},
  pages={27124--27133},
  year={2024}
}

@inproceedings{koleilat2024medclip,
  title={Medclip-sam: Bridging text and image towards universal medical image segmentation},
  author={Koleilat, Taha and Asgariandehkordi, Hojat and Rivaz, Hassan and Xiao, Yiming},
  booktitle={International conference on medical image computing and computer-assisted intervention},
  pages={643--653},
  year={2024},
  organization={Springer}
}

@inproceedings{lai2024lisa,
  title={Lisa: Reasoning segmentation via large language model},
  author={Lai, Xin and Tian, Zhuotao and Chen, Yukang and Li, Yanwei and Yuan, Yuhui and Liu, Shu and Jia, Jiaya},
  booktitle={Proceedings of the IEEE/CVF Conference on Computer Vision and Pattern Recognition},
  pages={9579--9589},
  year={2024}
}

@article{li2023llava,
  title={Llava-med: Training a large language-and-vision assistant for biomedicine in one day},
  author={Li, Chunyuan and Wong, Cliff and Zhang, Sheng and Usuyama, Naoto and Liu, Haotian and Yang, Jianwei and Naumann, Tristan and Poon, Hoifung and Gao, Jianfeng},
  journal={Advances in Neural Information Processing Systems},
  volume={36},
  pages={28541--28564},
  year={2023}
}

@inproceedings{kirillov2023segment,
  title={Segment anything},
  author={Kirillov, Alexander and Mintun, Eric and Ravi, Nikhila and Mao, Hanzi and Rolland, Chloe and Gustafson, Laura and Xiao, Tete and Whitehead, Spencer and Berg, Alexander C and Lo, Wan-Yen and others},
  booktitle={Proceedings of the IEEE/CVF international conference on computer vision},
  pages={4015--4026},
  year={2023}
}

@inproceedings{liu2023clip,
  title={Clip-driven universal model for organ segmentation and tumor detection},
  author={Liu, Jie and Zhang, Yixiao and Chen, Jie-Neng and Xiao, Junfei and Lu, Yongyi and A Landman, Bennett and Yuan, Yixuan and Yuille, Alan and Tang, Yucheng and Zhou, Zongwei},
  booktitle={Proceedings of the IEEE/CVF international conference on computer vision},
  pages={21152--21164},
  year={2023}
}

@misc{cheng2023sammed2d,
    title={SAM-Med2D},
    author={Junlong Cheng and Jin Ye and Zhongying Deng and Jianpin Chen and Tianbin Li and Haoyu Wang and Yanzhou Su and Ziyan Huang and Jilong Chen and Lei Jiang and Hui Sun and Junjun He and Shaoting Zhang and Min Zhu and Yu Qiao},
    year={2023},
    eprint={2308.16184},
    archivePrefix={arXiv},
    primaryClass={cs.CV}
}

@inproceedings{cheng2025interactive,
  title={Interactive medical image segmentation: A benchmark dataset and baseline},
  author={Cheng, Junlong and Fu, Bin and Ye, Jin and Wang, Guoan and Li, Tianbin and Wang, Haoyu and Li, Ruoyu and Yao, He and Cheng, Junren and Li, JingWen and others},
  booktitle={Proceedings of the Computer Vision and Pattern Recognition Conference},
  pages={20841--20851},
  year={2025}
}

@inproceedings{rasheed2024glamm,
  title={Glamm: Pixel grounding large multimodal model},
  author={Rasheed, Hanoona and Maaz, Muhammad and Shaji, Sahal and Shaker, Abdelrahman and Khan, Salman and Cholakkal, Hisham and Anwer, Rao M and Xing, Eric and Yang, Ming-Hsuan and Khan, Fahad S},
  booktitle={Proceedings of the IEEE/CVF Conference on Computer Vision and Pattern Recognition},
  pages={13009--13018},
  year={2024}
}

@inproceedings{chen2024causalclipseg,
  title={Causalclipseg: Unlocking clip’s potential in referring medical image segmentation with causal intervention},
  author={Chen, Yaxiong and Wei, Minghong and Zheng, Zixuan and Hu, Jingliang and Shi, Yilei and Xiong, Shengwu and Zhu, Xiao Xiang and Mou, Lichao},
  booktitle={International Conference on Medical Image Computing and Computer-Assisted Intervention},
  pages={77--87},
  year={2024},
  organization={Springer}
}

@inproceedings{hu2024lga,
  title={Lga: A language guide adapter for advancing the sam model’s capabilities in medical image segmentation},
  author={Hu, Jihong and Li, Yinhao and Sun, Hao and Song, Yu and Zhang, Chujie and Lin, Lanfen and Chen, Yen-Wei},
  booktitle={International Conference on Medical Image Computing and Computer-Assisted Intervention},
  pages={610--620},
  year={2024},
  organization={Springer}
}

@inproceedings{xie2024simtxtseg,
  title={Simtxtseg: Weakly-supervised medical image segmentation with simple text cues},
  author={Xie, Yuxin and Zhou, Tao and Zhou, Yi and Chen, Geng},
  booktitle={International Conference on Medical Image Computing and Computer-Assisted Intervention},
  pages={634--644},
  year={2024},
  organization={Springer}
}

@inproceedings{wang2022cris,
  title={Cris: Clip-driven referring image segmentation},
  author={Wang, Zhaoqing and Lu, Yu and Li, Qiang and Tao, Xunqiang and Guo, Yandong and Gong, Mingming and Liu, Tongliang},
  booktitle={Proceedings of the IEEE/CVF conference on computer vision and pattern recognition},
  pages={11686--11695},
  year={2022}
}

@article{zhu2025internvl3,
  title={Internvl3: Exploring advanced training and test-time recipes for open-source multimodal models},
  author={Zhu, Jinguo and Wang, Weiyun and Chen, Zhe and Liu, Zhaoyang and Ye, Shenglong and Gu, Lixin and Tian, Hao and Duan, Yuchen and Su, Weijie and Shao, Jie and others},
  journal={arXiv preprint arXiv:2504.10479},
  year={2025}
}

@article{xu2025lingshu,
  title={Lingshu: A Generalist Foundation Model for Unified Multimodal Medical Understanding and Reasoning},
  author={Xu, Weiwen and Chan, Hou Pong and Li, Long and Aljunied, Mahani and Yuan, Ruifeng and Wang, Jianyu and Xiao, Chenghao and Chen, Guizhen and Liu, Chaoqun and Li, Zhaodonghui and others},
  journal={arXiv preprint arXiv:2506.07044},
  year={2025}
}

@article{shen2025vlm,
  title={Vlm-r1: A stable and generalizable r1-style large vision-language model},
  author={Shen, Haozhan and Liu, Peng and Li, Jingcheng and Fang, Chunxin and Ma, Yibo and Liao, Jiajia and Shen, Qiaoli and Zhang, Zilun and Zhao, Kangjia and Zhang, Qianqian and others},
  journal={arXiv preprint arXiv:2504.07615},
  year={2025}
}

@article{li2025enhancing,
  title={Enhancing Abnormality Grounding for Vision Language Models with Knowledge Descriptions},
  author={Li, Jun and Liu, Che and Bai, Wenjia and Arcucci, Rossella and Bercea, Cosmin I and Schnabel, Julia A},
  journal={arXiv preprint arXiv:2503.03278},
  year={2025}
}

@article{huang2025medseg,
  title={MedSeg-R: Reasoning Segmentation in Medical Images with Multimodal Large Language Models},
  author={Huang, Yu and Peng, Zelin and Zhao, Yichen and Yang, Piao and Yang, Xiaokang and Shen, Wei},
  journal={arXiv preprint arXiv:2506.10465},
  year={2025}
}

@article{trinh2025prs,
  title={PRS-Med: Position Reasoning Segmentation with Vision-Language Model in Medical Imaging},
  author={Trinh, Quoc-Huy and Nguyen, Minh-Van and Peng, Jung and Bagci, Ulas and Jha, Debesh},
  journal={arXiv preprint arXiv:2505.11872},
  year={2025}
}

@article{sun2025enhancing,
  title={Enhancing Step-by-Step and Verifiable Medical Reasoning in MLLMs},
  author={Sun, Haoran and Jiang, Yankai and Lou, Wenjie and Zhang, Yujie and Li, Wenjie and Wang, Lilong and Liu, Mianxin and Liu, Lei and Wang, Xiaosong},
  journal={arXiv preprint arXiv:2506.16962},
  year={2025}
}

@article{tong2025medisee,
  title={MediSee: Reasoning-based Pixel-level Perception in Medical Images},
  author={Tong, Qinyue and Lu, Ziqian and Liu, Jun and Zheng, Yangming and Lu, Zheming},
  journal={arXiv preprint arXiv:2504.11008},
  year={2025}
}

@article{liu2025seg,
  title={Seg-zero: Reasoning-chain guided segmentation via cognitive reinforcement},
  author={Liu, Yuqi and Peng, Bohao and Zhong, Zhisheng and Yue, Zihao and Lu, Fanbin and Yu, Bei and Jia, Jiaya},
  journal={arXiv preprint arXiv:2503.06520},
  year={2025}
}

@article{Qwen2.5-VL,
  title={Qwen2.5-VL Technical Report},
  author={Bai, Shuai and Chen, Keqin and Liu, Xuejing and Wang, Jialin and Ge, Wenbin and Song, Sibo and Dang, Kai and Wang, Peng and Wang, Shijie and Tang, Jun and Zhong, Humen and Zhu, Yuanzhi and Yang, Mingkun and Li, Zhaohai and Wan, Jianqiang and Wang, Pengfei and Ding, Wei and Fu, Zheren and Xu, Yiheng and Ye, Jiabo and Zhang, Xi and Xie, Tianbao and Cheng, Zesen and Zhang, Hang and Yang, Zhibo and Xu, Haiyang and Lin, Junyang},
  journal={arXiv preprint arXiv:2502.13923},
  year={2025}
}

@article{guo2025deepseek,
  title={Deepseek-r1: Incentivizing reasoning capability in llms via reinforcement learning},
  author={Guo, Daya and Yang, Dejian and Zhang, Haowei and Song, Junxiao and Zhang, Ruoyu and Xu, Runxin and Zhu, Qihao and Ma, Shirong and Wang, Peiyi and Bi, Xiao and others},
  journal={arXiv preprint arXiv:2501.12948},
  year={2025}
}

@article{lin2025healthgpt,
  title={Healthgpt: A medical large vision-language model for unifying comprehension and generation via heterogeneous knowledge adaptation},
  author={Lin, Tianwei and Zhang, Wenqiao and Li, Sijing and Yuan, Yuqian and Yu, Binhe and Li, Haoyuan and He, Wanggui and Jiang, Hao and Li, Mengze and Song, Xiaohui and others},
  journal={arXiv preprint arXiv:2502.09838},
  year={2025}
}

@article{sellergren2025medgemma,
  title={MedGemma Technical Report},
  author={Sellergren, Andrew and Kazemzadeh, Sahar and Jaroensri, Tiam and Kiraly, Atilla and Traverse, Madeleine and Kohlberger, Timo and Xu, Shawn and Jamil, Fayaz and Hughes, Cían and Lau, Charles and others},
  journal={arXiv preprint arXiv:2507.05201},
  year={2025}
}

@article{lai2025med,
  title={Med-r1: Reinforcement learning for generalizable medical reasoning in vision-language models},
  author={Lai, Yuxiang and Zhong, Jike and Li, Ming and Zhao, Shitian and Yang, Xiaofeng},
  journal={arXiv preprint arXiv:2503.13939},
  year={2025}
}

@article{ma2025medsam2,
  title={Medsam2: Segment anything in 3d medical images and videos},
  author={Ma, Jun and Yang, Zongxin and Kim, Sumin and Chen, Bihui and Baharoon, Mohammed and Fallahpour, Adibvafa and Asakereh, Reza and Lyu, Hongwei and Wang, Bo},
  journal={arXiv preprint arXiv:2504.03600},
  year={2025}
}

@article{da2024segment,
  title={Segment as You Wish--Free-Form Language-Based Segmentation for Medical Images},
  author={Da, Longchao and Wang, Rui and Xu, Xiaojian and Bhatia, Parminder and Kass-Hout, Taha and Wei, Hua and Xiao, Cao},
  journal={arXiv preprint arXiv:2410.12831},
  year={2024}
}

@article{wang2024healthg,
  title={Cogvlm: Visual expert for pretrained language models},
  author={Wang, Weihan and Lv, Qingsong and Yu, Wenmeng and Hong, Wenyi and Qi, Ji and Wang, Yan and Ji, Junhui and Yang, Zhuoyi and Zhao, Lei and XiXuan, Song and others},
  journal={Advances in Neural Information Processing Systems},
  volume={37},
  pages={121475--121499},
  year={2024}
}

@article{luo2024vividmed,
  title={Vividmed: Vision language model with versatile visual grounding for medicine},
  author={Luo, Lingxiao and Tang, Bingda and Chen, Xuanzhong and Han, Rong and Chen, Ting},
  journal={arXiv preprint arXiv:2410.12694},
  year={2024}
}

@article{chen2024huatuogpt,
  title={Huatuogpt-vision, towards injecting medical visual knowledge into multimodal llms at scale},
  author={Chen, Junying and Gui, Chi and Ouyang, Ruyi and Gao, Anningzhe and Chen, Shunian and Chen, Guiming Hardy and Wang, Xidong and Zhang, Ruifei and Cai, Zhenyang and Ji, Ke and others},
  journal={arXiv preprint arXiv:2406.19280},
  year={2024}
}

@article{gao2024mini,
  title={Mini-internvl: A flexible-transfer pocket multimodal model with 5\% parameters and 90\% performance},
  author={Gao, Zhangwei and Chen, Zhe and Cui, Erfei and Ren, Yiming and Wang, Weiyun and Zhu, Jinguo and Tian, Hao and Ye, Shenglong and He, Junjun and Zhu, Xizhou and others},
  journal={arXiv preprint arXiv:2410.16261},
  year={2024}
}

@article{shao2024deepseekmath,
  title={Deepseekmath: Pushing the limits of mathematical reasoning in open language models},
  author={Shao, Zhihong and Wang, Peiyi and Zhu, Qihao and Xu, Runxin and Song, Junxiao and Bi, Xiao and Zhang, Haowei and Zhang, Mingchuan and Li, YK and Wu, Y and others},
  journal={arXiv preprint arXiv:2402.03300},
  year={2024}
}

@misc{openai2024o1,
  author = {OpenAI},
  title = {OpenAI o1},
  year = {2024},
  howpublished = {\url{https://openai.com/o1/}},
}

@misc{gpt4o,
  author       = {{OpenAI}},
  title        = {{GPT‑4o (GPT‑4 Omni)}},
  howpublished = {Online},
  year         = {2024},
  month        = may,
  day          = 13,
  url          = {https://openai.com/index/hello-gpt-4o/}
}

@misc{gemini2.5_flash,
  author       = {{Google}},
  title        = {{Gemini-2.5-Flash}},
  howpublished = {Online; GA release},
  year         = {2025},
  month        = jun,
  day          = 17,
  url          = {https://cloud.google.com/vertex-ai/generative-ai/docs/models/gemini/2-5-flash}
}

@article{ma2024segment,
  title={Segment anything in medical images},
  author={Ma, Jun and He, Yuting and Li, Feifei and Han, Lin and You, Chenyu and Wang, Bo},
  journal={Nature Communications},
  volume={15},
  number={1},
  pages={654},
  year={2024},
  publisher={Nature Publishing Group UK London}
}

@article{wei2022chain,
  title={Chain-of-thought prompting elicits reasoning in large language models},
  author={Wei, Jason and Wang, Xuezhi and Schuurmans, Dale and Bosma, Maarten and Xia, Fei and Chi, Ed and Le, Quoc V and Zhou, Denny and others},
  journal={Advances in neural information processing systems},
  volume={35},
  pages={24824--24837},
  year={2022}
}

@article{uesato2022solving,
  title={Solving math word problems with process-and outcome-based feedback},
  author={Uesato, Jonathan and Kushman, Nate and Kumar, Ramana and Song, Francis and Siegel, Noah and Wang, Lisa and Creswell, Antonia and Irving, Geoffrey and Higgins, Irina},
  journal={arXiv preprint arXiv:2211.14275},
  year={2022}
}

@article{chen2021transunet,
  title={Transunet: Transformers make strong encoders for medical image segmentation},
  author={Chen, Jieneng and Lu, Yongyi and Yu, Qihang and Luo, Xiangde and Adeli, Ehsan and Wang, Yan and Lu, Le and Yuille, Alan L and Zhou, Yuyin},
  journal={arXiv preprint arXiv:2102.04306},
  year={2021}
}

@article{li2024abdomenatlas,
  title={Abdomenatlas: A large-scale, detailed-annotated, \& multi-center dataset for efficient transfer learning and open algorithmic benchmarking},
  author={Li, Wenxuan and Qu, Chongyu and Chen, Xiaoxi and Bassi, Pedro RAS and Shi, Yijia and Lai, Yuxiang and Yu, Qian and Xue, Huimin and Chen, Yixiong and Lin, Xiaorui and others},
  journal={Medical Image Analysis},
  volume={97},
  pages={103285},
  year={2024},
  publisher={Elsevier}
}

@article{zhao2024biomedparse,
  title={BiomedParse: a biomedical foundation model for image parsing of everything everywhere all at once},
  author={Zhao, Theodore and Gu, Yu and Yang, Jianwei and Usuyama, Naoto and Lee, Ho Hin and Naumann, Tristan and Gao, Jianfeng and Crabtree, Angela and Abel, Jacob and Moung-Wen, Christine and others},
  journal={arXiv preprint arXiv:2405.12971},
  year={2024}
}

@article{ye2023sa,
  title={Sa-med2d-20m dataset: Segment anything in 2d medical imaging with 20 million masks},
  author={Ye, Jin and Cheng, Junlong and Chen, Jianpin and Deng, Zhongying and Li, Tianbin and Wang, Haoyu and Su, Yanzhou and Huang, Ziyan and Chen, Jilong and Jiang, Lei and others},
  journal={arXiv preprint arXiv:2311.11969},
  year={2023}
}

@inproceedings{liu2021slake,
  title={Slake: A semantically-labeled knowledge-enhanced dataset for medical visual question answering},
  author={Liu, Bo and Zhan, Li-Ming and Xu, Li and Ma, Lin and Yang, Yan and Wu, Xiao-Ming},
  booktitle={2021 IEEE 18th international symposium on biomedical imaging (ISBI)},
  pages={1650--1654},
  year={2021},
  organization={IEEE}
}

@article{he2020pathvqa,
  title={Pathvqa: 30000+ questions for medical visual question answering},
  author={He, Xuehai and Zhang, Yichen and Mou, Luntian and Xing, Eric and Xie, Pengtao},
  journal={arXiv preprint arXiv:2003.10286},
  year={2020}
}

@article{lau2018dataset,
  title={A dataset of clinically generated visual questions and answers about radiology images},
  author={Lau, Jason J and Gayen, Soumya and Ben Abacha, Asma and Demner-Fushman, Dina},
  journal={Scientific data},
  volume={5},
  number={1},
  pages={1--10},
  year={2018},
  publisher={Nature Publishing Group} 
}

@article{li2024language,
  title={Language-guided Medical Image Segmentation with Target-informed Multi-level Contrastive Alignments},
  author={Li, Mingjian and Meng, Mingyuan and Ye, Shuchang and Fulham, Michael and Bi, Lei and Kim, Jinman},
  journal={arXiv preprint arXiv:2412.13533},
  year={2024}
}

@article{huang2024cross,
  title={Cross-modal conditioned reconstruction for language-guided medical image segmentation},
  author={Huang, Xiaoshuang and Li, Hongxiang and Cao, Meng and Chen, Long and You, Chenyu and An, Dong},
  journal={IEEE Transactions on Medical Imaging},
  year={2024},
  publisher={IEEE}
}

@article{li2023lvit,
  title={Lvit: language meets vision transformer in medical image segmentation},
  author={Li, Zihan and Li, Yunxiang and Li, Qingde and Wang, Puyang and Guo, Dazhou and Lu, Le and Jin, Dakai and Zhang, You and Hong, Qingqi},
  journal={IEEE transactions on medical imaging},
  volume={43},
  number={1},
  pages={96--107},
  year={2023},
  publisher={IEEE}
}

@inproceedings{hu2016segmentation,
  title={Segmentation from natural language expressions},
  author={Hu, Ronghang and Rohrbach, Marcus and Darrell, Trevor},
  booktitle={European conference on computer vision},
  pages={108--124},
  year={2016},
  organization={Springer}
}

@inproceedings{yang2022lavt,
 title={Lavt: Language-aware vision transformer for referring image segmentation},
 author={Yang, Zhao and Wang, Jiaqi and Tang, Yansong and Chen, Kai and Zhao, Hengshuang and Torr, Philip HS},
 booktitle={Proceedings of the IEEE/CVF conference on computer vision and pattern recognition},
 pages={18155--18165},
 year={2022}
}

@inproceedings{lin2024stable,
  title={Stable diffusion segmentation for biomedical images with single-step reverse process},
  author={Lin, Tianyu and Chen, Zhiguang and Yan, Zhonghao and Yu, Weijiang and Zheng, Fudan},
  booktitle={International Conference on Medical Image Computing and Computer-Assisted Intervention},
  pages={656--666},
  year={2024},
  organization={Springer}
}

@article{diao2025driverx,
  title={DriveRX: A Vision-Language Reasoning Model for Cross-Task Autonomous Driving},
  author={Diao, Muxi and Yang, Lele and Yin, Hongbo and Wang, Zhexu and Wang, Yejie and Tian, Daxin and Liang, Kongming and Ma, Zhanyu},
  journal={arXiv preprint arXiv:2505.20665},
  year={2025}
}

@article{wang2025harnessing,
  title={Harnessing Caption Detailness for Data-Efficient Text-to-Image Generation},
  author={Wang, Xinran and Diao, Muxi and Liu, Yuanzhi and Wang, Chunyu and Liang, Kongming and Ma, Zhanyu and Guo, Jun},
  journal={arXiv preprint arXiv:2505.15172},
  year={2025}
}

@article{song2024cs,
  title={Cs-bench: A comprehensive benchmark for large language models towards computer science mastery},
  author={Song, Xiaoshuai and Diao, Muxi and Dong, Guanting and Wang, Zhengyang and Fu, Yujia and Qiao, Runqi and Wang, Zhexu and Fu, Dayuan and Wu, Huangxuan and Liang, Bin and others},
  journal={arXiv preprint arXiv:2406.08587},
  year={2024}
}

@article{wang2025cinetechbench,
  title={CineTechBench: A Benchmark for Cinematographic Technique Understanding and Generation},
  author={Wang, Xinran and Xu, Songyu and Shan, Xiangxuan and Zhang, Yuxuan and Diao, Muxi and Duan, Xueyan and Huang, Yanhua and Liang, Kongming and Ma, Zhanyu},
  journal={arXiv preprint arXiv:2505.15145},
  year={2025}
}

@inproceedings{diao2025seas,
  title={Seas: Self-evolving adversarial safety optimization for large language models},
  author={Diao, Muxi and Li, Rumei and Liu, Shiyang and Liao, Guogang and Wang, Jingang and Cai, Xunliang and Xu, Weiran},
  booktitle={Proceedings of the AAAI Conference on Artificial Intelligence},
  volume={39},
  pages={23778--23786},
  year={2025}
}

@inproceedings{qiao2025we,
  title={We-math: Does your large multimodal model achieve human-like mathematical reasoning?},
  author={Qiao, Runqi and Tan, Qiuna and Dong, Guanting and MinhuiWu, MinhuiWu and Sun, Chong and Song, Xiaoshuai and Wang, Jiapeng and Gongque, Zhuoma and Lei, Shanglin and Zhang, Yifan and others},
  booktitle={Proceedings of the 63rd Annual Meeting of the Association for Computational Linguistics (Volume 1: Long Papers)},
  pages={20023--20070},
  year={2025}
}

\clearpage
\newpage
\appendix
\setcounter{secnumdepth}{2}
\renewcommand{\thesection}{\Alph{section}}
\renewcommand{\thesubsection}{\Alph{section}.\arabic{subsection}}

\section{More Details on UMRG Task} \label{apd:task}

\subsection{Referring Image Segmentation}
Referring Image Segmentation (RIS) aims to segment a specific object within an image according to a natural language expression. First introduced in \cite{hu2016segmentation}, RIS addresses the need for fine-grained, language-driven visual understanding, enabling users to interact with visual systems in a more intuitive and flexible manner. Unlike traditional segmentation tasks, RIS supports flexible object descriptions, enabling intuitive interaction.

Formally, the RIS task can be defined as follows: given an image \( I \in \mathbb{R}^{H \times W \times 3} \) and a natural language expression \( Q = \{q_1, q_2, \dots, q_n\} \), the goal is to predict a binary segmentation mask \( M \in \{0,1\}^{H \times W} \) that identifies the region corresponding to the object referred to by \( Q \).Recent methods fall into three main categories:

\paragraph{Explicit Language Reasoning.} This line of work narrows the language-vision gap by generating reasoning chains from text.
LISA~\cite{lai2024lisa} introduces reasoning supervision via a frozen LLM and achieves strong zero-shot performance, but its token-level reasoning and reliance on supervision limit generalization. Seg-Zero~\cite{liu2025seg} improves on this with a decoupled architecture trained by reinforcement learning, enabling explicit reasoning and superior performance without annotated reasoning data.

\paragraph{Fine-grained Cross-modal Alignment.} 
These methods improve pixel-text alignment via fine-grained feature interactions.
CRIS~\cite{wang2022cris} employs contrastive learning for region-expression alignment but struggles with hard negatives and compositional queries. LAVT~\cite{yang2022lavt} enhances cross-modal encoding via early fusion, though its rigid architecture limits flexibility. GLaMM~\cite{rasheed2024glamm} enables multi-turn pixel-level grounding but requires extensive pretraining and infrastructure, hindering adaptability.

\paragraph{Foundation Model Augmentation.}
This category integrates general-purpose segmentation models for enhanced control.
SAM4MLLM~\cite{chen2024sam4mllm} integrates SAM with a MLLM through refinement modules, improving mask accuracy though still limited by semantic misalignment.

\subsection{Language-Guided Medical Image Segmentation}
Language-Guided Medical Image Segmentation (LGMIS) aims to segment anatomical structures or pathological regions in medical images based on natural language instructions.  It supports flexible expressions across modalities.

Given a medical image \( I \in \mathbb{R}^{H \times W \times C} \), typically grayscale or multi-channel (e.g., CT, MRI), and a clinical instruction \( Q = \{q_1, q_2, \dots, q_n\} \), the objective is to generate a segmentation mask \( M \in \{0,1\}^{H \times W} \) that delineates anatomical or pathological structures referenced by \( Q \). Existing methods fall into two key categories:

\paragraph{Prompt Generation and Geometric Awareness.}  To support interactive and context-sensitive segmentation, some methods generate language prompts dynamically and enforce geometric consistency. FLanS~\cite{da2024segment} uses retrieval-augmented prompts with a geometry-aware model for orientation-consistent segmentations. Besides, CLIP-based models~\cite{liu2023clip} inject general language priors to support zero-shot anatomical segmentation.

\paragraph{Foundation Model Integration.}Another line of research enhances LGMIS by combining pretrained vision-language and segmentation models under weak or zero-shot supervision. MedCLIP-SAM~\cite{koleilat2024medclip} fuses BiomedCLIP and SAM with ScoreCAM-based supervision to achieve weakly supervised yet accurate segmentation.

\subsection{Referring Lesion Segmentation}
Referring Lesion Segmentation (RLS) aims to identify lesion regions in medical images based on natural language descriptions. Compared to RIS, RLS presents unique challenges: the complexity of medical terminology, irregular lesion morphology, and the often vague nature of clinical language. Solving this task requires precise cross-modal understanding and robustness to ambiguous inputs.

Formally, given a medical image \( I \in \mathbb{R}^{H \times W \times C} \) and a lesion-focused referring phrase \( Q = \{q_1, q_2, \dots, q_n\} \), the task is to generate a binary mask \( M \in \{0,1\}^{H \times W} \) that accurately localizes the lesion referred to by \( Q \). Recent methods fall into two main categories:

\paragraph{Semantic Alignment and Reasoning.} 
These approaches aim to enhance the alignment between clinical language and visual features, often through contrastive learning, causal reasoning, or language-conditioned modeling.
Li et al.~\cite{li2024language} apply multi-level contrastive learning, while CausalCLIPSeg~\cite{chen2024causalclipseg} introduces causal interventions to suppress spurious correlations. LViT~\cite{li2023lvit} incorporates language-guided attention into transformers for holistic lesion understanding.

\paragraph{Foundation Model Adaptation and Prompt Guidance.}
This direction focuses on extending the capabilities of pretrained segment models to the medical domain with minimal supervision.
LGA~\cite{hu2024lga} enables controllable segmentation by injecting language cues into SAM via a lightweight adapter. SimTxtSeg~\cite{xie2024simtxtseg} and~\cite{huang2024cross} utilize prompt-based pseudo-labeling and language-conditioned reconstruction for weakly supervised alignment.

\subsection{Unified Medical Reasoning Grounding}
Previous tasks have been based on the assumption that clear references to segmented objects are available, which is often challenging in real-world scenarios. Even methods that allow for free-form language descriptions still require unambiguous prompts. In contrast, UMRG starts from a vague reference to anatomical structures and ultimately generates pixel-level segmentation results. This approach alleviates the burden of annotation in the overall workflow and aligns with real-world contexts. In terms of methodology, prior tasks that involve reasoning focused on clarifying the segmentation objects, often decoupling them from the actual images. Our task, for the first time, associates reasoning with grounding, emphasizing the image cues and a CoT aimed at grounding. The process can be formulated as follows:

Given a medical image $\mathcal{I}$ and a clinical query $\mathcal{Q}$ containing implicit referring expressions, the model $\mathbf{G}$ outputs a bounding box $\mathcal{B}$, two semantic key points $\mathcal{P}_1$ and $\mathcal{P}_2$, and a pixel-level segmentation mask $\mathcal{M}$. The Task is:
\begin{equation}
\{\mathcal{T},\mathcal{B},\mathcal{P}_1,\mathcal{P}_2,\mathcal{M}\}=\mathbf{G}(\mathcal{I},\mathcal{Q}).
\label{apd:umrg_task}
\end{equation}

\section{More Details on U-MRG-14K Dataset} \label{apd:dataset}

\begin{figure*}[t]
    \centering
    \includegraphics[width=1\linewidth]{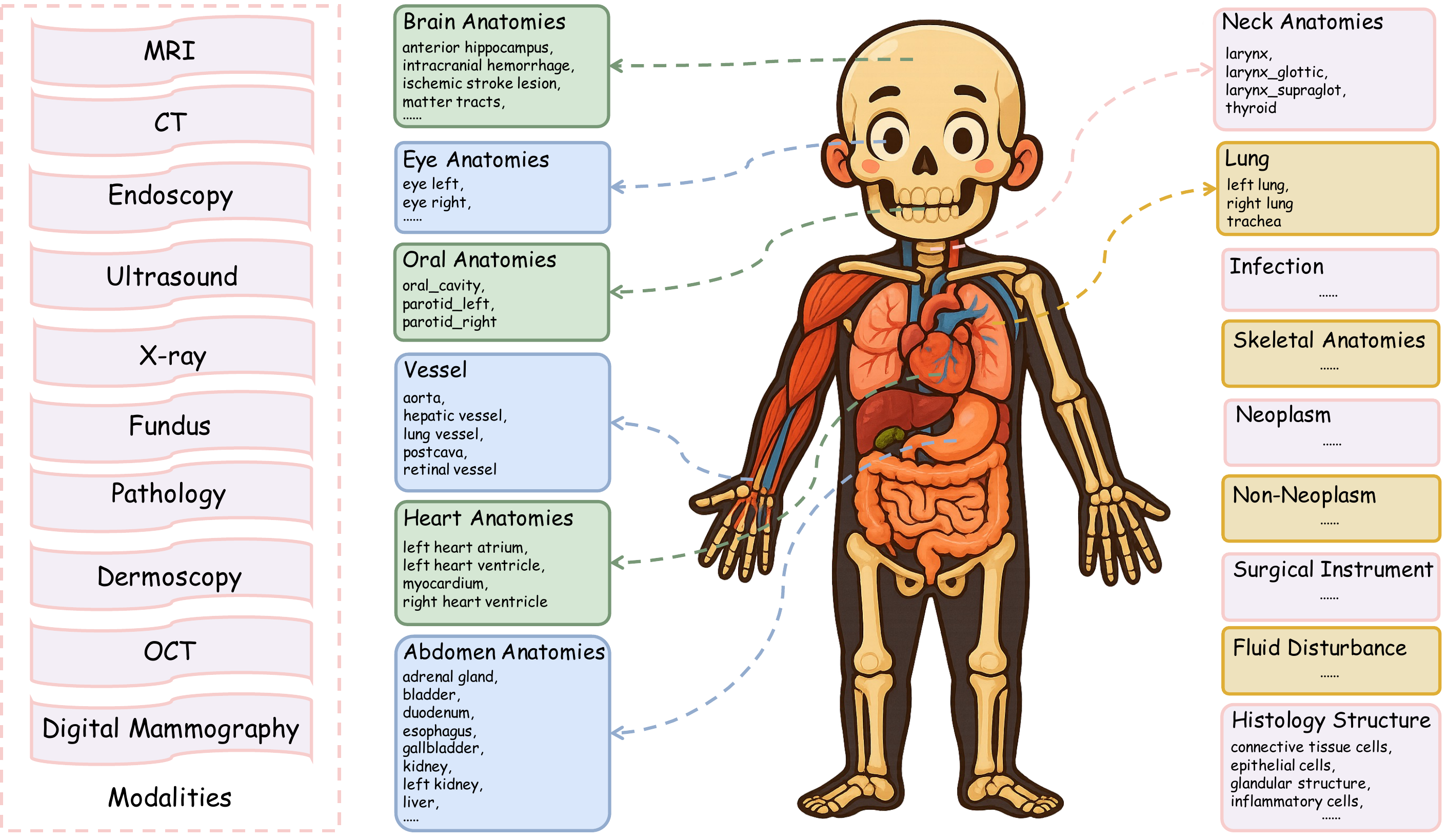}
    \caption{Overview of Modalities and Hierarchical Category Structure in the U-MRG-14k Dataset.}
    \label{fig:modal_supercate}
\end{figure*}

\subsection{Meta Information of U-MRG-14K}

Following the established methodologies for constructing diverse datasets \cite{qiao2025we,song2024cs,diao2025seas,diao2025driverx,wang2025harnessing,wang2025cinetechbench}, we design and curate U-MRG-14K with a focus on medical image reasoning across multiple modalities and hierarchical categories.

\paragraph{Comprehensive Image Annotations.}
To support the generation of faithful and context-aware QA data, we construct enriched meta information for each image. Specifically, we manually annotate each image with key attributes, including \textbf{imaging modality}, \textbf{subject type} (patient or healthy individual), the \textbf{category of the localized structure} (organ or anatomical region), and its broader \textbf{super-category} (e.g., \textit{body system} or \textit{scan region}). This meta information ensures consistency across diverse sources and helps establish a structured understanding of the image content. Building on these annotations, we use GPT-4o to generate detailed descriptions using a three-component prompting strategy: \textbf{task definition}, \textbf{information injection}, and \textbf{task refinement}.

\paragraph{Prompt Construction.}
In the task definition part, we specify the system role and generation principles to constrain the model toward producing accurate and distinctive descriptions. The prompts are iteratively refined to avoid common pitfalls such as diagnostic suggestions or overly generic outputs. For the information injection part, the annotated attributes are translated into natural language with explicit explanations to enhance the model’s comprehension. To precisely anchor the described region, we also provide the model with the actual \textbf{segmentation mask}, its \textbf{bounding box}, the point \textbf{farthest from the mask boundary}, and the point \textbf{relatively farthest from it}. To support object-specific adaptation, we categorize the target regions as normal structures, lesions, or ambiguous objects, prompting the model to construct the description from appropriate perspectives—for instance, emphasizing \textit{physiological function} for normal structures, \textit{clinical impact} for lesions, and \textit{strict visual fidelity} for ambiguous regions while explicitly avoiding unwarranted assumptions.

\paragraph{Description Generation.}
Using these carefully constructed prompts, we generate two complementary descriptions per image–mask pair: (i) a \textbf{short description} focused on intuitive visual features (e.g., \textit{shape}, \textit{texture}, \textit{absolute and relative location}), while avoiding medical terminology; and (ii) a \textbf{long description} that integrates both medical and imaging knowledge, emphasizing the distinctiveness of the region within its category and in contrast to surrounding structures. Notably, we observe that for long descriptions, the model tends to default to general medical knowledge rather than precise, image-grounded observations. To address this, we require that \textbf{at least half} of the content in the long description be directly grounded in observable visual features. This generation strategy is specifically designed to bridge the gap observed in the Unified Medical Reasoning Grounding (\textbf{UMRG}) task, where models often struggle to connect textual interpretations with concrete visual evidence, particularly in clinically nuanced or ambiguous cases. To ensure the quality and reliability of the generated descriptions, all outputs undergo a manual review process, focusing on accuracy, visual relevance, and clinical plausibility.Although we use these descriptions to generate QA data, in the UMRG task, models still struggle to understand and utilize this information directly from the image. Nevertheless, we believe these detailed annotations can serve as a stepping stone toward building a general framework for reasoning grounding in medical images. The example of final generated \textit{meta information} is illustrated in Appendix~\ref{case:meta_infor}.

\subsection{QA Formats of U-MRG-14K}

\paragraph{Context-Aware Prompt Conditioning.}
For each super-category, we design a short \textbf{category scope} prompt to set the GPT-4o’s clinical focus before QA generation. If the super-category is pathological (e.g., \textit{neoplasm}, \textit{infection}), the prompt guides the model to reason about abnormal findings such as tissue changes, lesion extent, and diagnostic uncertainty. If the super-category is  anatomical (e.g., \textit{Lung}, \textit{Abdomen}), it shifts focus to normal structure, spatial relations, and physiological function, while discouraging disease assumptions.
A short list of fine-grained subclasses (e.g., \textit{left lung}, \textit{right kidney}) is provided for context, but GPT-4o is instructed not to repeat these terms. This conditioning ensures that all questions follow the intended clinical perspective and remain deliberately vague.

\paragraph{Schema-Guided QA-Format Generation.}
After setting the clinical scope, we present GPT-4o with one structured prompt that requests exactly $N$ question–answer formats (default $N = 20$). The prompt explains that each question should resemble a vague inquiry from a patient, pointing to the target only through visual or functional clues while avoiding technical labels. Each answer is limited to five sentences and must describe a clear, step by step visual reasoning path without offering a diagnosis. To ensure variety, the instructions demand that wording, cue type, and reasoning style differ across formats. Finally, the model must return its output as a strict JSON object that contains the super-category name and a list of \{\textit{id}, \textit{question}, \textit{answer}\}. Because this schema is embedded in the prompt, the result is immediately machine-verifiable and ready for downstream use. Combined with the context-aware prompt, this procedure yields QA formats that are clinically sound, broadly applicable within each super-category, and fully consistent with the UMRG evaluation protocol. The example of final generated \textit{QA formats} is illustrated in Appendix~\ref{case:qa_formats}.

\subsection{QA Pairs of U-MRG-14K}

\paragraph{QA Pair Design for the UMRG Task.}
The QA pair data is designed to serve the \textbf{Unified Medical Reasoning Grounding (UMRG) task}, which is closely aligned with real-world scenarios. The detailed reasoning data included also provides room for future expansion in subsequent SFT work. To achieve this, we designed a specific format for our question-answering pairs that differs significantly from traditional VQA datasets. Our deliberately \textbf{ambiguous questions} are crafted to train the model's foundational visual reasoning and localization capabilities, rather than simple object recognition. Critically, the corresponding answers are designed as explicit Chain-of-Thought (\textbf{CoT}) reasoning pathways. The intention is that these detailed logical inferences are particularly beneficial for bootstrapping a model's reasoning abilities from a cold start.

\paragraph{A Multi-part Prompting Framework.}
To generate QA pairs that meet the above design, we developed a detailed, multi-part Prompting Framework that guides the model through a sequence of understanding, reasoning, and generation; such a structured approach is essential for managing the complexity of clinical reasoning. This framework is composed of three core components delivered in a single, cohesive prompt. First, it \textbf{assigns the model the role} of a professional radiologist; this initial instruction primes the model to activate its domain-specific knowledge and adopt a professional, analytical tone, rather than a conversational one. Second, it grounds the model in facts by injecting a rich, multimodal context. This includes the medical image, a segmentation mask, and extensive metadata such as \textit{imaging modality, patient health status, anatomical classifications (super-category, category), precise spatial coordinates (bbox, key points), and textual descriptions}. Providing this comprehensive evidence base is vital to minimize hallucination, a common failure mode for LLMs, and ensures all reasoning is anchored in verifiable data. Third, it defines the core task: the model must \textbf{revise a given QA template} to align with the provided image content. This revision process is governed by a strict set of rules designed to elicit deep reasoning. The question must be revised to be vague and indirect, grounded in the region's visual attributes without revealing the category name, which forces the model to engage in genuine visual search rather than simple keyword matching. The answer must follow a \textbf{step-by-step logical path} based solely on observable visual features, remain non-diagnostic, and clearly identify the true category, making its reasoning process transparent and auditable. To enforce this deep visual reasoning, the most notable principle is the \textbf{“Pretend the Mask is Unavailable” Principle}. This instruction is critical as it compels the model to mimic human expert cognition—analyzing the broader anatomical context to progressively narrow down and localize the specific finding, rather than taking a shortcut by simply referring to the mask's coordinates. This entire strategy culminates in a structured JSON output, which facilitates reliable downstream processing and automated evaluation of the generated dataset. The example of final generated \textit{QA pairs} is illustrated in Appendix~\ref{case:qa_pairss}.

\begin{figure}[t]
    \centering
    \includegraphics[width=0.8\linewidth]{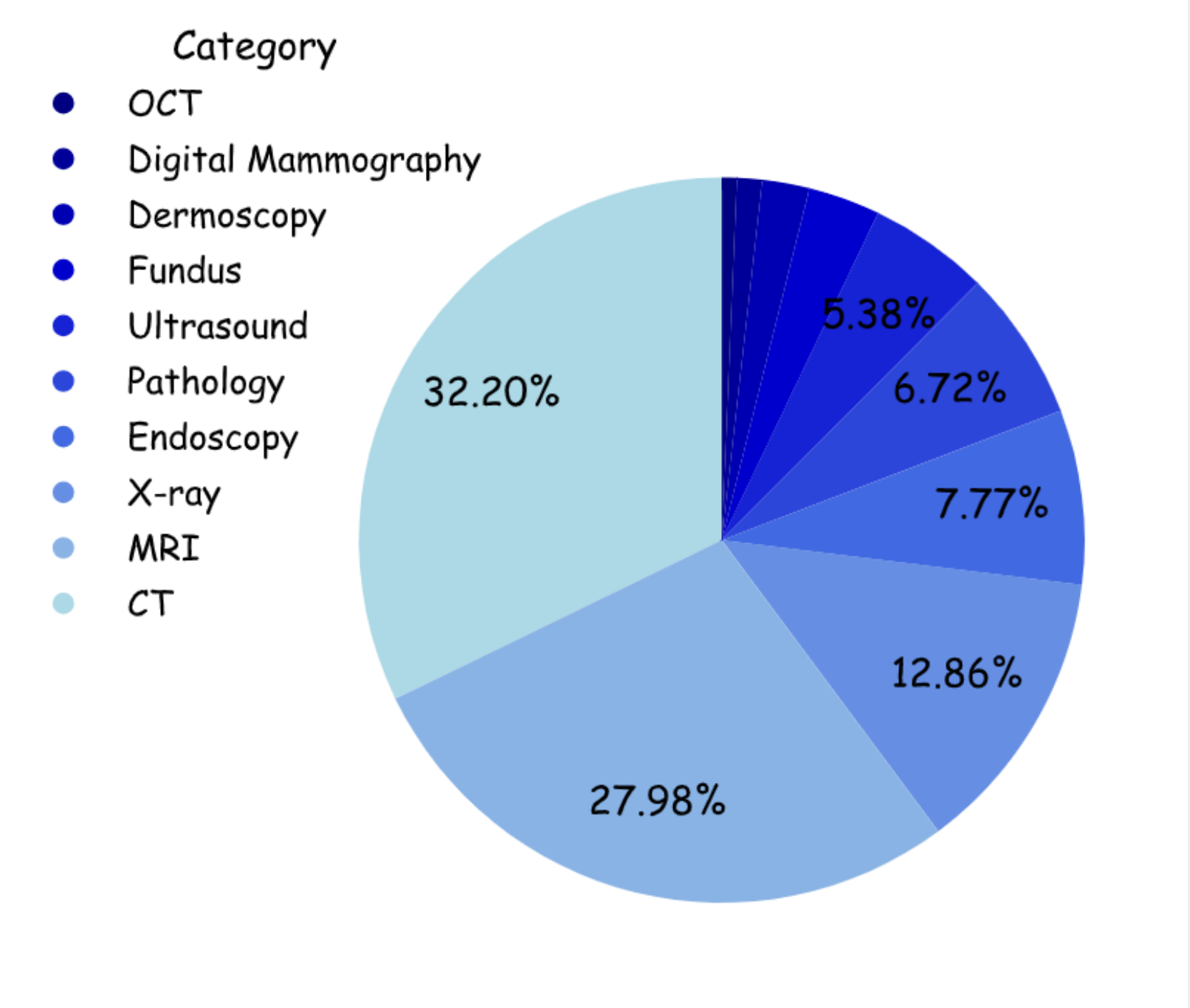}
    \caption{Pie chart illustrating the distribution of imaging modalities in the whole U-MRG-14K dataset.}
    \label{fig:statistics_modal}
\end{figure}

\begin{figure}[t]
    \centering
    \includegraphics[width=0.6\linewidth]{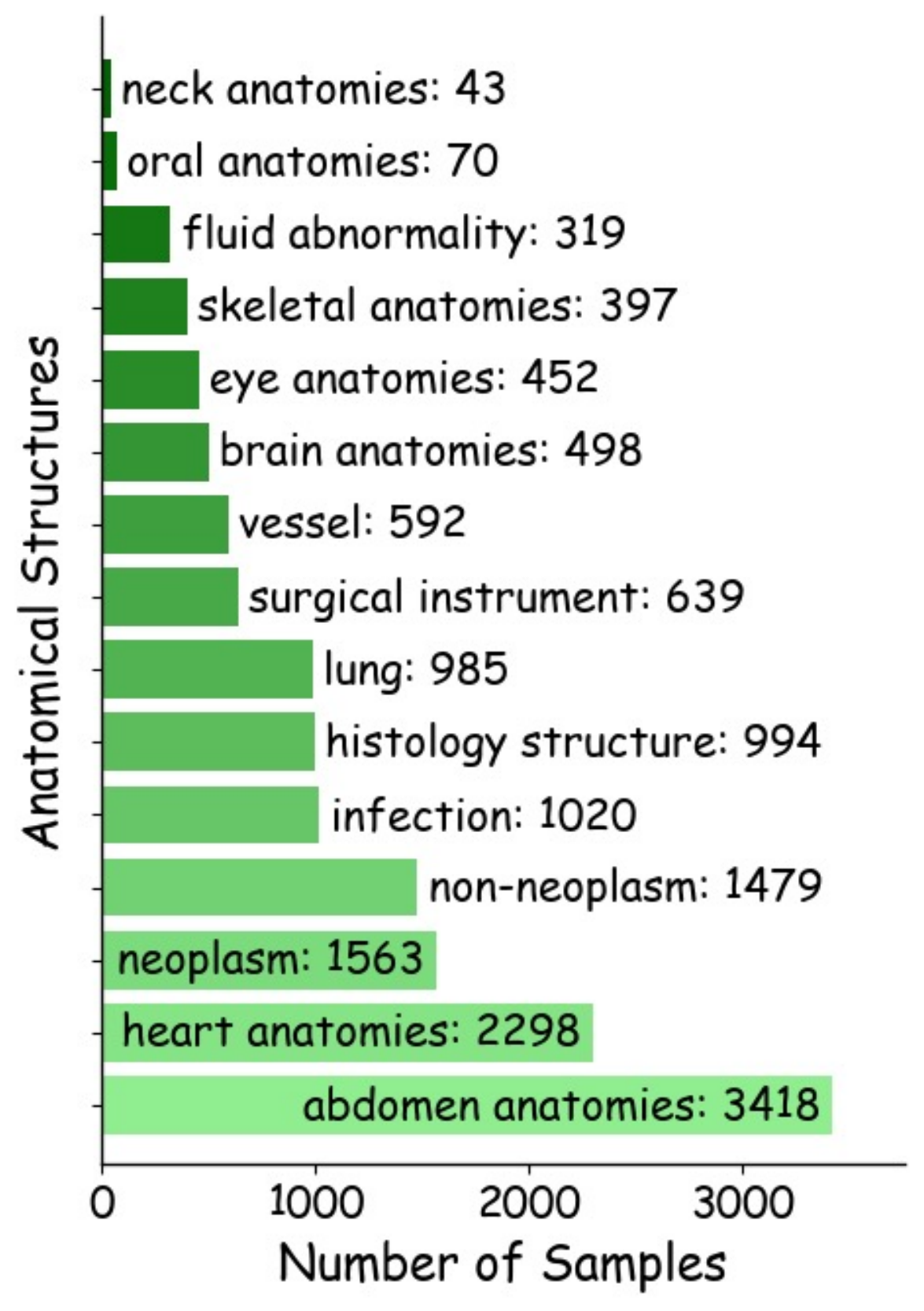}
    \caption{Bar chart illustrating the distribution of super-categories in the whole U-MRG-14K dataset.}
    \label{fig:statistics_super}
\end{figure}

\begin{figure}[t]
    \centering
    \includegraphics[width=1\linewidth]{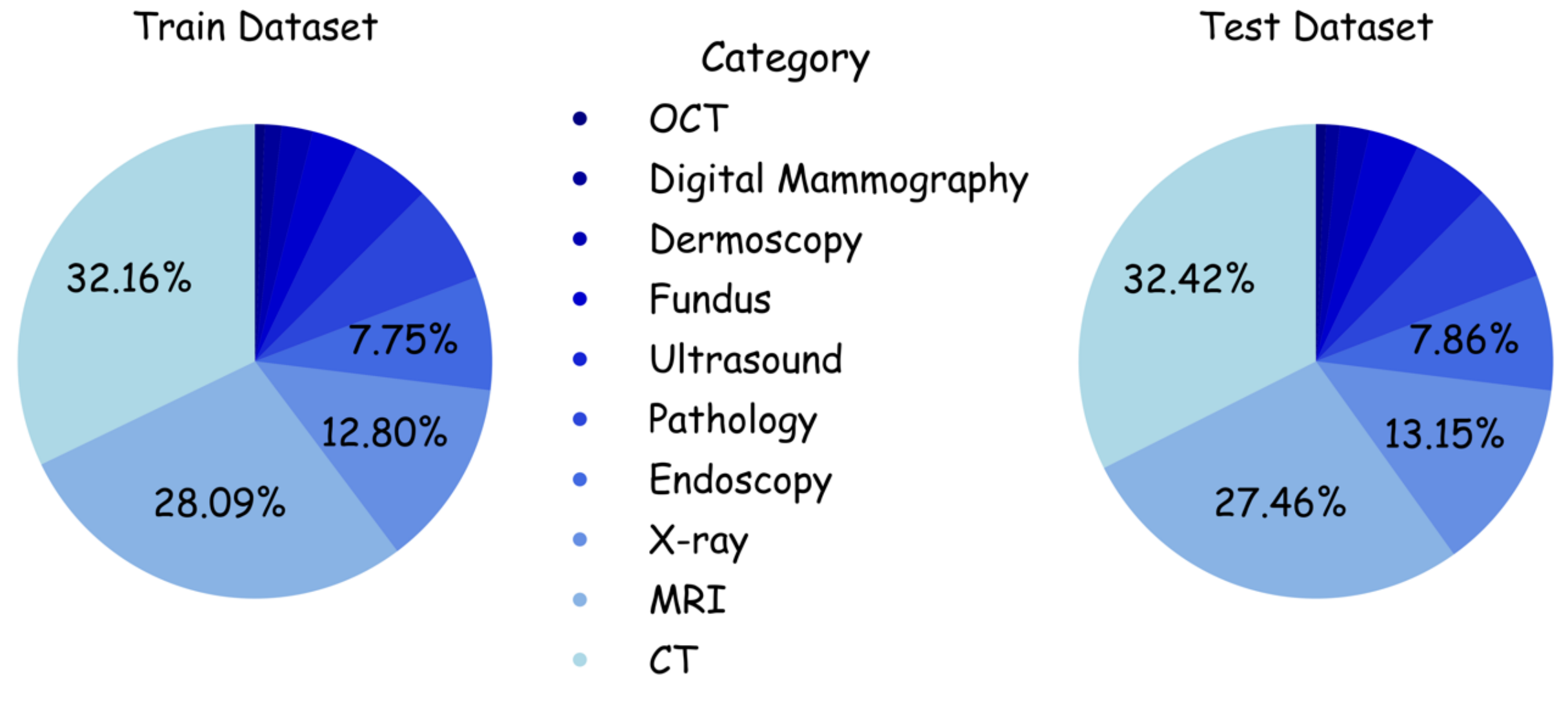}
    \caption{Pie chart illustrating the distribution of imaging modalities in the U-MRG-14K dataset. The \textbf{left} panel shows the percentage of samples per modality in the training set, while the \textbf{right} panel shows the corresponding distribution in the test set.}
    \label{fig:statistics_pie}
\end{figure}

\begin{figure}[t]
    \centering
    \includegraphics[width=1\linewidth]{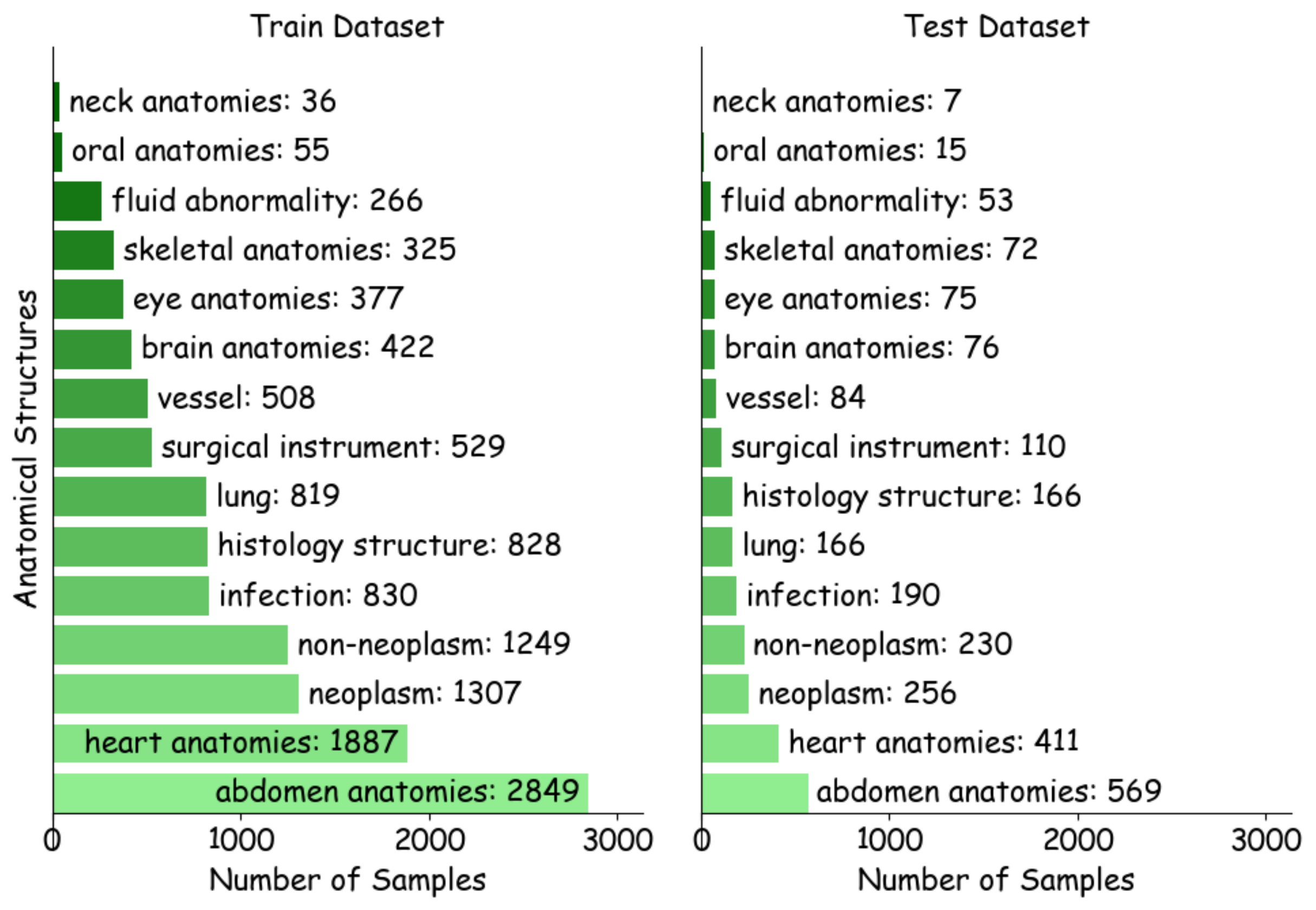}
    \caption{Bar chart illustrating the distribution of super-categories in the U-MRG-14K dataset. The \textbf{left} panel shows the number of samples per super-category in the training set, while the \textbf{right} panel shows the corresponding distribution in the test set.}
    \label{fig:statistics_bar}
\end{figure}

\subsection{Dataset Statistics} \label{apd:statistics}
U-MRG-14K is, to our knowledge, the first dataset that combines implicit clinical questions with pixel annotations (\textit{bounding boxes}, \textit{interior key points}, and \textit{masks}), across a wide range of imaging studies. It spans \textbf{10 medical modalities} (CT, MRI, ultrasound, histology, etc.), \textbf{15 super-categories}, and \textbf{108 fine-grained categories}. A summary of the overall distribution is shown in Fig.~\ref{fig:modal_supercate}. The specific proportions of these imaging modalities and super-categories are illustrated in Fig.~\ref{fig:statistics_modal} and Fig.~\ref{fig:statistics_super}.

\paragraph{Detailed Distribution.}
The four largest super-categories in the dataset, namely \textit{abdomen}, \textit{heart}, \textit{neoplasm}, and \textit{non-neoplasm}, collectively account for \textbf{59 \%} of all samples. This highlights the dataset's alignment with the predominant trends in current open-source medical datasets.
We partition U-MRG-14K into distinct training and testing splits to support systematic evaluation.
The distribution of imaging modalities within each split is shown in Fig.~\ref{fig:statistics_pie}, and the super-category composition is illustrated in Fig.~\ref{fig:statistics_bar}.

\paragraph{Future Extension.}
We are actively expanding the underlying collection pipeline and will release updated versions that broaden modality and category coverage, ultimately providing a more comprehensive resource to support the development and evaluation of solutions for the UMRG task.

\section{More Details on MedReasoner Framework}

\subsection{Group Relative Policy Optimization.}
We utilize Group Relative Policy Optimization (GRPO) \cite{shao2024deepseekmath} as our reinforcement learning strategy. is an efficient reinforcement learning algorithm that eliminates the need for a value network by leveraging group-wise relative advantages. For each input query $q$, GRPO samples a group of $G$ outputs $\{o_i\}_{i=1}^G$ from the old policy $\pi_{\text{old}}$, scores them using a reward model $r_\phi$, and computes normalized relative advantages within the group. The policy $\pi_\theta$ is updated by maximizing the following objective:
\begin{equation}
    \begin{aligned}
        J_{\mathrm{GRPO}}(\theta) = \mathbb{E}_{q, \{o_i\}_{i=1}^G \sim \pi_{\text{old}}} \Bigg[ 
        \frac{1}{G} \sum_{i=1}^{G} \frac{1}{|o_i|} \sum_{t=1}^{|o_i|} \Bigg( \\
        \min\Big( 
        \frac{\pi_{\theta}(o_{i,t} \mid q, o_{i,<t})}{\pi_{\text{old}}(o_{i,t} \mid q, o_{i,<t})} \hat{A}_{i,t}, \\
        \text{clip}\big(
        \frac{\pi_{\theta}(o_{i,t} \mid q, o_{i,<t})}{\pi_{\text{old}}(o_{i,t} \mid q, o_{i,<t})}, 1{-}\epsilon, 1{+}\epsilon
        \big) \hat{A}_{i,t} \Big) \\
        - \beta \, D_{\mathrm{KL}}\left[\pi_\theta \,\|\, \pi_{\text{ref}}\right] 
        \Bigg) 
        \Bigg]
    \end{aligned}
\end{equation}
where $r_{i,t} = \frac{\pi_{\theta}(o_{i,t}|q, o_{i,<t})}{\pi_{\text{old}}(o_{i,t}|q, o_{i,<t})}$ is the token-level importance ratio, and $\hat{A}_{i,t}$ denotes the normalized advantage computed via one of the following supervision strategies:

\paragraph{Outcome Supervision.} A single scalar reward $r_i$ is assigned to each sampled output $o_i$ by the reward model. The group-wise normalized reward is computed as:
\begin{equation}
\hat{A}_{i,t} = \tilde{r}_i = \frac{r_i - \mu_r}{\sigma_r}, \quad \forall t,
\end{equation}
where $\mu_r$ and $\sigma_r$ denote the mean and standard deviation of $\{r_1, ..., r_G\}$.

\paragraph{Process Supervision.} Step-level rewards $r_{i}^{(j)}$ are assigned to each intermediate reasoning step based on its contribution, and the advantage is computed accordingly as:
\begin{equation}
\hat{A}_{i,t} = \sum_{j: \text{index}(j) \ge t} \tilde{r}_{i}^{(j)}, \quad \text{where } \tilde{r}_{i}^{(j)} = \frac{r_{i}^{(j)} - \mu_R}{\sigma_R}.
\end{equation}
GRPO aligns well with the relative nature of reward models trained on pairwise preference data, and it significantly reduces the computational burden by avoiding the training of a separate value network.

\subsection{User Prompts} \label{apd:user_prompts}

\paragraph{Reasoning User Prompt.}
The design of these prompts is guided by principles that ensure precise, clinically grounded visual grounding within a tightly defined output format. The prompt \textbf{explicitly defines the task} as localizing anatomical or pathological regions across diverse medical imaging modalities, and makes clear to the model that natural language questions may lack explicit spatial references—requiring inference based on clinical context and visual cues. To address this, the prompt enforces \textbf{a structured reasoning process} resembling clinical diagnostic logic: generating a hypothesis based on medical context, systematically inspecting visual features, and iteratively refining the inference. This process culminates in visual grounding, achieved by aligning domain knowledge with observable cues such as shape, density, texture, and structural variation. The \textbf{required output} consists of a <think> block articulating unambiguous clinical reasoning, followed by an <answer> block containing a precise bounding box and two interior key points. This design promotes accuracy, reproducibility, and interpretability, while explicitly restricting diagnostic speculation and minimizing ambiguity in localization. In all experiments presented in this work, any response explicitly labeled as \textit{reason} was generated using this prompt.

\paragraph{Direct User Prompt.}
This prompt specifies a direct localization task in medical images, requiring spatial outputs \textbf{without an explicit reasoning trace}. The model receives a natural language question and must respond solely with an <answer> block containing a structured json object: a tight bounding box enclosing the target region and two distinct key points. Unlike reasoning-based prompts, this version provides no guidance on interpreting ambiguous queries or incorporating clinical context. Any response explicitly labeled as \textit{without reason} was generated using this prompt.

\begin{figure*}[t]
    \centering
    \includegraphics[width=1\linewidth]{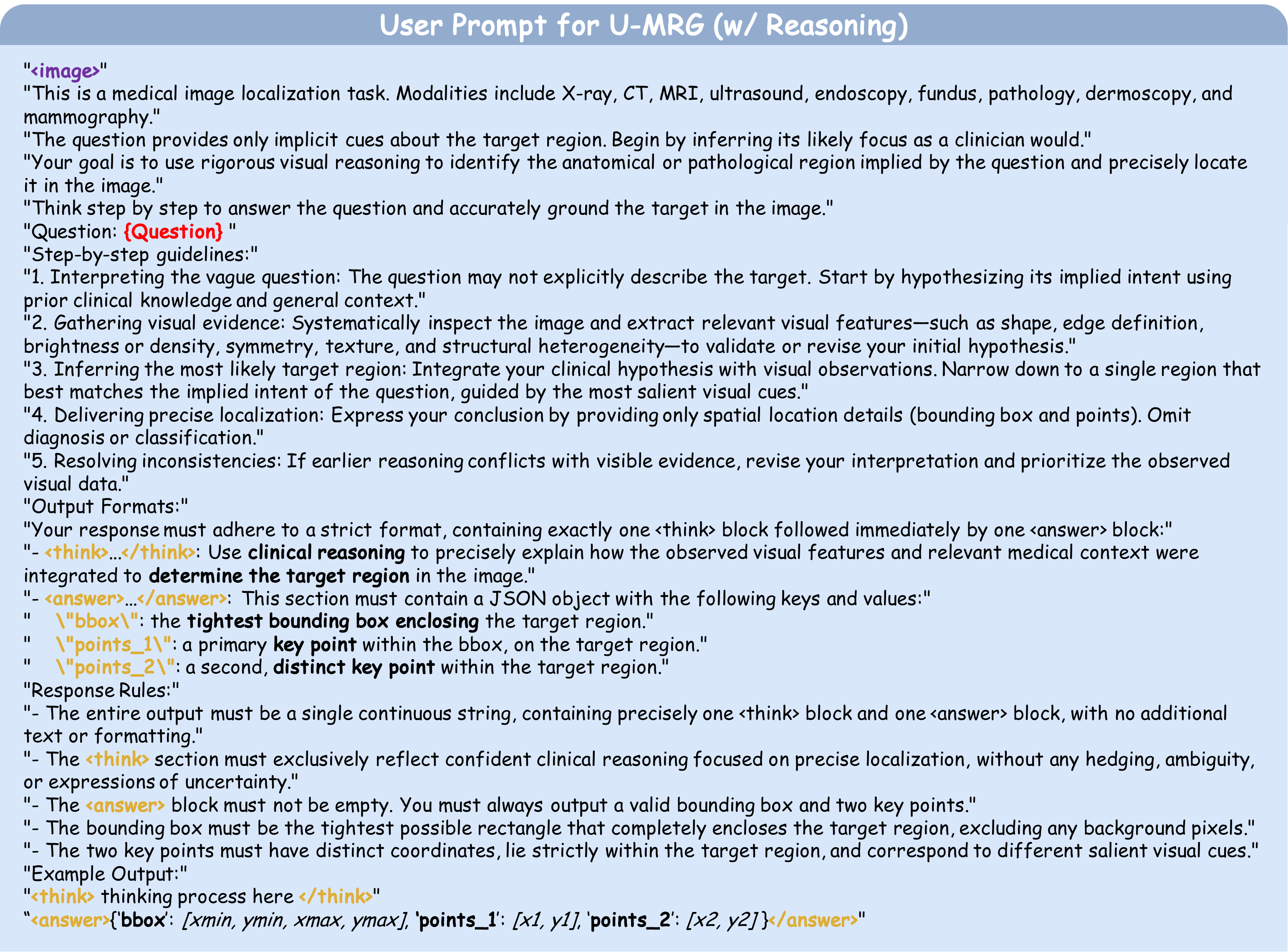}
    \caption{\textbf{Reasoning User Prompt.} Prompt variant used in RL training and \textit{with-reasoning} evaluation. The model must (1) generate a \texttt{<think>} block that walks through step-by-step visual reasoning, then (2) output an \texttt{<answer>} block containing a JSON object with a tight bounding box and two interior key points.}
    \label{fig:reason_prompt}
\end{figure*}

\begin{figure*}[t]
    \centering
    \includegraphics[width=1\linewidth]{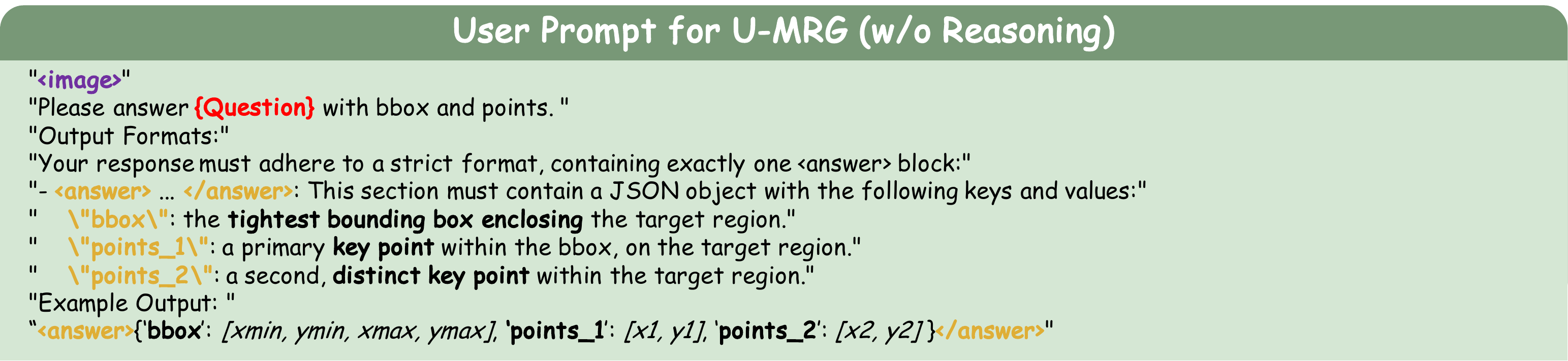}
    \caption{\textbf{Direct User Prompt.} Prompt variant for the \textit{without-reasoning} baseline. The model skips the explicit reasoning trace and returns only the \texttt{<answer>} block with a JSON object with a tight bounding box and two interior key points.}
    \label{fig:direct_prompt}
\end{figure*}

\subsection{Reward Functions} \label{apd:reward_function}

\paragraph{BBox IoU Reward.}
The \textbf{IoU reward} quantifies the spatial overlap between a predicted bounding box $\mathcal{B}_p$ and a ground truth bounding box $\mathcal{B}_g$. It is defined as the ratio of their intersection area to their union area:
\begin{equation}
\mathbb{R}_{\mathrm{iou}} = \frac{\mathrm{Area}(\mathcal{B}_p \cap \mathcal{B}_g)}{\mathrm{Area}(\mathcal{B}_p \cup \mathcal{B}_g)}.
\end{equation}
This metric ranges from 0 (no overlap) to 1 (perfect overlap), with higher values indicating better localization accuracy. 

\paragraph{BBox Alignment Reward.}
The \textbf{Alignment reward} measures the average L1 distance between corresponding corner coordinates of the predicted bounding box $\mathcal{B}_p$ and the ground truth bounding box $\mathcal{B}_g$. This distance is then normalized by the diagonal length of $\mathcal{B}_g$ to ensure scale invariance. Formally, it's expressed as:
\begin{equation}
\mathbb{R}_{\mathrm{align}} = \frac{1}{4} \sum_{i=1}^{4} \left| \mathcal{B}_p^{(i)} - \mathcal{B}_g^{(i)} \right|.
\end{equation}
A lower $\mathbb{R}_{\mathrm{align}}$ value signifies superior positional alignment, making it a direct indicator of how closely the predicted box's corners match those of the ground truth.

\paragraph{BBox Scale Reward.}
The \textbf{Scale reward} measures structural consistency between a predicted box and its ground truth, considering both area and aspect ratio. It computes the Euclidean distance between the logarithmic differences of the box areas and aspect ratios:
\begin{equation}
\mathbb{R}_{\mathrm{scale}} = \sqrt{(\Delta \log A)^2 + (\Delta \log R)^2},
\end{equation}
where $\Delta \log A$ and $\Delta \log R$ denote the logarithmic differences in box area and aspect ratio, respectively. A smaller $\mathbb{R}_{\mathrm{scale}}$ value indicates superior structural alignment, reflecting a better match in shape and proportionality.

\paragraph{Points Dice (pDice) Reward.}
The \textbf{Points Dice (pDice) reward} evaluates the spatial correspondence between a predicted keypoint pair $\mathcal{P}_p = \{\mathbf{p}_1^p, \mathbf{p}_2^p\}$ and a ground-truth keypoint pair $\mathcal{P}_g = \{\mathbf{p}_1^g, \mathbf{p}_2^g\}$. This reward models each point pair as the diameter of a circle, $O_p$ and $O_g$ respectively. The Dice score is then computed to quantify the spatial overlap between these circles:
\begin{equation}
\mathbb{R}_{\mathrm{pdice}} = \frac{2 \cdot \mathrm{Area}(O_p \cap O_g)}{\mathrm{Area}(O_p) + \mathrm{Area}(O_g)}.
\end{equation}
A higher $\mathbb{R}_{\mathrm{pdice}}$ value indicates better spatial alignment and consistency between the regions defined by the predicted and ground-truth keypoint pairs.

\paragraph{Points Alignment Reward.}
The \textbf{Alignment reward} quantifies the positional accuracy of predicted keypoints by computing the mean absolute error between corresponding points in a predicted pair $\mathcal{P}_p = \{\mathbf{p}_1^p, \mathbf{p}_2^p\}$ and a ground truth pair $\mathcal{P}_g = \{\mathbf{p}_1^g, \mathbf{p}_2^g\}$. It is formulated as:
\begin{equation}
\mathbb{R}_{\mathrm{align}} = \frac{1}{2} \sum_{i=1}^{2} \left( |x_i^p - x_i^g| + |y_i^p - y_i^g| \right).
\end{equation}
A lower $\mathbb{R}_{\mathrm{align}}$ value signifies superior positional alignment, indicating that the predicted keypoints are precisely located relative to their ground truth counterparts.

\paragraph{Points Angle Reward.}
The \textbf{Angle reward} quantifies the angular consistency between a predicted keypoint pair and its ground-truth counterpart. It computes the absolute cosine similarity between their respective direction vectors, $\mathbf{v}_p = \mathbf{p}_2^p - \mathbf{p}_1^p$ and $\mathbf{v}_g = \mathbf{p}_2^g - \mathbf{p}_1^g$:
\begin{equation}
\mathbb{R}_{\mathrm{angle}} = \left| \cos\left( \theta \right) \right| = \left| \frac{\langle \mathbf{v}_p, \mathbf{v}_g \rangle}{\| \mathbf{v}_p \|_2 \cdot \| \mathbf{v}_g \|_2} \right|,
\end{equation}
where $\theta$ is the angle between the vectors. A higher $\mathbb{R}_{\mathrm{angle}}$ value (closer to 1) indicates superior angular alignment, signifying that the orientation defined by the predicted points closely matches that of the ground-truth.

\subsection{Smoothing and Penalization} \label{apd:smoothing_penalization}
\paragraph{Smoothing.}
To enhance training stability and provide a more nuanced differentiation of prediction quality, all reward components are processed by dedicated smoothing functions. These transformations convert raw reward values or distances into a smoothed range, making them more suitable for reinforcement learning optimization.

\begin{figure}[t]
    \centering
    \includegraphics[width=0.7\linewidth]{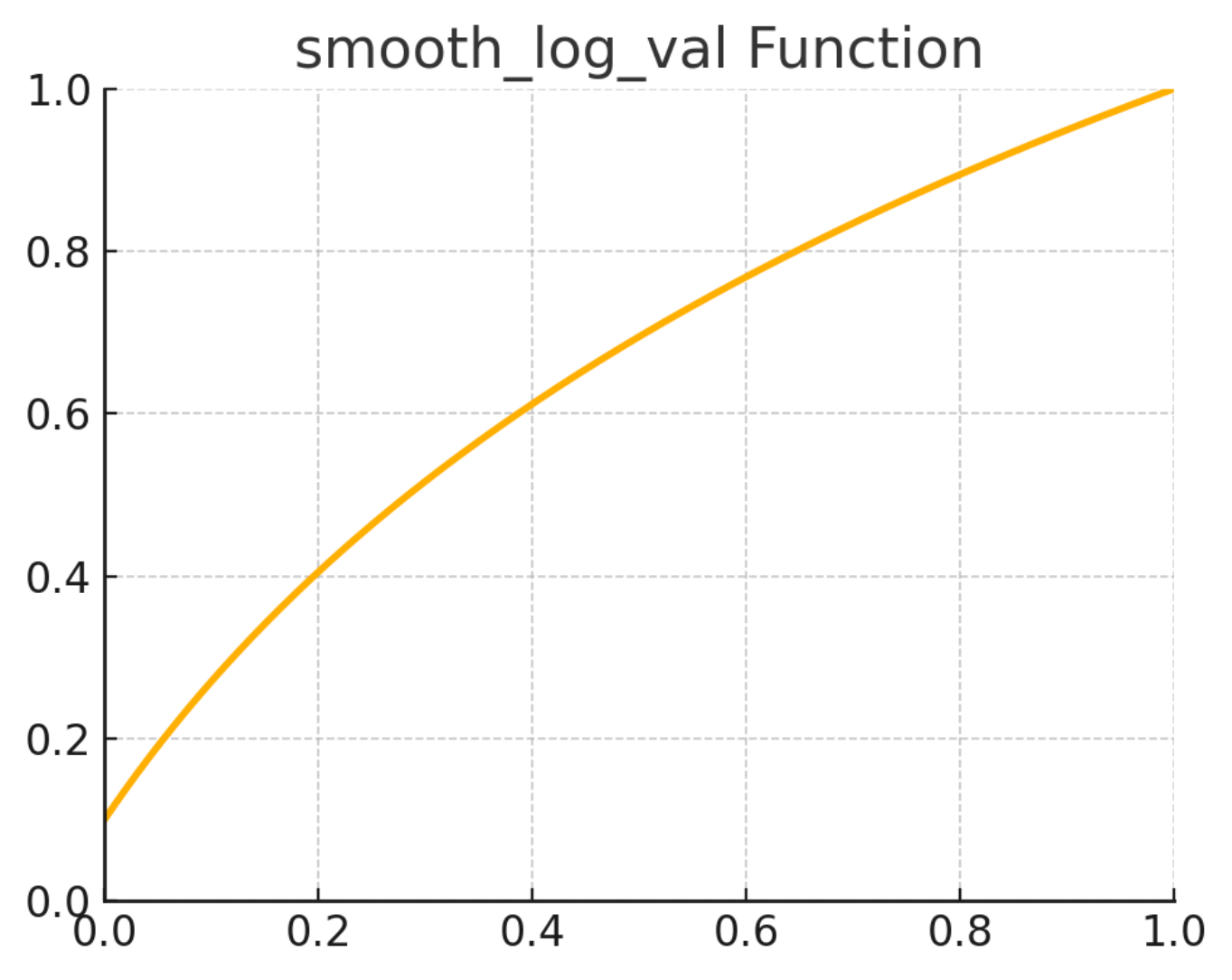}
    \caption{Visualization of the logarithmic smoothing function, showing the smoothing reward growth from $0.1$ to $1.0$ over the input range $[0,1]$.}
    \label{fig:smoothing_log}
\end{figure}

For rewards where a higher value indicates better performance (e.g., $\mathbb{R}_{\mathrm{iou}}$, $\mathbb{R}_{\mathrm{pdice}}$, and $\mathbb{R}_{\mathrm{angle}}$), \textbf{logarithmic smoothing} is applied. This function, given by:
\begin{equation}
\mathcal{S}_{\log}(r; k) = \frac{\log(k r + 1)}{\log(k + 1)},
\end{equation}
maps the raw reward $r \in [0, 1]$ to a smoothed value. The parameter $k$ (default $k = 3$) acts as a \textbf{scaling factor}, controlling the curvature of the logarithmic function. This smoothing compresses higher reward values while expanding the differences among lower reward values, providing stronger gradients for initial improvements when the raw reward is small. This behavior is illustrated in Fig.~\ref{fig:smoothing_log}.

\begin{figure}[t]
    \centering
    \includegraphics[width=0.7\linewidth]{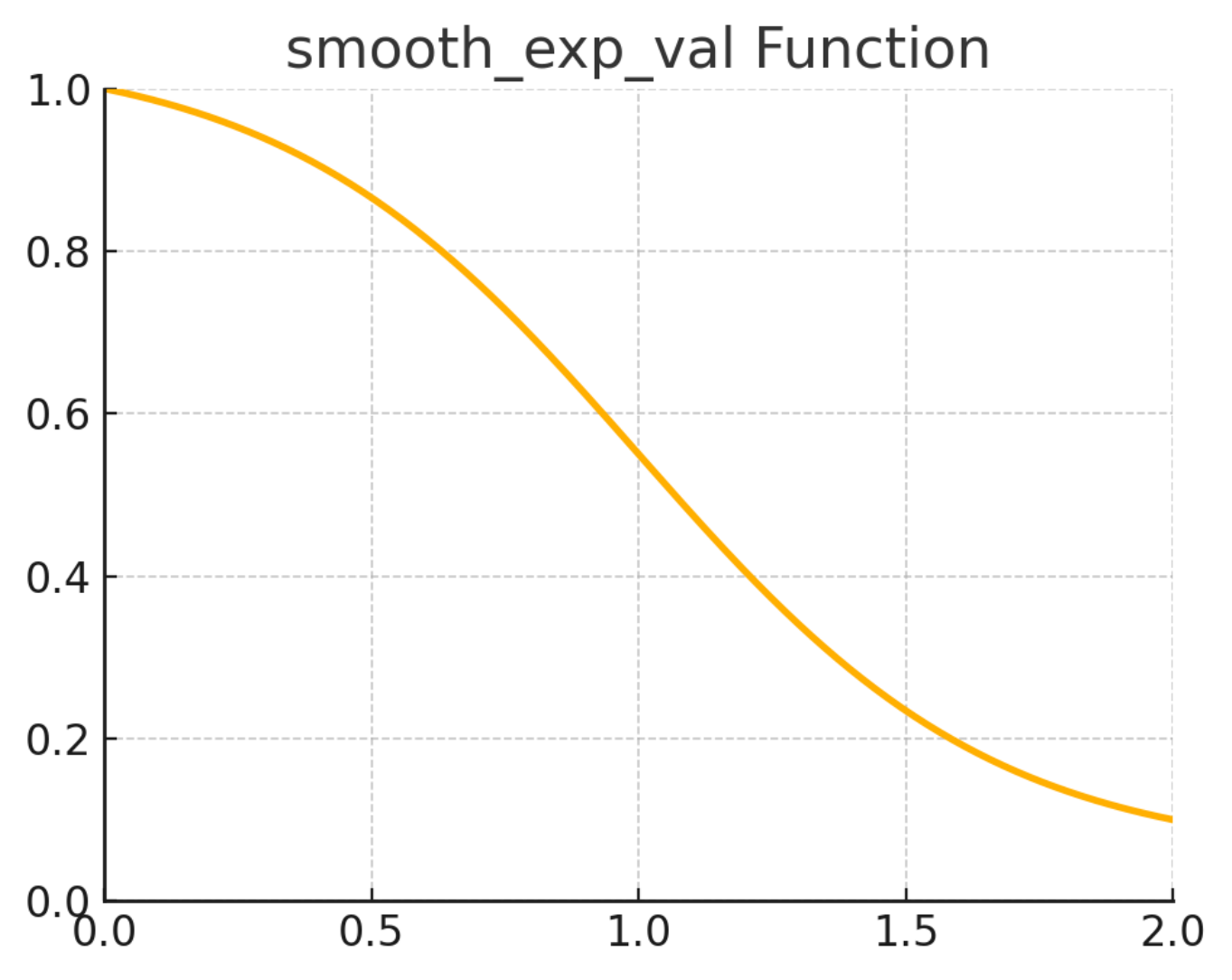}
    \caption{Visualization of the exponential smoothing function, illustrating a smooth exponential decay from $1.0$ to $0.1$ over the input range $[0,2]$, centered around $x=1.0$.}
    \label{fig:smoothing_exp}
\end{figure}

For rewards where a lower value indicates better performance (e.g., $\mathbb{R}_{\mathrm{align}}$ and $\mathbb{R}_{\mathrm{scale}}$), \textbf{exponential smoothing} is utilized. This function, defined as:
\begin{equation}
\mathcal{S}_{\exp}(d; k, c) = \frac{1}{1 + e^{k(d - c)}},
\end{equation}
transforms the normalized distance $d \in [0, 2]$ into a smoothed reward. Here, $k$ (default $k = 3$) controls the \textbf{steepness} of the sigmoid-like curve, and $c$ (default $c = 1$) represents the \textbf{target center}, shifting the inflection point of the curve. This smoothing assigns higher rewards when the distance $d$ is small and rapidly decreases the reward as $d$ increases, effectively penalizing larger deviations more aggressively. This behavior is visualized in Fig.~\ref{fig:smoothing_exp}.

\paragraph{Penalization.}
After applying smoothing functions, we further refine the reward signal through a penalization function $\mathcal{N}(\cdot)$. The primary purpose of this penalization is to softly down-weight unreliable predictions, ensuring that the learning agent is not overly rewarded for outputs that are spatially implausible or inconsistent.
For each smoothed reward, two validity scores, $v_1$ and $v_2$, are computed. These scores are designed to reflect the spatial plausibility of the predicted output based on different criteria.

For \textbf{BBox-based rewards} ($\mathbb{R}_{\mathrm{iou}}$, $\mathbb{R}_{\mathrm{align}}$, $\mathbb{R}_{\mathrm{scale}}$): 
(1) One validity score checks if the ground truth points are largely contained within the predicted bounding box. This ensures the predicted box correctly encloses the target.
(2) Another score assesses the plausibility of the predicted bounding box's area relative to the ground truth. For instance, if a predicted box is excessively large or small compared to the ground truth, it would receive a lower validity score.

For \textbf{Points-based rewards} ($\mathbb{R}_{\mathrm{pdice}}$, $\mathbb{R}_{\mathrm{align}}$, $\mathbb{R}_{\mathrm{angle}}$):
(1) One validity score ensures the predicted keypoints are within a plausible spatial range, such as within the ground truth box or image boundaries.
(2) Another score evaluates the spatial spread or separation of the predicted keypoints. If the points are too close together (e.g., collapsing to a single point) or too far apart, it suggests an unreliable prediction.

The final reward is adjusted using the following formula:
\begin{equation}
\mathcal{N}(r; v_1, v_2) = \lambda r + (1 - \lambda) r \cdot \frac{v_1 + v_2}{2},
\end{equation}
where $r$ is the smoothed reward, $v_1$ and $v_2$ are the two validity scores (typically ranging from 0 to 1, with 1 indicating high validity and 0.5 indicating moderate validity in the provided context), and $\lambda$ is a mixing coefficient (default $\lambda = 0.7$). This formula linearly combines the smoothed reward with a weighted average of the smoothed reward scaled by the validity scores. The parameter $\lambda$ controls the influence of the raw smoothed reward versus the validity-adjusted reward. A higher $\lambda$ places more emphasis on the smoothed reward, while a lower $\lambda$ allows the validity scores to more significantly penalize unreliable predictions. This penalization mechanism acts as a soft constraint, discouraging the model from making outputs that, despite potentially having a reasonable IoU or alignment, are fundamentally illogical in their spatial configuration.

\section{More Details on Experiments}

\subsection{Implementation Details} \label{apd:implementation_details}
We adopt Lingshu-7B~\cite{xu2025lingshu} and \textbf{}MedSAM2~\cite{ma2025medsam2} as our default CRM and ASM, respectively. MedReasoner is trained on an 8 NVIDIA A100-80G GPUs with the veRL~\cite{sheng2025hybridflow}. Training utilized a total batch size of 5 and 16 samples per step. The initial learning rate is set to 1e-6. In terms of generation configuration, we follow the same settings as used in Lingshu, setting the repetition penalty to 1.05 and the temperature to 0.1, while employing argmax sampling.

\subsection{Model Details}

\paragraph{GPT-4o~\cite{gpt4o}}GPT-4o is a multimodal, decoder-only closed-source language model capable of processing text, vision, and audio within a unified architecture. It is trained end-to-end on mixed-modality data and optimized for reasoning, instruction following, and real-time interaction. GPT-4o achieves strong performance across many standard academic benchmarks.

\paragraph{Gemini 2.5 Flash~\cite{gemini2.5_flash}}
Gemini 2.5 Flash is a lightweight text-to-text decoder model derived from Google's Gemini series, optimized for efficiency and low-latency applications. Despite its compact design, it supports multimodal inputs (text, code, image, audio, video) and features a 1M-token context window. The model excels in reasoning and language understanding benchmarks.

\paragraph{Qwen2.5-VL~\cite{Qwen2.5-VL}}
Qwen2.5-VL extends the Qwen2.5 language model with advanced multimodal capabilities, integrating a vision encoder enhanced by dynamic resolution training, window attention, SwiGLU, and RMSNorm. It achieves robust visual reasoning performance across charts, texts, and layouts.

\paragraph{InternVL3~\cite{chen2024internvl}}
InternVL3 is a multimodal decoder-based model built upon a Qwen2.5-derived language backbone and a newly pre-trained vision encoder, following the ViT-MLP-LLM architecture. It adopts Native Multimodal Pre-Training on text, image, and video data, enabling strong long-context understanding and tool-use reasoning across domains such as 3D vision and GUI agents.

\paragraph{Med-R1~\cite{lai2025med}}
Med-R1 is a reinforcement learning-enhanced vision-language model tailored for medical tasks across eight imaging modalities. It employs Group Relative Policy Optimization (GRPO) to improve generalization in tasks such as disease diagnosis and lesion grading, surpassing Qwen2-VL-72B by a significant margin.

\paragraph{MiniInternVL~\cite{gao2024mini}}
MiniInternVL, part of InternVL2.0, is a compact multimodal instruction-tuned model built with InternViT-300M, an MLP projector, and Phi-3-mini-128k. It supports high-resolution images and long-video inputs, demonstrating efficient performance in diverse vision-language tasks.

\paragraph{MedGemma~\cite{sellergren2025medgemma}}
MedGemma is based on Gemma 3 and integrates a MedSigLIP encoder pre-trained on de-identified medical images, with language components trained on medical text. Optimized for instruction tuning, the model supports medical applications such as report summarization and diagnosis explanation.

\paragraph{HuatuoGPT~\cite{chen2024huatuogpt}}
HuatuoGPT is a Chinese medical instruction-tuned LLM based on LLaMA, designed for diagnostic consultations. It is fine-tuned using synthetic instructions from ChatGPT and real doctor-patient dialogues. The model demonstrates strong performance in symptom interpretation and treatment recommendation.

\paragraph{Lingshu~\cite{xu2025lingshu}}
Lingshu is a medical MLLM built on Qwen2.5-VL, integrating a vision encoder, LLM, and projection module. Trained via a multi-stage process with RL using verifiable rewards, it leverages over 5 million multimodal and textual medical samples. It supports unified understanding across multiple imaging types

\paragraph{Chiron-o1~\cite{sun2025enhancing}}
Chiron-o1 is an MLLM fine-tuned from InternVL using a collaborative search architecture involving mentor-trainee feedback loops to enhance reasoning. It excels in benchmark reasoning tasks by generating interpretable reasoning paths.

\paragraph{MedSAM~\cite{ma2024segment}}
Built upon the Segment Anything Model (SAM), MedSAM is fine-tuned on a large-scale dataset of 1.57 million image-mask pairs, spanning 10 medical imaging modalities and over 30 cancer types. MedSAM adopts a promptable segmentation approach using bounding boxes to flexibly specify regions of interest, which enables clinicians to guide segmentation according to varied needs.

\paragraph{SAM-Med2D~\cite{cheng2023sammed2d}}
SAM-Med2D bridges this domain gap by fine-tuning SAM on a large-scale medical image dataset comprising approximately 4.6 million images and 19.7 million masks. Unlike earlier adaptations which focused on limited prompt types or small datasets, SAM-Med2D supports a full spectrum of prompts—points, bounding boxes, and masks—enabling robust interactive segmentation across complex clinical scenarios.

\paragraph{MedSAM2~\cite{ma2025medsam2}}
MedSAM2 is a generalized auto-tracking model designed for universal 2D and 3D medical image segmentation. Built upon SAM2, MedSAM2 treats medical image segmentation as a video object tracking task to unify the processing of unordered 2D slices and volumetric 3D scans. A key innovation is its self-sorting memory bank, which dynamically selects and resamples embeddings. 

\paragraph{SAM4MLLM~\cite{chen2024sam4mllm}}
SAM4MLLM integrates the Segment Anything Model (SAM) with MLLMs to perform Referring Expression Segmentation using a purely text-based training objective. It introduces Prompt-Point Generation and Proactive Query mechanisms to generate and refine segmentation prompts. Notably, the model retains the original MLLM architecture, enabling pixel-level localization without structural changes.

\paragraph{VLM-R1~\cite{shen2025vlm}}
VLM-R1 adapts rule-based reinforcement learning from DeepSeek-R1 for vision tasks like REC and OVD, using a novel odLength metric to avoid reward hacking. Built on Qwen2.5VL-3B, it enables complex visual reasoning behaviors such as emergent "aha" moments. The framework enhances generalizability via task-specific visual feedback signals.

\paragraph{SegZero~\cite{liu2025seg}}
SegZero combines Qwen2.5-VL and SAM2 within a reinforcement learning framework for zero-shot reasoning segmentation. It decouples reasoning and segmentation by using CoT prompts and a frozen segmentation module. Trained without annotated reasoning data, it achieves SOTA results on ReasonSeg.

\subsection{Out-of-Distribution Experiments}

To evaluate generalization, we created a biased training set, \textbf{U-MRG-6K}, using 6K images from the 5 most frequent categories in U-MRG-14K. The remaining images from the other 9 categories served as the out-of-distribution (OOD) test set. We then trained the CRM on U-MRG-6K using (1) SFT and (2) three RL variants. We evaluated their performance under different prompting strategies.

Table~\ref{tab:ablation_ood} shows that SFT suffers from severe phrase overfitting: without explicit reasoning, it refused 91\% of OOD queries, and even with reasoning, its accuracy dropped below the base model. In contrast, all RL variants outperformed the base and SFT models across all metrics, confirming the benefit of reinforcement learning for implicit-query grounding when the test distribution shifts. Furthermore, both RL variants (excluding the base) were robust to prompt variations in OOD situation, maintaining grounding performance even without detailed task instructions, whereas SFT heavily relied on reasoning-path prompts to avoid refusals.

\begin{table}[t]
\centering
\begin{tabular*}{\linewidth}{@{\extracolsep{\fill}}lrrrrr}
\toprule
Method & Reason & IoU$\uparrow$ & pDice$\uparrow$ & Dice$\uparrow$ & \# Ref.$\downarrow$ \\
\midrule
SFT & \ding{55} & 0.32 & 0.08 & 0.55 & 1081 \\
SFT & \ding{51} & 6.32 & 1.56 & 13.45 & 0 \\
\midrule
RL-Base & \ding{55} & 8.11 & 1.93 & 14.72 & 1 \\
RL-Base & \ding{51} & 16.09 & 9.78 & 24.91 & 0 \\
RL-Hard & \ding{55} & 16.08 & 10.85 & 23.28 & 1 \\
RL-Hard & \ding{51} & 16.92 & \underline{11.85} & 24.34 & 0 \\
RL-Soft & \ding{55} & \underline{17.27} & \textbf{12.02} & \underline{25.2} & 0 \\
RL-Soft & \ding{51} & \textbf{17.67} & 11.33 & \textbf{26.72} & 0 \\
\bottomrule
\end{tabular*}
\caption{Performance on the OOD split with Lingshu-7B as the CRM. Models are trained on \textbf{U-MRG-6K} (five frequent categories) and tested on unseen classes.}
\label{tab:ablation_ood}
\end{table}

\begin{figure*}[t]
    \centering

    \begin{subfigure}[b]{\textwidth}
        \centering
        \includegraphics[width=\linewidth]{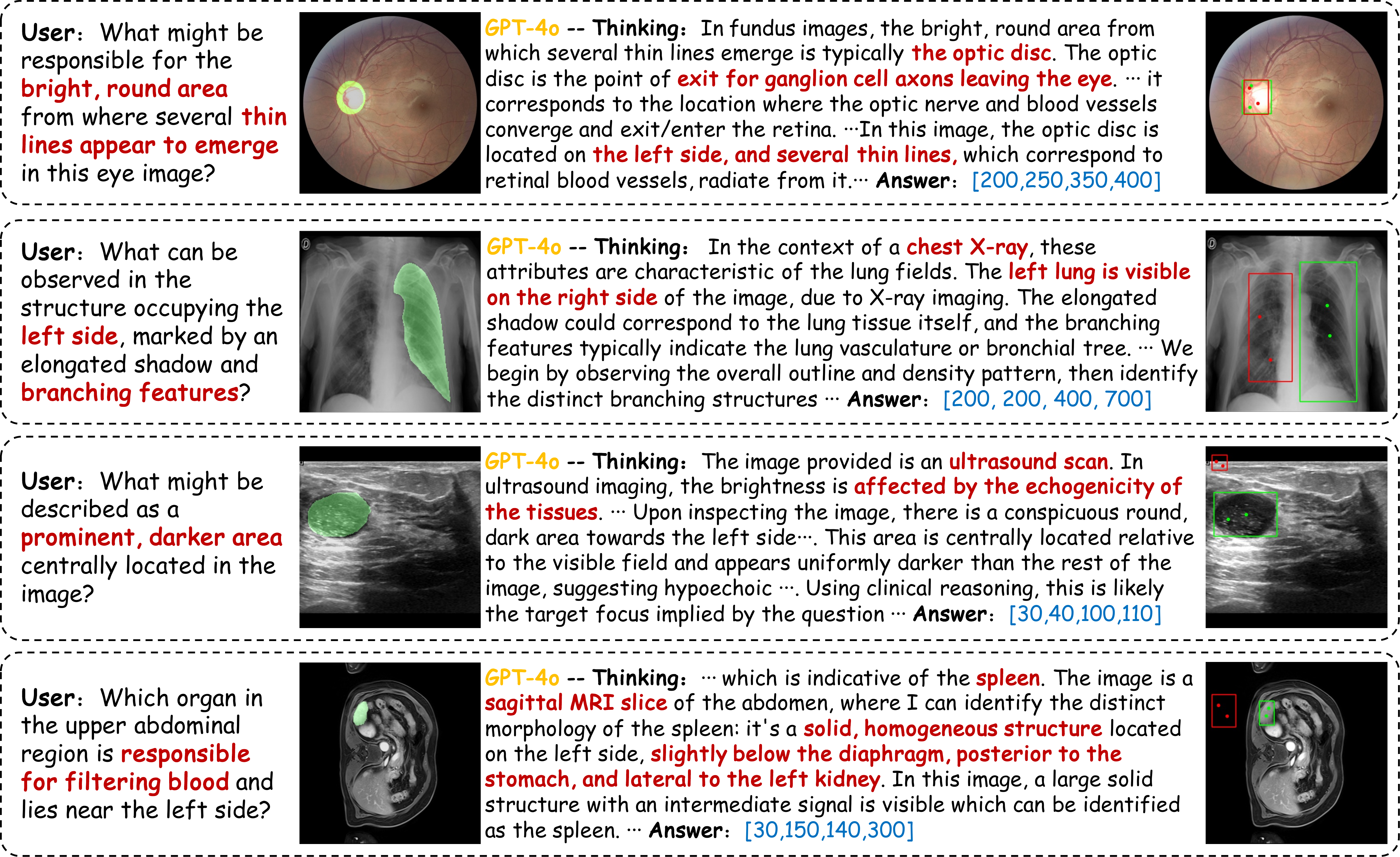}
        \label{fig:qualitative1}
    \end{subfigure}

    \begin{subfigure}[b]{\textwidth}
        \centering
        \includegraphics[width=\linewidth]{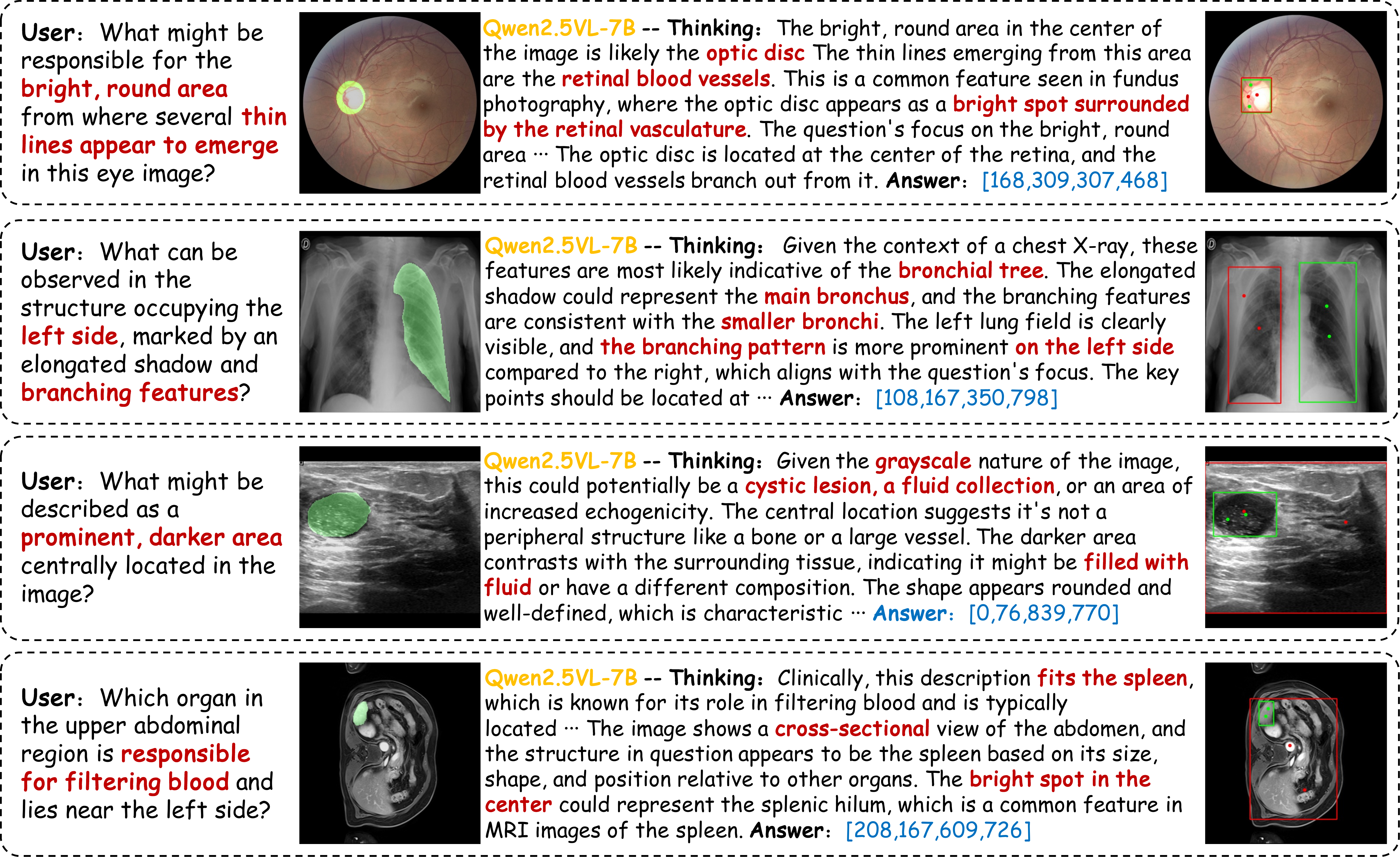}
        \label{fig:qualitative2}
    \end{subfigure}

    \caption{The top four rows show GPT-4o’s responses in different cases, while the bottom four rows present those of Qwen2.5VL-7B. For each case, the columns from left to right represent the ambiguous query, the image input with the ground-truth mask, the model’s reasoning process, and the model’s output including the predicted bounding box and keypoints.}

    \label{fig:qualitative_1_to_2}
\end{figure*}

\begin{figure*}[t]
    \centering

    \begin{subfigure}[b]{\textwidth}
        \centering
        \includegraphics[width=\linewidth]{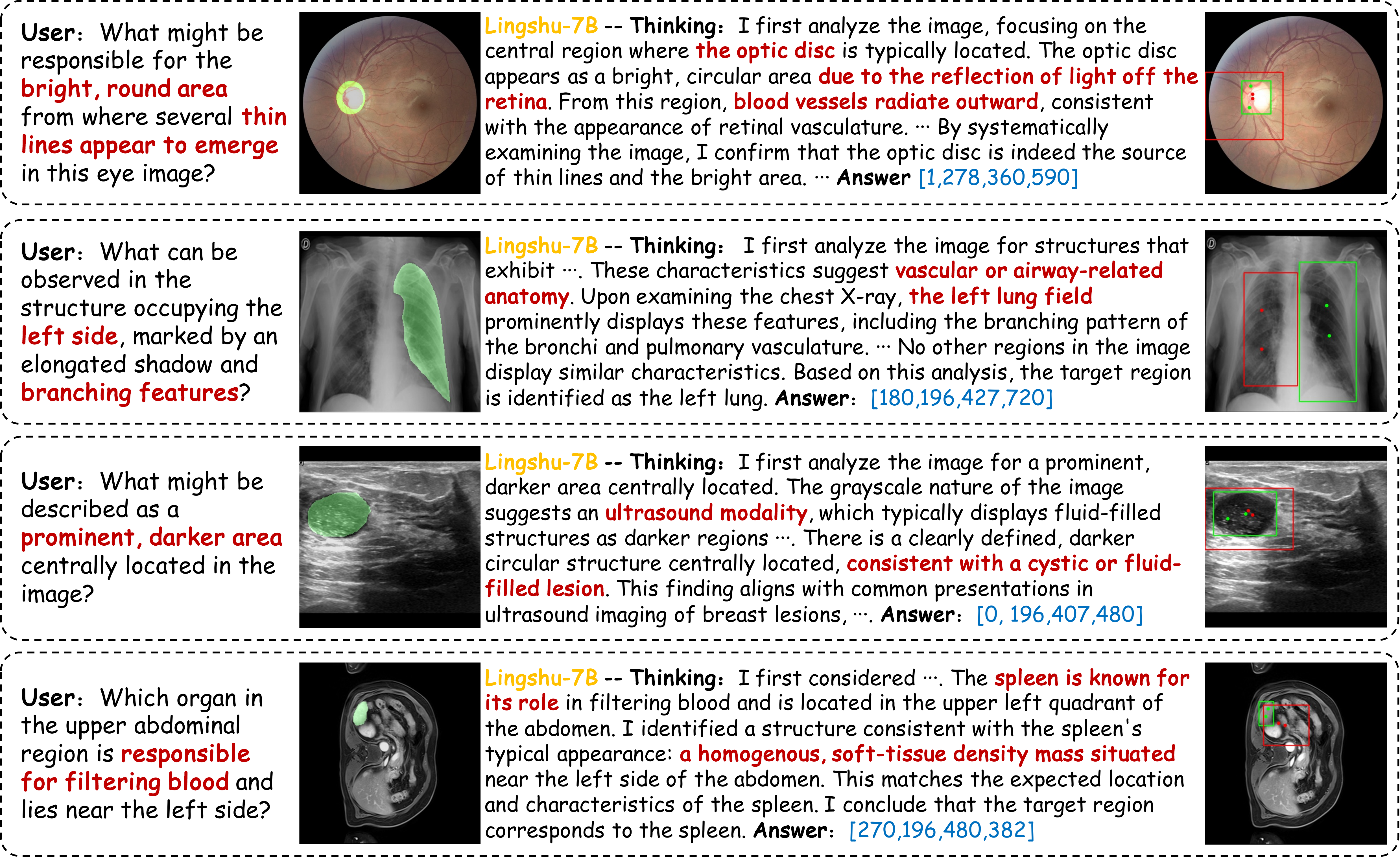}
        \label{fig:qualitative3}
    \end{subfigure}

    \begin{subfigure}[b]{\textwidth}
        \centering
        \includegraphics[width=\linewidth]{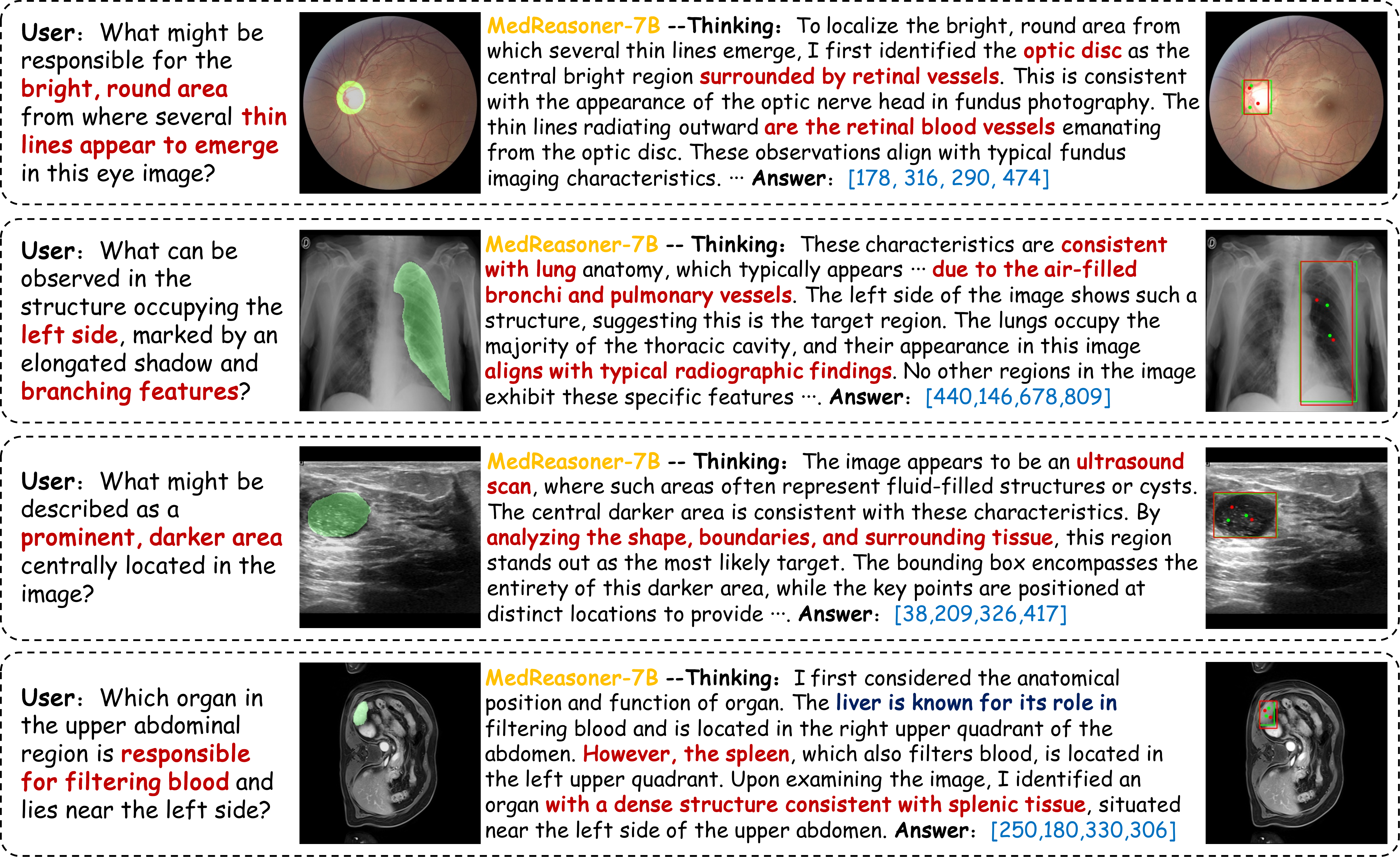}
        \label{fig:qualitative4}
    \end{subfigure}

    \caption{The top four rows show Lingshu-7B’s responses in different cases, while the bottom four rows present those of ours. For each case, the columns from left to right represent the ambiguous query, the image input with the ground-truth mask, the model’s reasoning process, and the model’s output including the predicted bounding box and keypoints.}

    \label{fig:qualitative_3_to_4}
\end{figure*}

\subsection{Qualitative Results} \label{apd:qualitative_results}
Based on our observations of the output generated by various models, we found that GPT-4o, despite being a closed-source commercial model capable of \textbf{producing comprehensive reasoning paths} and generally correct interpretations of ambiguous questions, fails to translate this understanding into accurate grounding. This shortcoming reveals \textbf{a lack of pixel-level comprehension} of medical images, as evidenced by its tendency to produce bounding boxes with coordinates rounded to multiples of 10.

Qwen2.5VL, a powerful open-source general-purpose MLLM, demonstrates better alignment between textual and visual content and is \textbf{able to produce relatively precise grounding} results. However, its limited medical knowledge hinders its ability to \textbf{infer correct targets from ambiguous queries}, and it lacks domain-specific understanding of uncommon medical imaging modalities such as histopathological slides and fundus images.

Lingshu, a model further trained on medical tasks, is capable of correctly reasoning and identifying targets in ambiguous contexts. Nonetheless, it remains inadequate in \textbf{translating natural language-based grounding into bounding boxes or point coordinates} that can be effectively utilized by segmentation models.

In contrast, our model, trained with a GRPO objective that \textbf{incorporates a pixel-level grounding reward}, successfully aligns natural language reasoning with pixel-level grounding. It achieves accurate interpretation of ambiguous referents and bridges the gap left by existing models.

\section{Broaden Impact}

\subsection{Societal Impact}
Our dataset is anticipated to have a significant positive impact on both medical research and clinical practice. By providing a high-quality, publicly available benchmark, it can accelerate the development and validation of novel image segmentation algorithms, thereby \textbf{pushing forward the broader scientific research process}. Models trained on this dataset can \textbf{effectively assist physicians in diagnosis} by enabling faster and more accurate delineation of pathological regions, which is crucial for treatment planning and disease monitoring. However, it is crucial to acknowledge potential risks. \textbf{Over-reliance on AI-driven tools without proper clinical oversight could lead to serious diagnostic errors}. Therefore, we advocate that these models should be used as assistive tools to augment, not replace, clinical expertise. The use of this dataset and any derived models for direct clinical decision-making without rigorous, independent validation and regulatory approval is strongly discouraged.

\subsection{Ethics Statement}
We have ensured strict adherence to all applicable ethical guidelines. Our dataset is compiled exclusively from \textbf{publicly available datasets where the original providers have explicitly affirmed that patients provided informed consent} for their data to be used in research. All data has been rigorously and consistently anonymized to remove any personal patient information and protect patient privacy before inclusion in our collection. This dataset is intended \textbf{solely for academic and non-commercial purposes. Any unauthorized commercial use is strictly prohibited.} We urge all users to respect this provision to maintain the integrity and ethical standing of this valuable resource.

\section{Case Study}

\subsection{Meta Information of U-MRG-14K} \label{case:meta_infor}
\begin{figure*}[t]
    \centering

    \begin{subfigure}[b]{\textwidth}
        \centering
        \includegraphics[width=\linewidth]{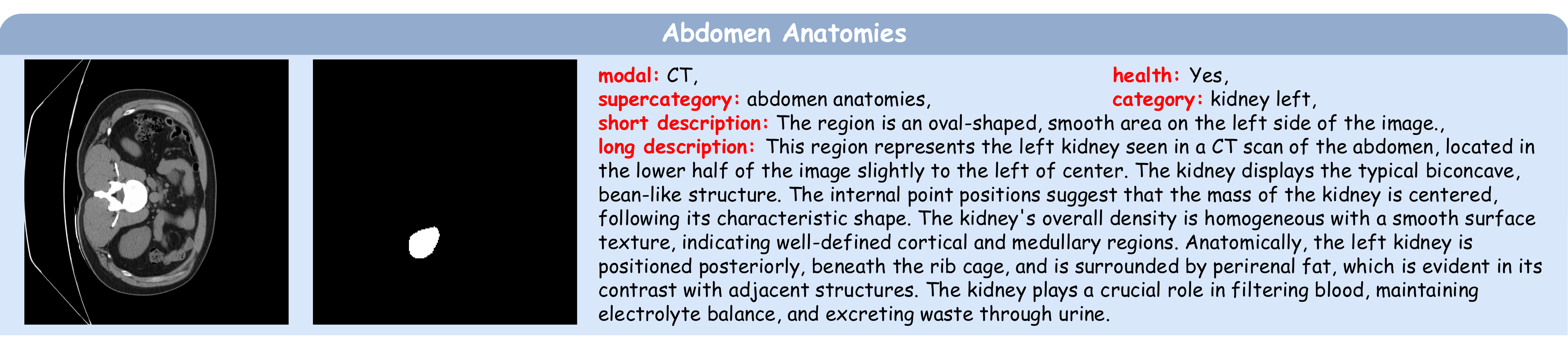}
        \label{fig:meta_infor_organ1}
    \end{subfigure}
    
    \begin{subfigure}[b]{\textwidth}
        \centering
        \includegraphics[width=\linewidth]{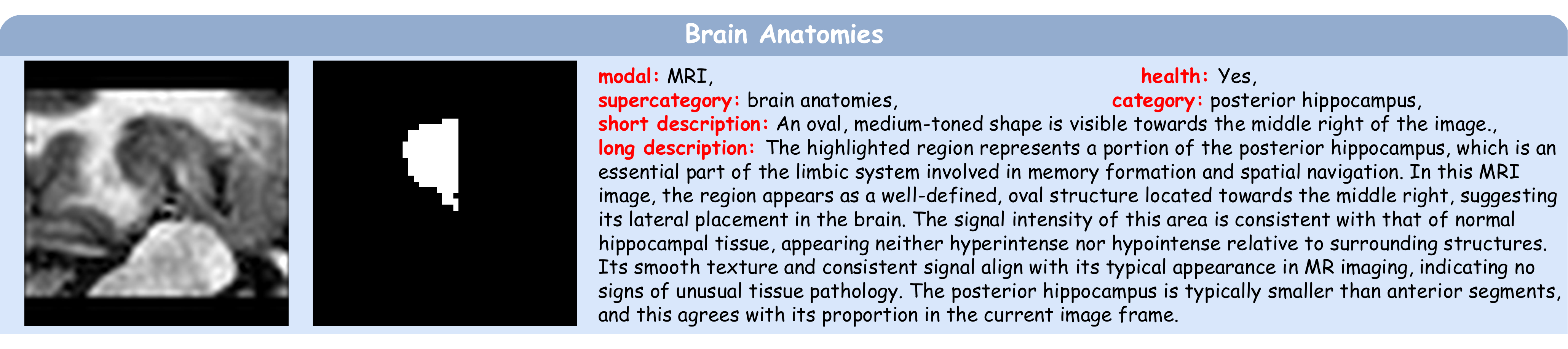}
        \label{fig:meta_infor_organ2}
    \end{subfigure}
    
    \begin{subfigure}[b]{\textwidth}
        \centering
        \includegraphics[width=\linewidth]{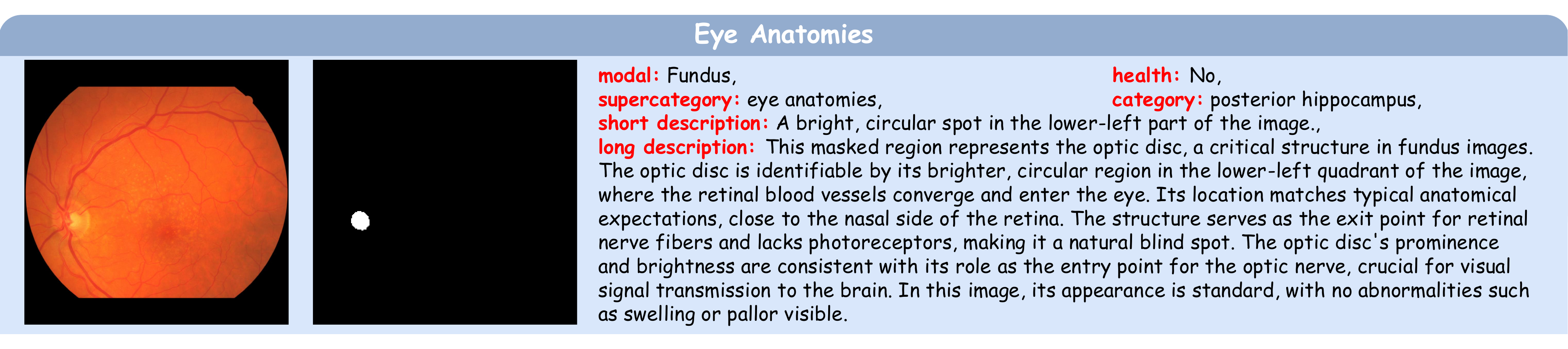}
        \label{fig:meta_infor_organ3}
    \end{subfigure}
    
    \caption{Examples of the meta information from three anatomical super-categories. From top to bottom: \textbf{Abdomen Anatomies}, \textbf{Brain Anatomies}, \textbf{Eye Anatomies}. Each case presents the image along with its: \textit{modal}, \textit{health}, \textit{super-category}, \textit{category}, \textit{short description} and \textit{long description}.}

    \label{fig:meta_infor_1_to_3}
\end{figure*}

\begin{figure*}[t]
    \centering

    \begin{subfigure}[b]{\textwidth}
        \centering
        \includegraphics[width=\linewidth]{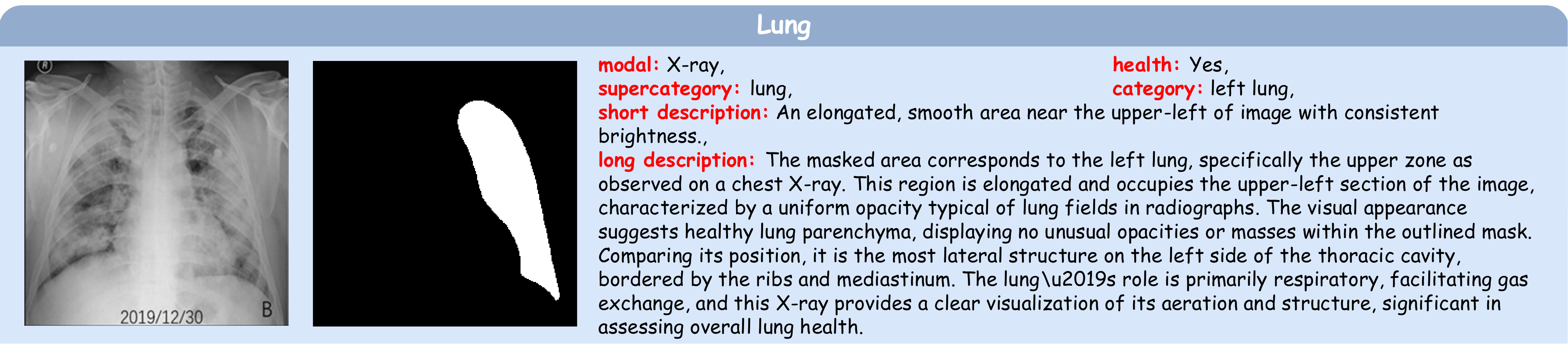}
        \label{fig:meta_infor_organ4}
    \end{subfigure}

    \begin{subfigure}[b]{\textwidth}
        \centering
        \includegraphics[width=\linewidth]{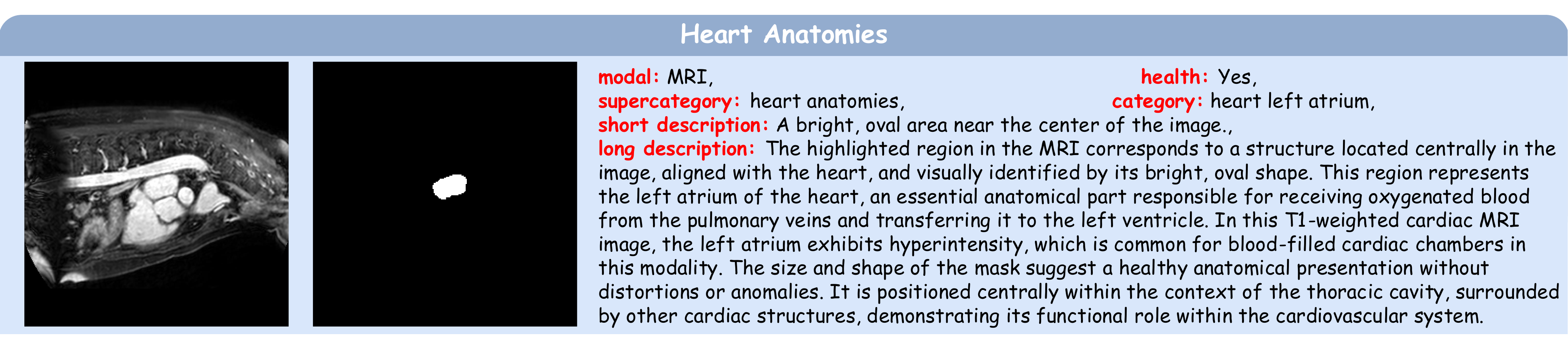}
        \label{fig:meta_infor_organ5}
    \end{subfigure}

    \begin{subfigure}[b]{\textwidth}
        \centering
        \includegraphics[width=\linewidth]{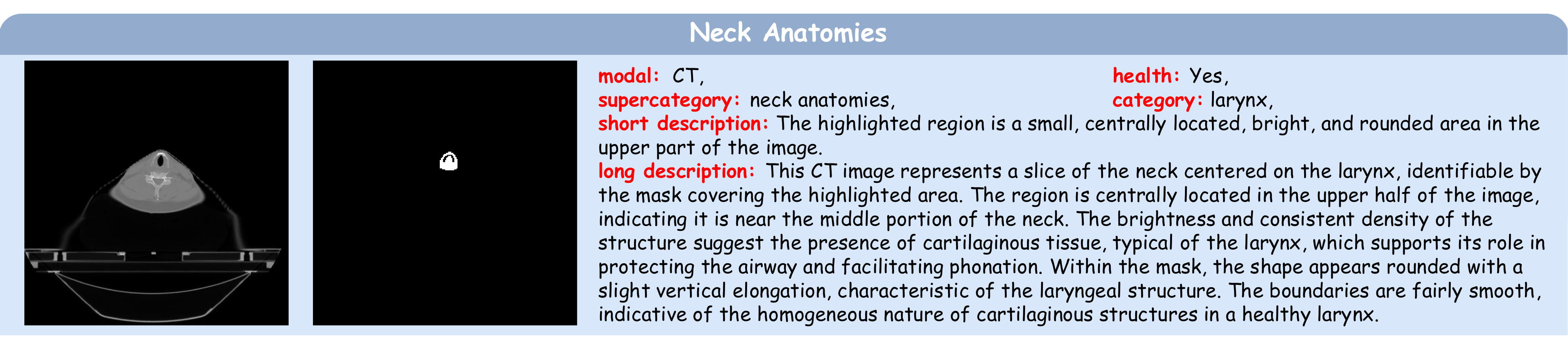}
        \label{fig:meta_infor_organ6}
    \end{subfigure}
    
    \caption{Examples of the meta information from three anatomical super-categories. From top to bottom: \textbf{Lung}, \textbf{Heart Anatomies}, \textbf{Neck Anatomies}. Each case presents the image along with its: \textit{modal}, \textit{health}, \textit{super-category}, \textit{category}, \textit{short description} and \textit{long description}.}

    \label{fig:meta_infor_4_to_6}
\end{figure*}

\begin{figure*}[t]
    \centering

    \begin{subfigure}[b]{\textwidth}
        \centering
        \includegraphics[width=\linewidth]{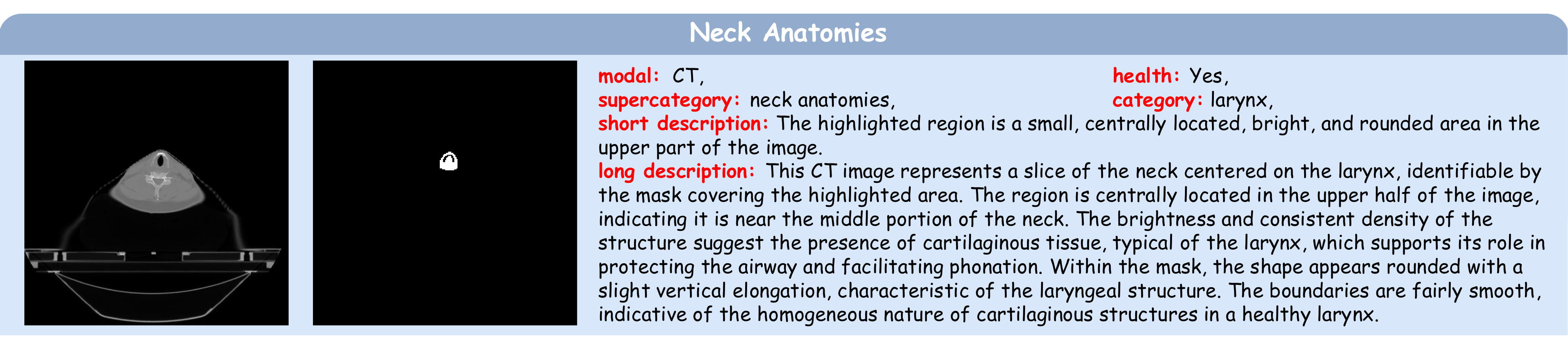}
        \label{fig:meta_infor_organ7}
    \end{subfigure}
    
    \begin{subfigure}[b]{\textwidth}
        \centering
        \includegraphics[width=\linewidth]{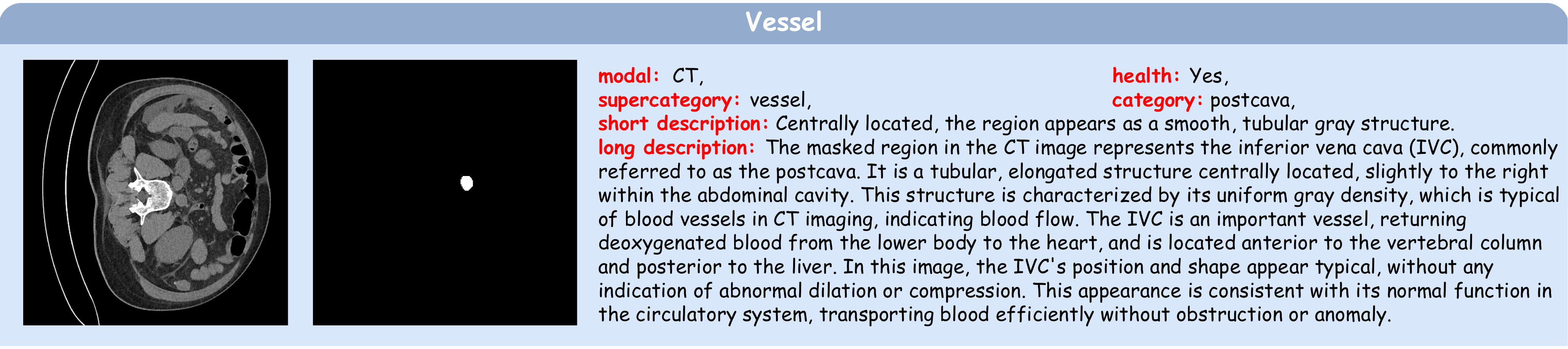}
        \label{fig:meta_infor_organ8}
    \end{subfigure}

    \begin{subfigure}[b]{\textwidth}
        \centering
        \includegraphics[width=\linewidth]{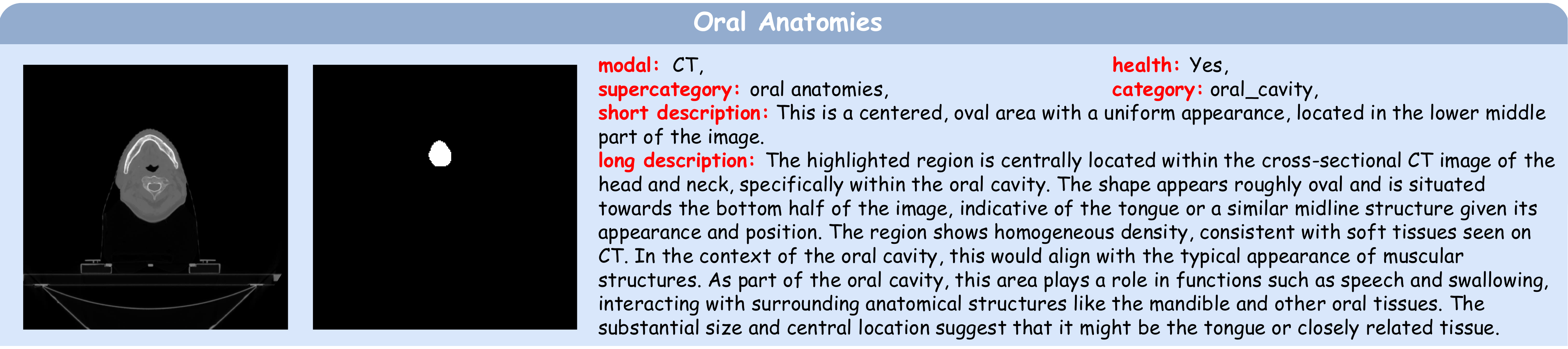}
        \label{fig:meta_infor_organ9}
    \end{subfigure}

    \caption{Examples of the meta information from three anatomical super-categories. From top to bottom: \textbf{Skeletal Anatomies}, \textbf{Vessel}, \textbf{Oral Anatomies}. Each case presents the image along with its: \textit{modal}, \textit{health}, \textit{super-category}, \textit{category}, \textit{short description} and \textit{long description}.}

    \label{fig:meta_infor_7_to_9}
\end{figure*}

\begin{figure*}[t]
    \centering

    \begin{subfigure}[b]{\textwidth}
        \centering
        \includegraphics[width=\linewidth]{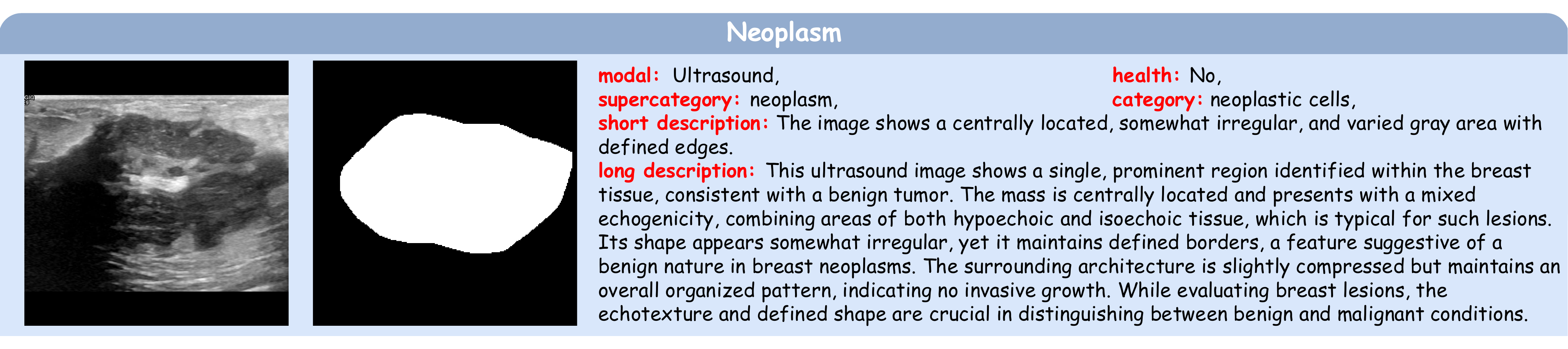}
        \label{fig:meta_infor_abnorm1}
    \end{subfigure}

    \begin{subfigure}[b]{\textwidth}
        \centering
        \includegraphics[width=\linewidth]{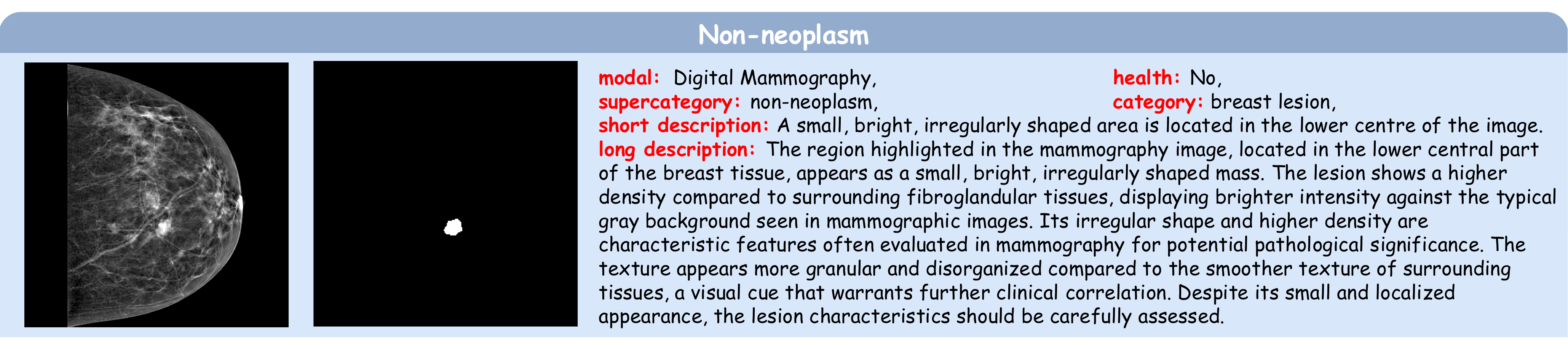}
        \label{fig:meta_infor_abnorm2}
    \end{subfigure}
    
    \begin{subfigure}[b]{\textwidth}
        \centering
        \includegraphics[width=\linewidth]{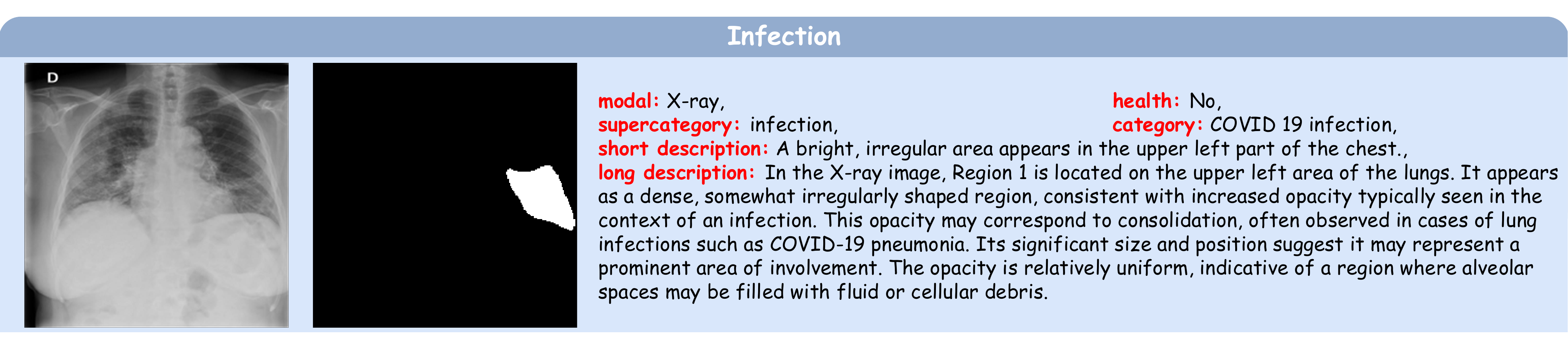}
        \label{fig:meta_infor_abnorm3}
    \end{subfigure}
    
    \begin{subfigure}[b]{\textwidth}
        \centering
        \includegraphics[width=\linewidth]{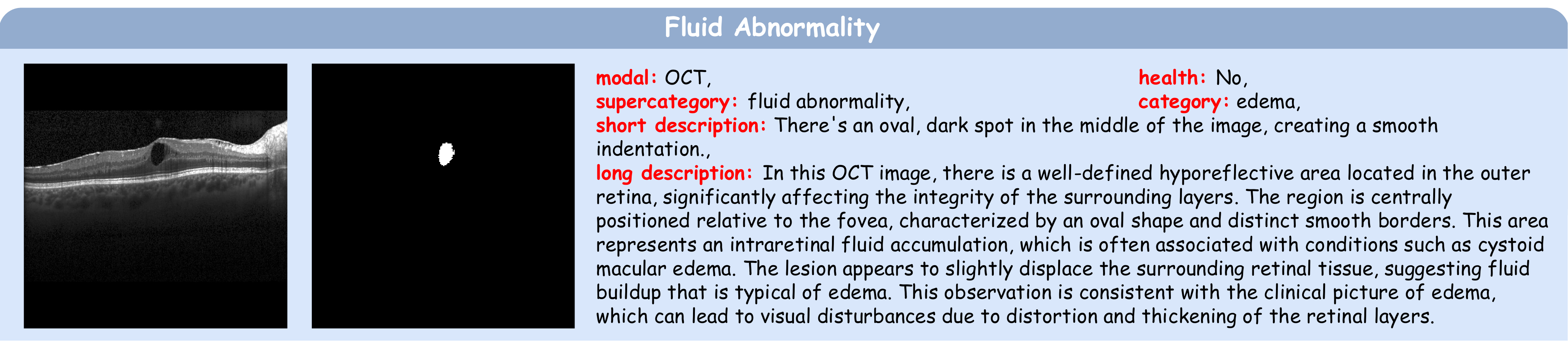}
        \label{fig:meta_infor_abnorm4}
    \end{subfigure}

    \caption{Examples of the meta information from four lesions super-categories. From top to bottom: \textbf{Neoplasm},  \textbf{Non-Neoplasm}, \textbf{Infection}, \textbf{Fluid Abnormality}. Each case presents the image along with its: \textit{modal}, \textit{health}, \textit{super-category}, \textit{category}, \textit{short description} and \textit{long description}.}

    \label{fig:meta_infor_10_to_13}
\end{figure*}

\begin{figure*}[t]
    \centering

    \vspace{50pt}

    \begin{subfigure}[b]{\textwidth}
        \centering
        \includegraphics[width=\linewidth]{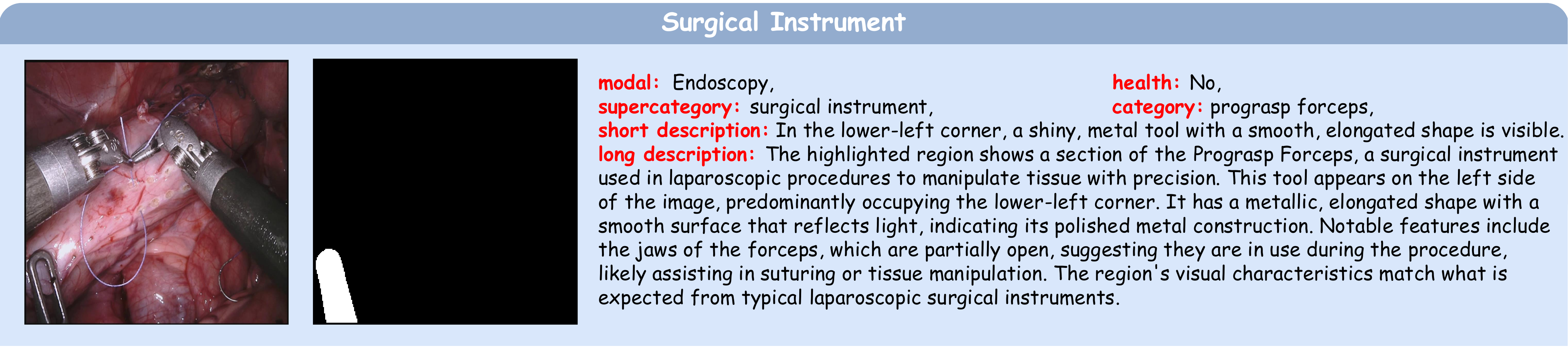}
        \label{fig:meta_infor_other1}
    \end{subfigure}

    \begin{subfigure}[b]{\textwidth}
        \centering
        \includegraphics[width=\linewidth]{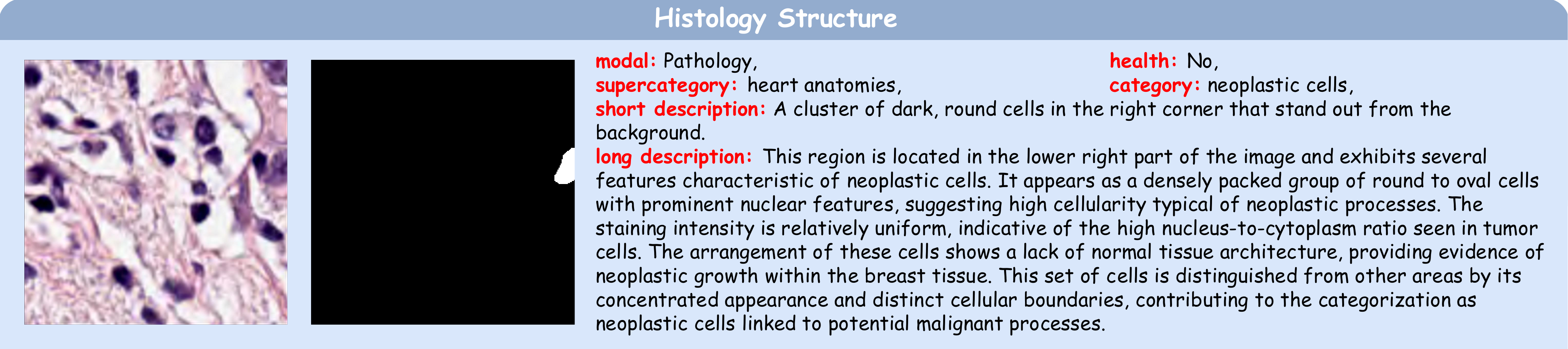}
        \label{fig:meta_infor_other2}
    \end{subfigure}

    \caption{Examples of the meta information from two other super-categories. From top to bottom: \textbf{Surgical Instrument},  \textbf{Histology Structure}. Each case presents the image along with its: \textit{modal}, \textit{health}, \textit{super-category}, \textit{category}, \textit{short description} and \textit{long description}.}

    \vspace{50pt}

    \label{fig:meta_infor_14_to_15}
\end{figure*}

We illustrate the \textbf{meta information} design using 15 representative cases, with one example selected from each super-category and arranged across Fig.~\ref{fig:meta_infor_1_to_3} to Fig.~\ref{fig:qa_pairs_14_to_15}.
For each image in the dataset, we construct a set of information that includes several key attributes, such as \textbf{imaging modality} and \textbf{subject health status}.
In addition, we provide a pair of carefully designed textual descriptions for each case: a \textbf{short description} that captures key visual cues in plain and intuitive language, and a \textbf{long description} that incorporates domain-specific knowledge to emphasize the region’s clinical relevance and distinctiveness.

\subsection{QA Formats of U-MRG-14K} \label{case:qa_formats}
\begin{figure*}[t]
    \centering

    \begin{subfigure}[b]{\textwidth}
        \centering
        \includegraphics[width=\linewidth]{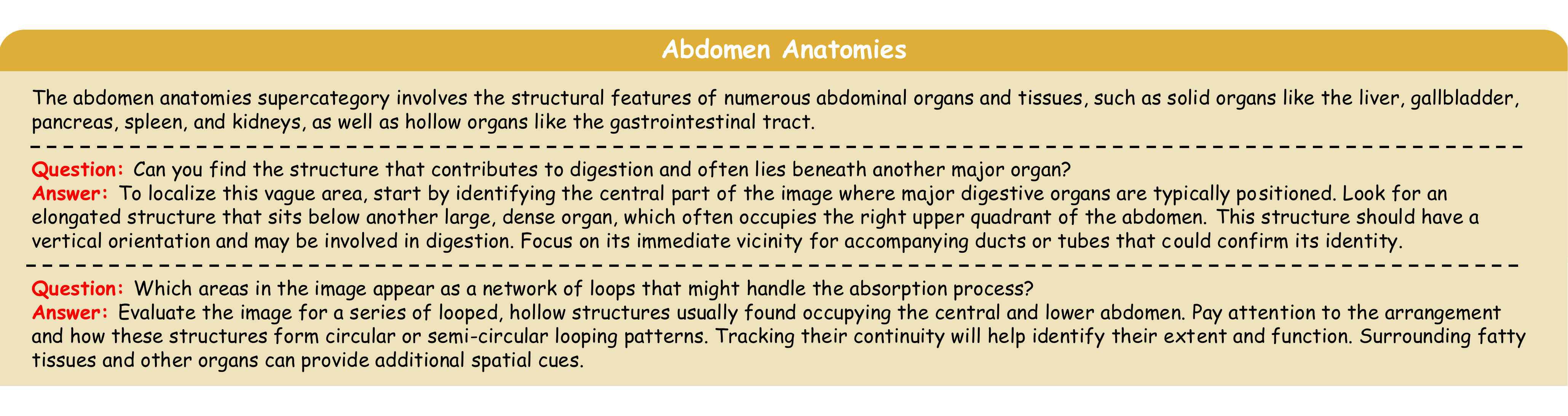}
        \label{fig:qa_formats_organ1}
    \end{subfigure}
    
    \begin{subfigure}[b]{\textwidth}
        \centering
        \includegraphics[width=\linewidth]{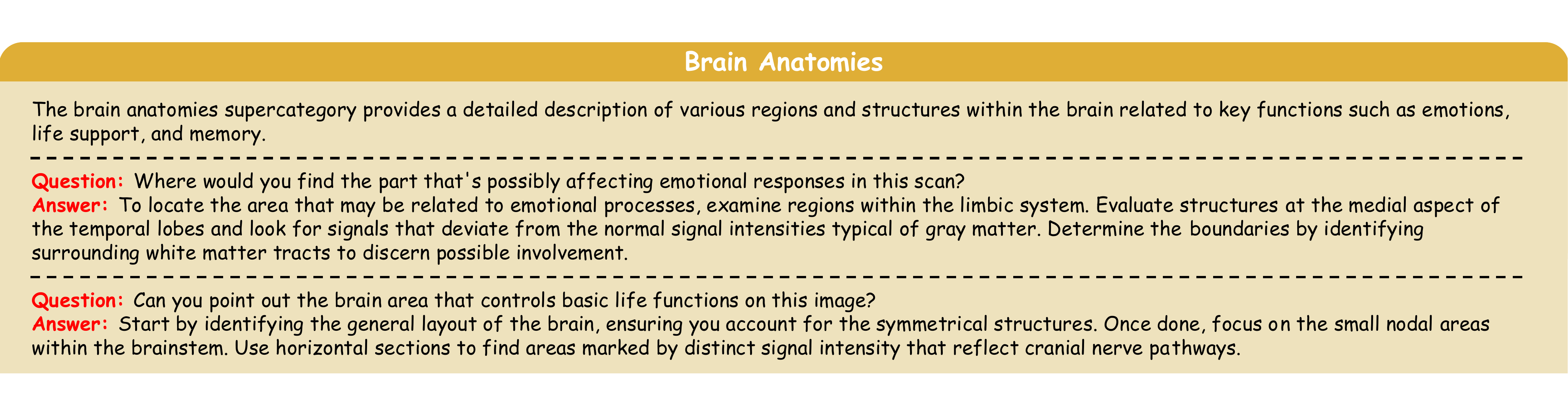}
        \label{fig:qa_formats_organ2}
    \end{subfigure}
    
    \begin{subfigure}[b]{\textwidth}
        \centering
        \includegraphics[width=\linewidth]{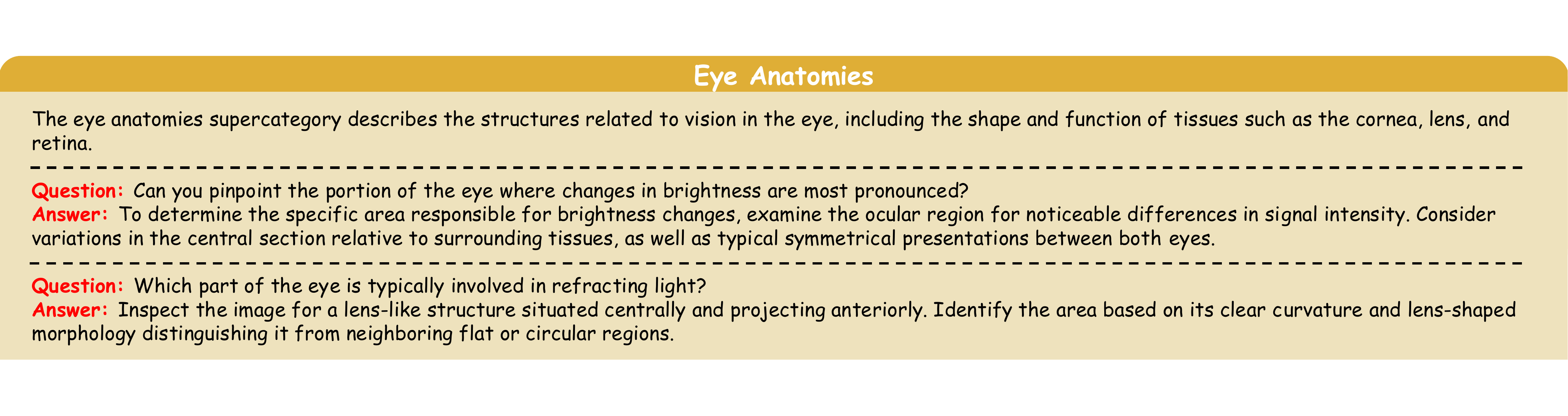}
        \label{fig:qa_formats_organ3}
    \end{subfigure}
    
    \caption{Examples of QA formats from three anatomical super-categories. From top to bottom: \textbf{Abdomen Anatomies}, \textbf{Brain Anatomies}, \textbf{Eye Anatomies}. Each case illustrates the intended meaning of the super-category and presents two distinct QA formats for it.}
    
    \label{fig:qa_formats_1_to_3}
\end{figure*}

\begin{figure*}[t]
    \centering

    \begin{subfigure}[b]{\textwidth}
        \centering
        \includegraphics[width=\linewidth]{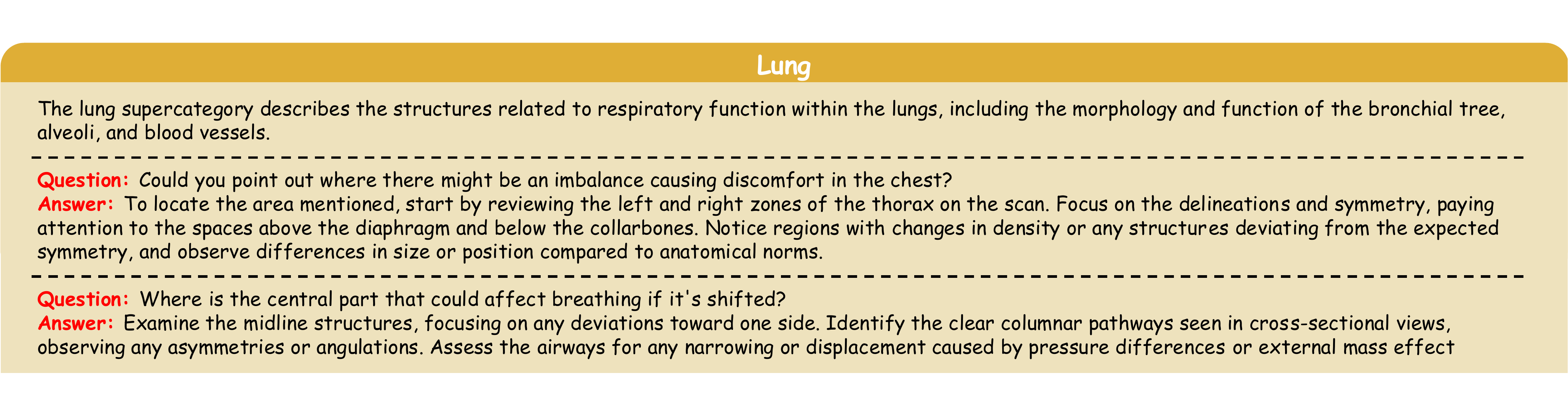}
        \label{fig:qa_formats_organ4}
    \end{subfigure}
    
    \begin{subfigure}[b]{\textwidth}
        \centering
        \includegraphics[width=\linewidth]{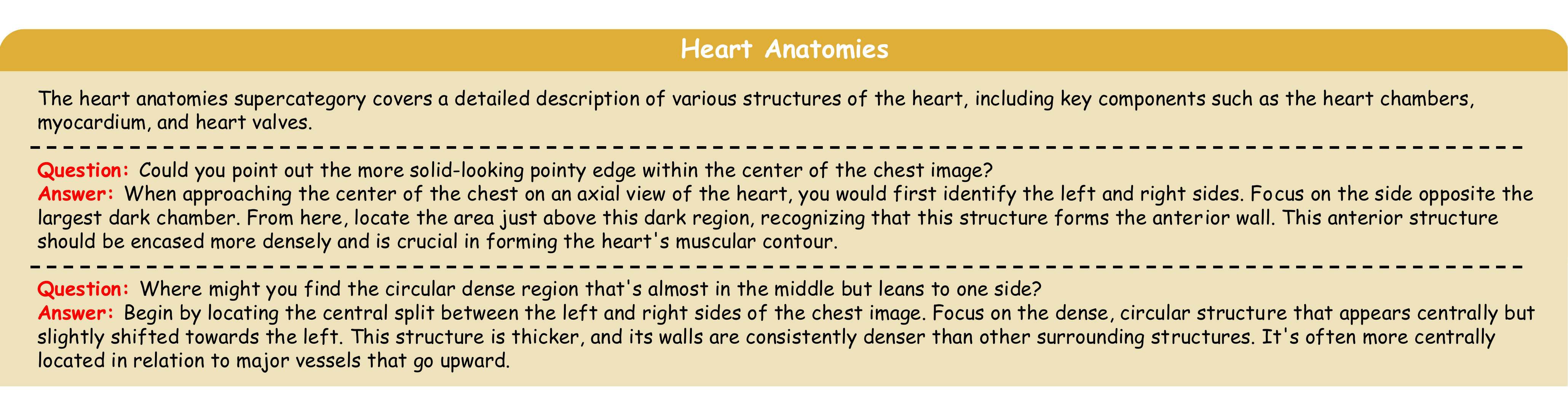}
        \label{fig:qa_formats_organ5}
    \end{subfigure}
    
    \begin{subfigure}[b]{\textwidth}
        \centering
        \includegraphics[width=\linewidth]{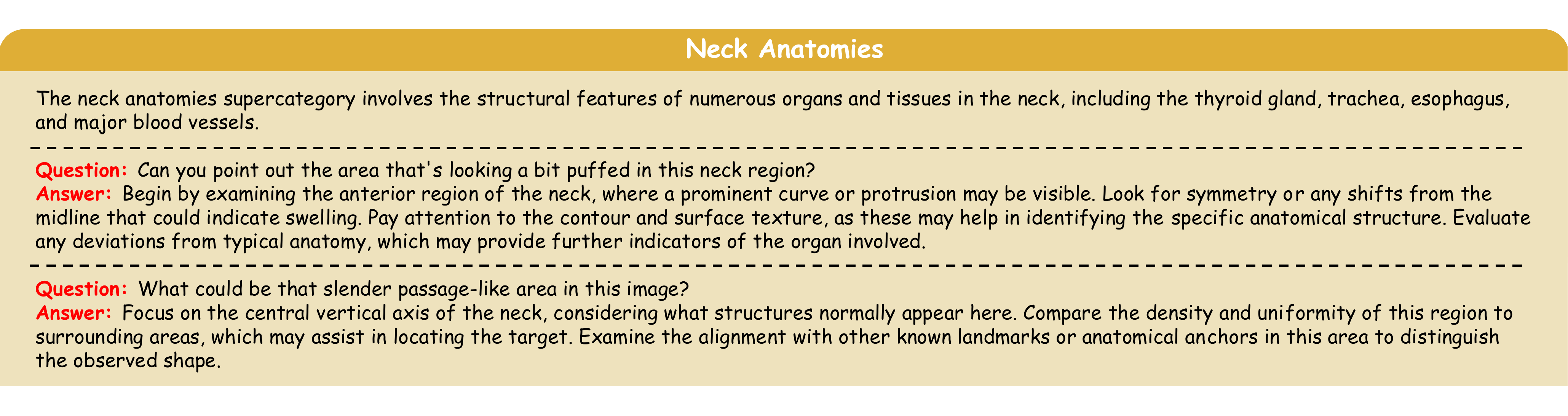}
        \label{fig:qa_formats_organ6}
    \end{subfigure}
    
    \caption{Examples of the meta information from three anatomical super-categories. From top to bottom: \textbf{Lung}, \textbf{Heart Anatomies}, \textbf{Neck Anatomies}. Each case illustrates the intended meaning of the super-category and presents two distinct QA formats for it.}

    \label{fig:qa_formats_4_to_6}
\end{figure*}

\begin{figure*}[t]
    \centering

    \begin{subfigure}[b]{\textwidth}
        \centering
        \includegraphics[width=\linewidth]{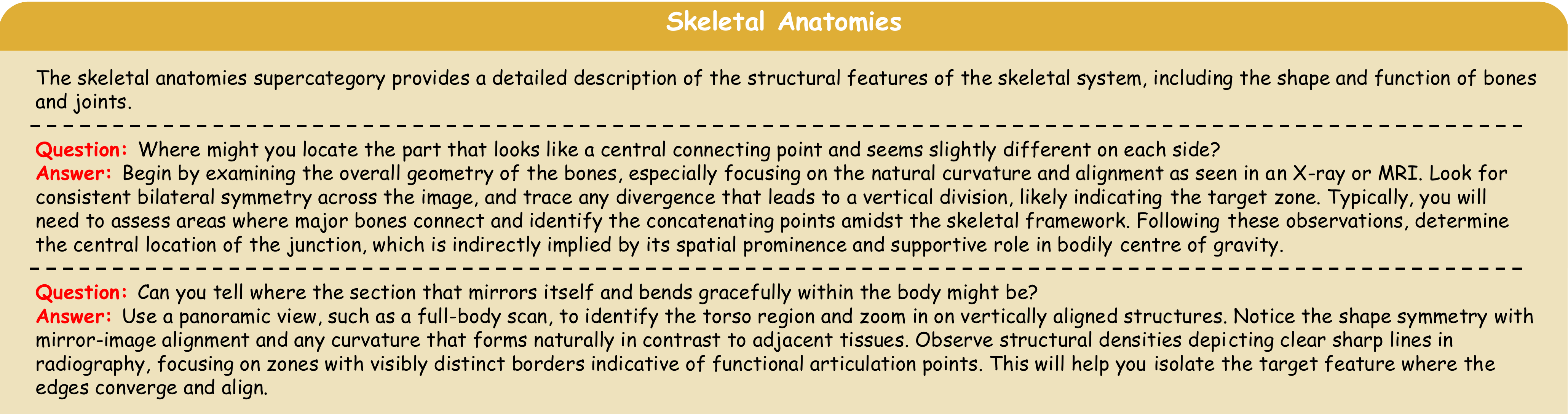}
        \label{fig:qa_formats_organ7}
    \end{subfigure}
    
    \begin{subfigure}[b]{\textwidth}
        \centering
        \includegraphics[width=\linewidth]{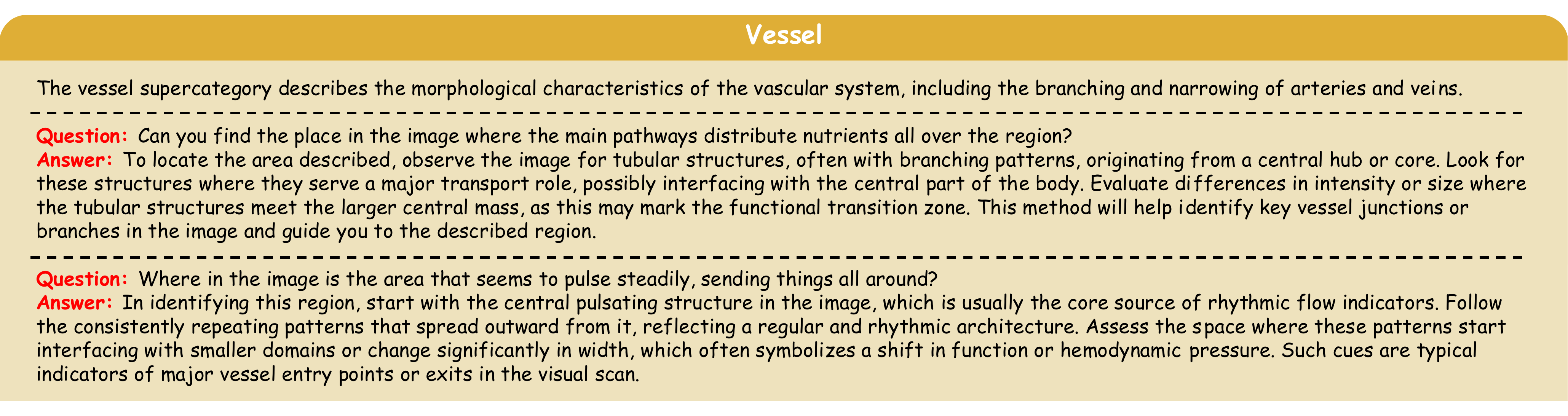}
        \label{fig:qa_formats_organ8}
    \end{subfigure}
    
    \begin{subfigure}[b]{\textwidth}
        \centering
        \includegraphics[width=\linewidth]{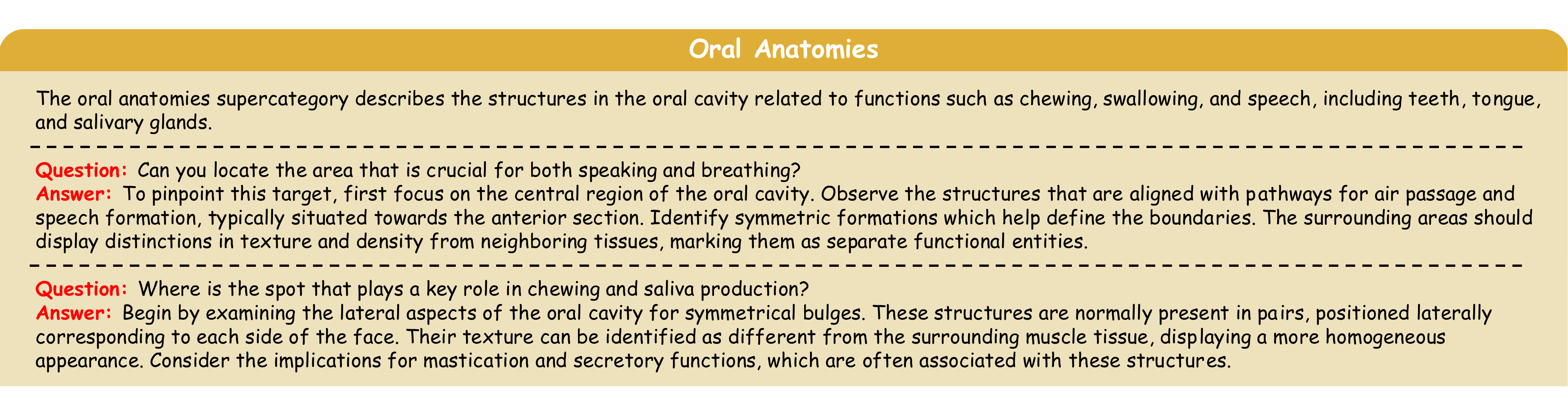}
        \label{fig:qa_formats_organ9}
    \end{subfigure}
    
    \caption{Examples of the meta information from three anatomical super-categories. From top to bottom: \textbf{Skeletal Anatomies}, \textbf{Vessel}, \textbf{Oral Anatomies}. Each case illustrates the intended meaning of the super-category and presents two distinct QA formats for it.}
    
    \label{fig:qa_formats_7_to_9}
\end{figure*}

\begin{figure*}[t]
    \centering

    \begin{subfigure}[b]{\textwidth}
        \centering
        \includegraphics[width=\linewidth]{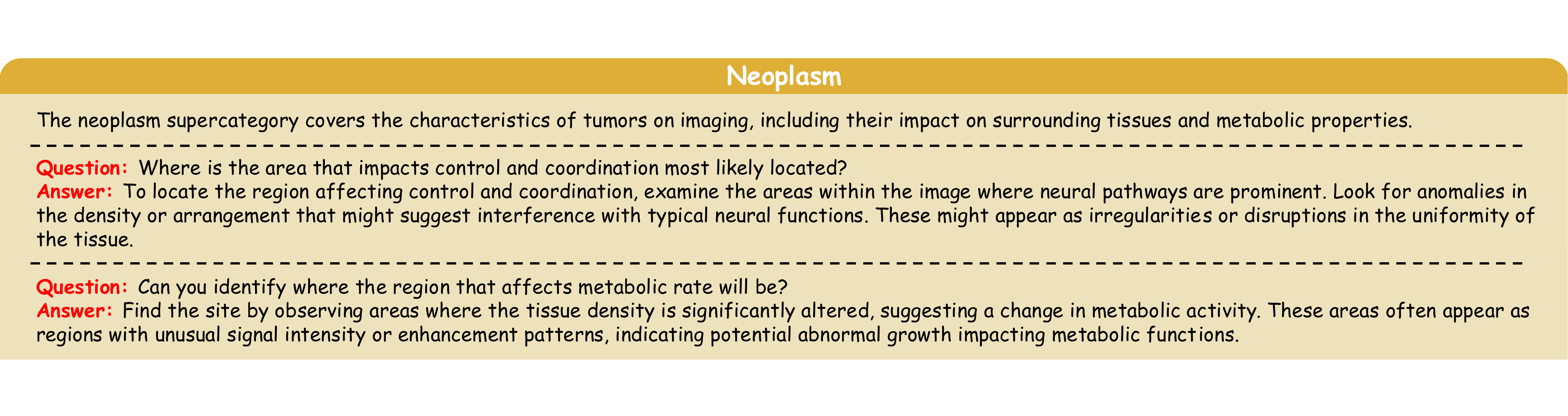}
        \label{fig:qa_formats_abnorm1}
    \end{subfigure}
    
    \begin{subfigure}[b]{\textwidth}
        \centering
        \includegraphics[width=\linewidth]{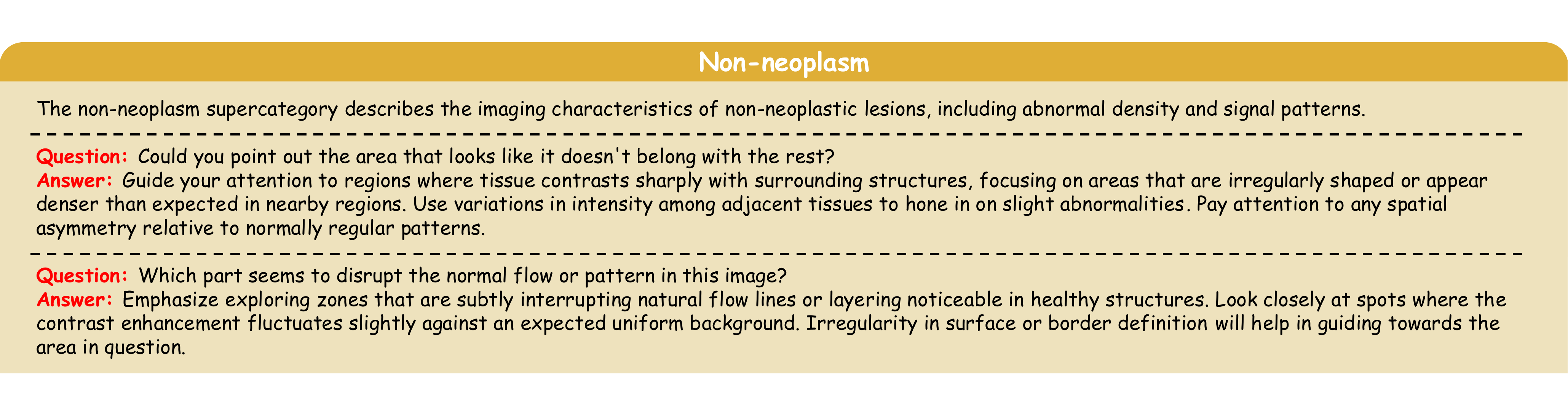}
        \label{fig:qa_formats_abnorm2}
    \end{subfigure}
    
    \begin{subfigure}[b]{\textwidth}
        \centering
        \includegraphics[width=\linewidth]{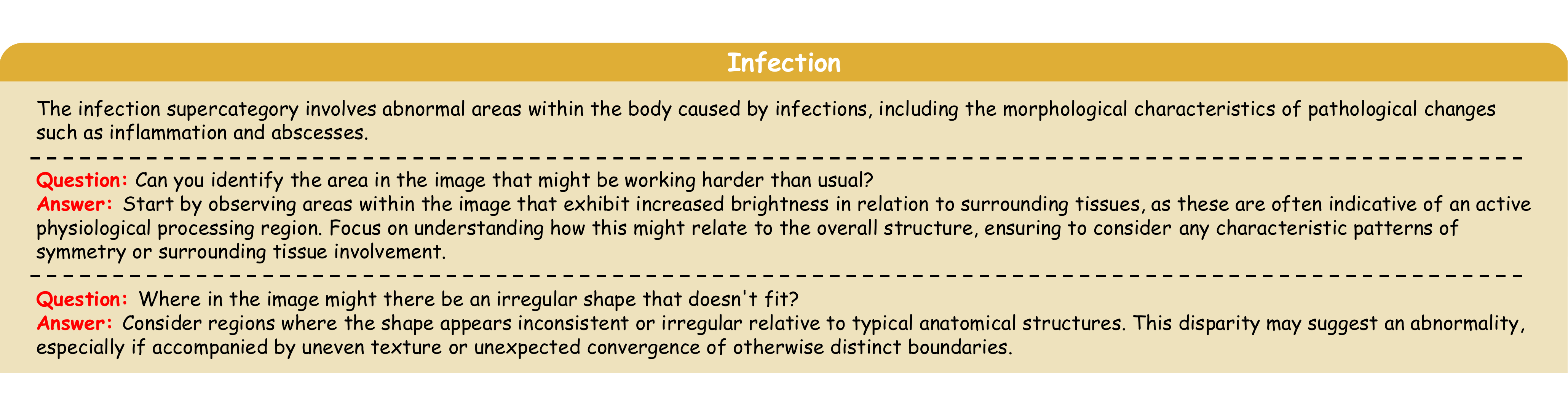}
        \label{fig:qa_formats_abnorm3}
    \end{subfigure}

    \begin{subfigure}[b]{\textwidth}
        \centering
        \includegraphics[width=\linewidth]{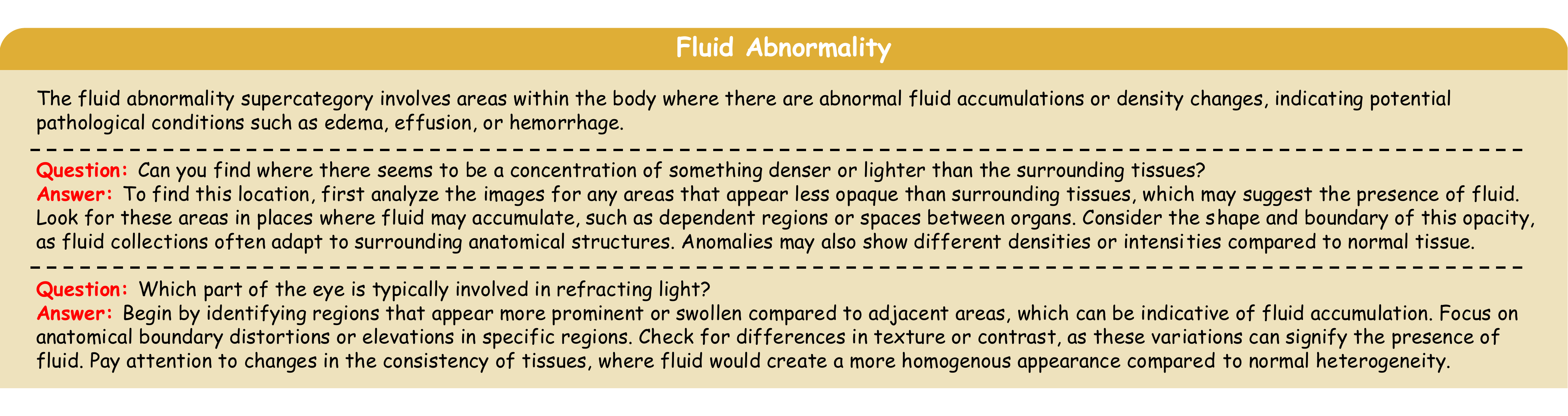}
        \label{fig:qa_formats_abnorm4}
    \end{subfigure}
    
    \caption{Examples of the meta information from four lesions super-categories. From top to bottom: \textbf{Neoplasm},  \textbf{Non-Neoplasm}, \textbf{Infection}, \textbf{Fluid Abnormality}. Each case illustrates the intended meaning of the super-category and presents two distinct QA formats for it.}

    \label{fig:qa_formats_10_to_13}
\end{figure*}

\begin{figure*}[t]
    \centering

    \vspace{50pt}

    \begin{subfigure}[b]{\textwidth}
        \centering
        \includegraphics[width=\linewidth]{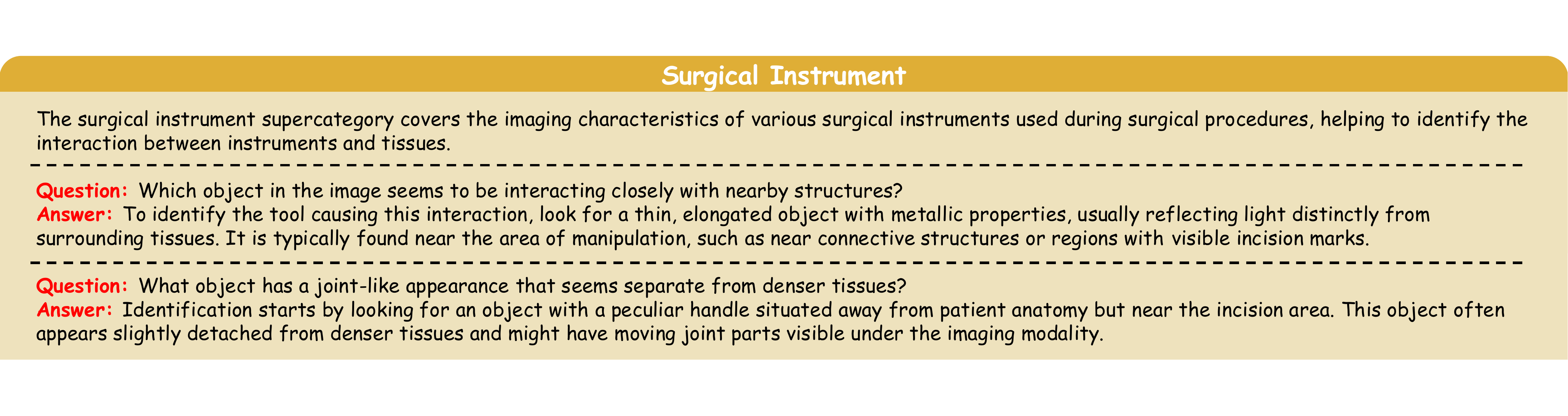}
        \label{fig:qa_formats_other1}
    \end{subfigure}
    
    \begin{subfigure}[b]{\textwidth}
        \centering
        \includegraphics[width=\linewidth]{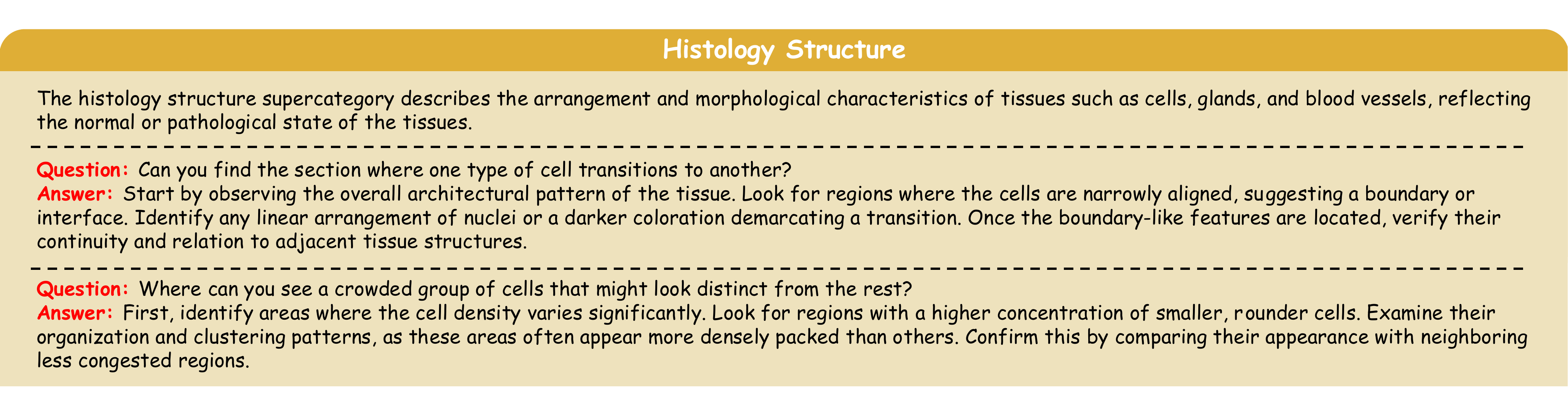}
        \label{fig:qa_formats_other2}
    \end{subfigure}
    
    \caption{Examples of the meta information from two other super-categories. From top to bottom: \textbf{Surgical Instrument},  \textbf{Histology Structure}. Each case illustrates the intended meaning of the super-category and presents two distinct QA formats for it.}

    \vspace{50pt}

    \label{fig:qa_formats_14_to_15}
\end{figure*}

We illustrate the \textbf{QA formats} design using 15 representative cases, with two examples selected from each super-category and arranged across Fig.~\ref{fig:qa_formats_1_to_3} to Fig.~\ref{fig:qa_formats_14_to_15}.
Each case includes a concise explanation of the corresponding super-category and showcases two representative QA formats that capture diverse \textit{query intents} and \textit{reasoning strategies}. The formats cover common clinical scenarios such as \textit{location reference}, \textit{attribute reasoning}, and \textit{structural inference}, reflecting how our prompt design accommodates linguistic ambiguity while aligning with medical grounding objectives.

\subsection{QA Pairs of U-MRG-14K} \label{case:qa_pairss}
\begin{figure*}[t]
    \centering

    \begin{subfigure}[b]{\textwidth}
        \centering
        \includegraphics[width=\linewidth]{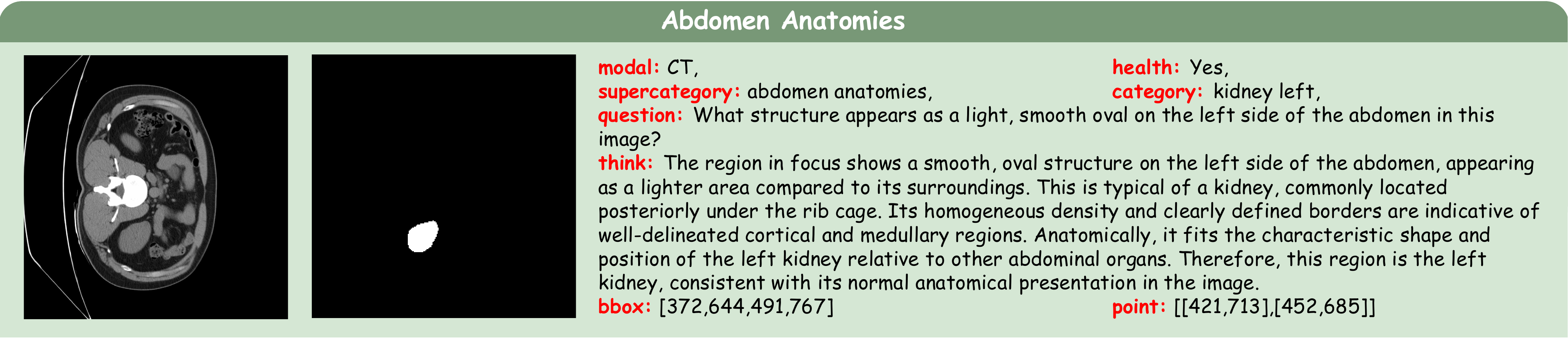}
        \label{fig:qa_pairs_organ1}
    \end{subfigure}
    
    \begin{subfigure}[b]{\textwidth}
        \centering
        \includegraphics[width=\linewidth]{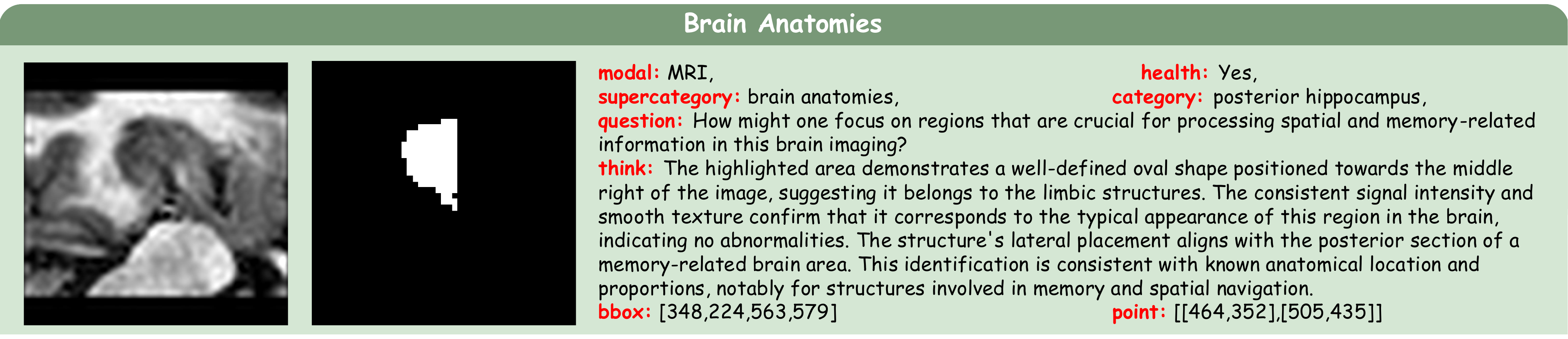}
        \label{fig:qa_pairs_organ2}
    \end{subfigure}
    
    \begin{subfigure}[b]{\textwidth}
        \centering
        \includegraphics[width=\linewidth]{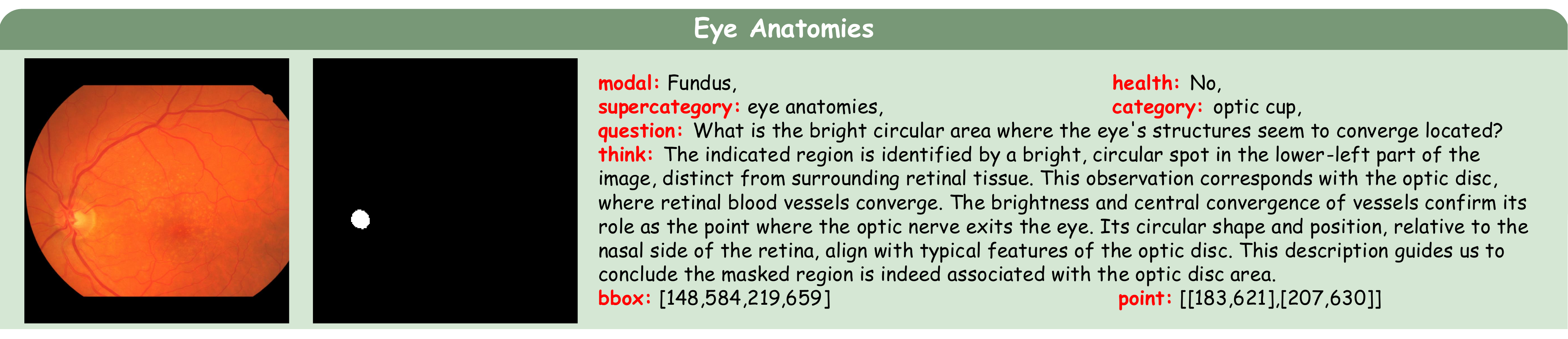}
        \label{fig:qa_pairs_organ3}
    \end{subfigure}
    
    \caption{Examples of the QA pairs from three anatomical super-categories. From top to bottom: \textbf{Abdomen Anatomies}, \textbf{Brain Anatomies}, \textbf{Eye Anatomies}. Each case presents the image along with its: \textit{modal}, \textit{health}, \textit{super-category}, \textit{category}, \textit{question}, \textit{think}, and \textit{answer (bbox, point)}.}

    \label{fig:qa_pairs_1_to_3}
\end{figure*}

\begin{figure*}[t]
    \centering

    \begin{subfigure}[b]{\textwidth}
        \centering
        \includegraphics[width=\linewidth]{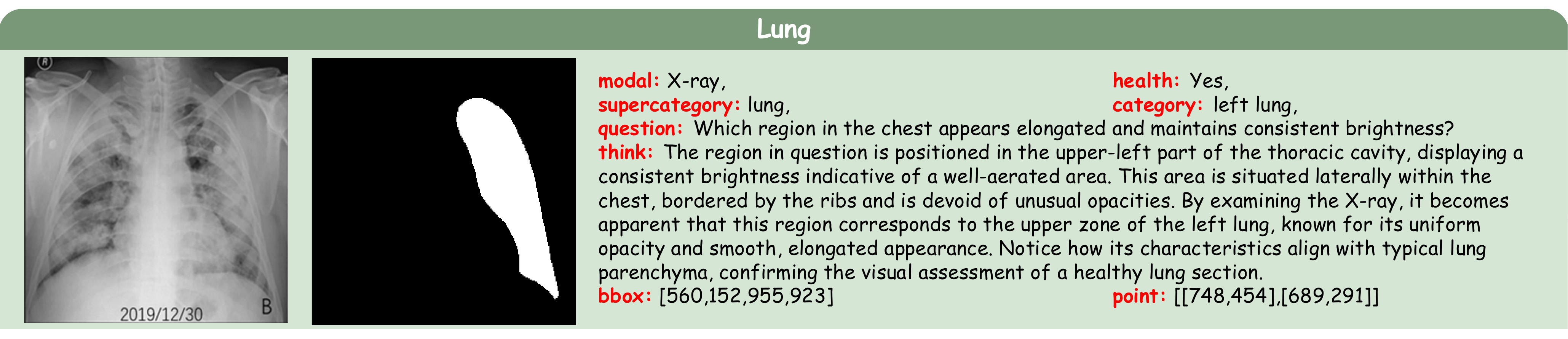}
        \label{fig:qa_pairs_organ4}
    \end{subfigure}

    \begin{subfigure}[b]{\textwidth}
        \centering
        \includegraphics[width=\linewidth]{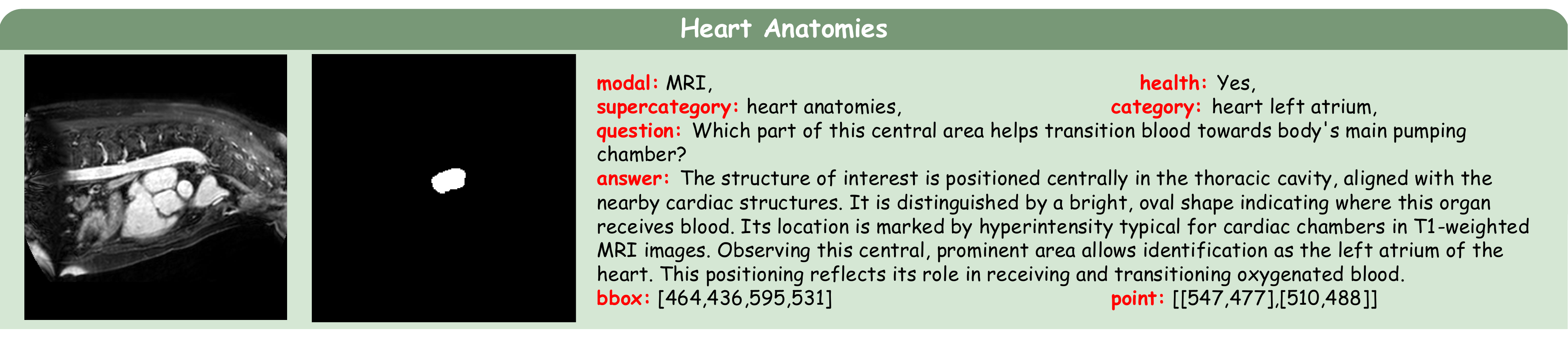}
        \label{fig:qa_pairs_organ5}
    \end{subfigure}

    \begin{subfigure}[b]{\textwidth}
        \centering
        \includegraphics[width=\linewidth]{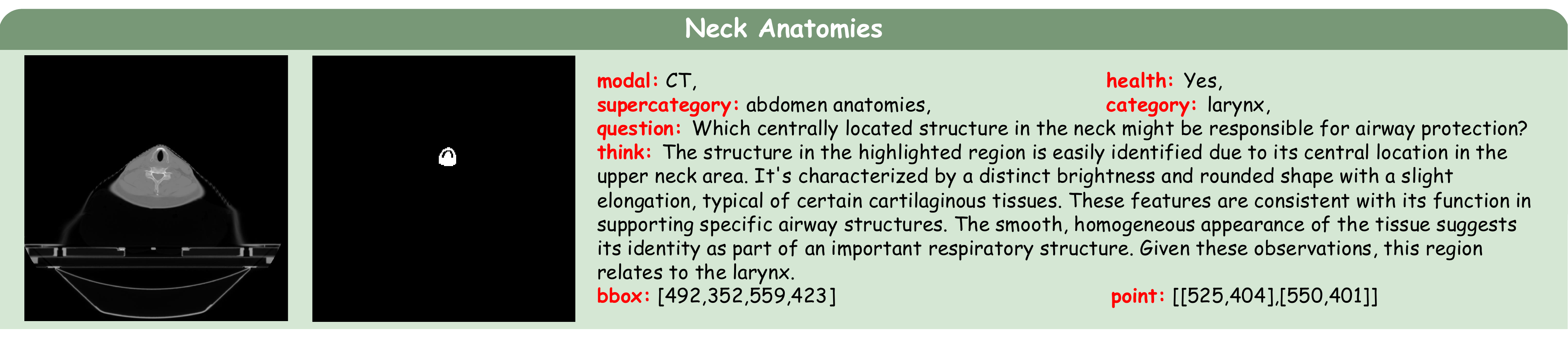}
        \label{fig:qa_pairs_organ6}
    \end{subfigure}
    
    \caption{Examples of the QA pairs from three anatomical super-categories. From top to bottom: \textbf{Lung}, \textbf{Heart Anatomies}, \textbf{Neck Anatomies}. Each case presents the image along with its: \textit{modal}, \textit{health}, \textit{super-category}, \textit{category}, \textit{question}, \textit{think}, and \textit{answer (bbox, point)}.}

    \label{fig:qa_pairs_4_to_6}
\end{figure*}

\begin{figure*}[t]
    \centering

    \begin{subfigure}[b]{\textwidth}
        \centering
        \includegraphics[width=\linewidth]{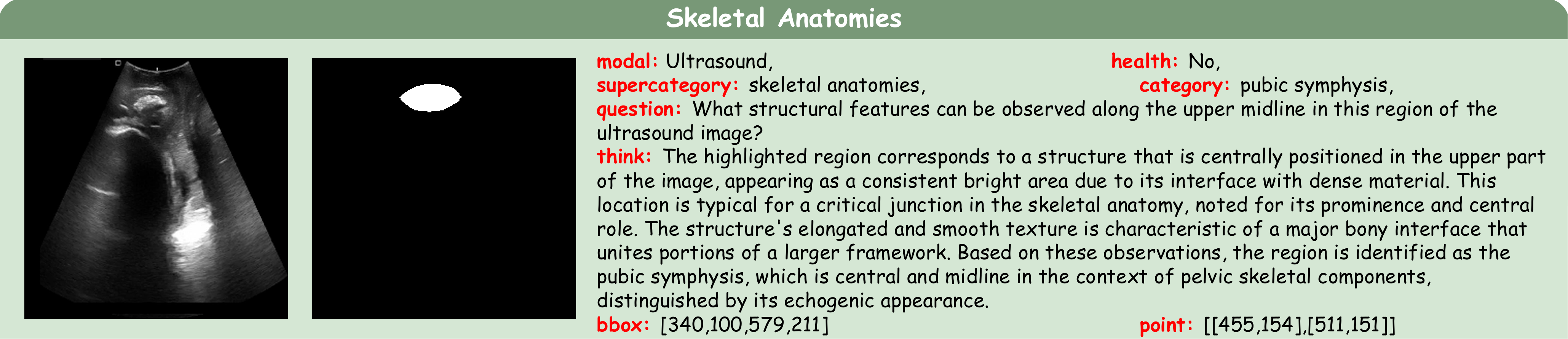}
        \label{fig:qa_pairs_organ7}
    \end{subfigure}
    
    \begin{subfigure}[b]{\textwidth}
        \centering
        \includegraphics[width=\linewidth]{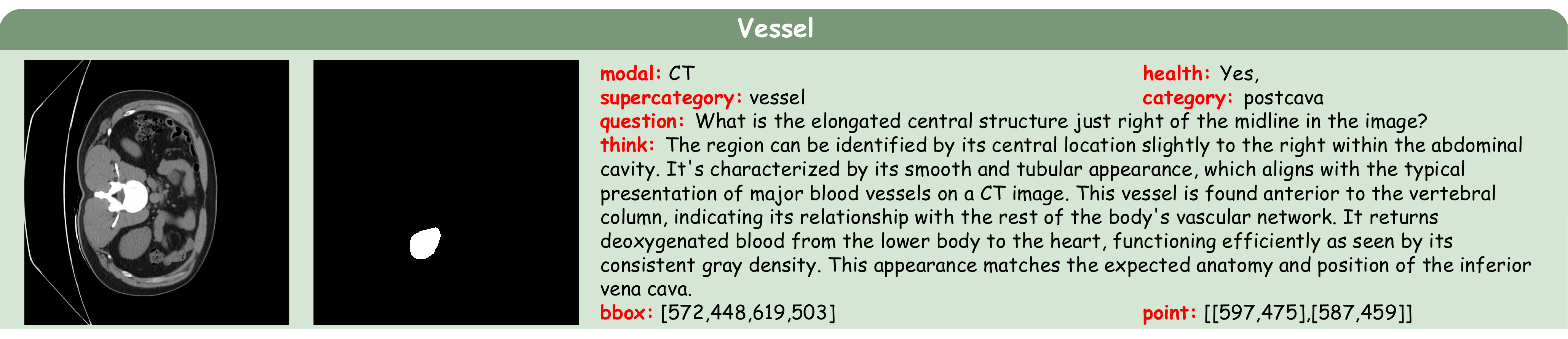}
        \label{fig:qa_pairs_organ8}
    \end{subfigure}

    \begin{subfigure}[b]{\textwidth}
        \centering
        \includegraphics[width=\linewidth]{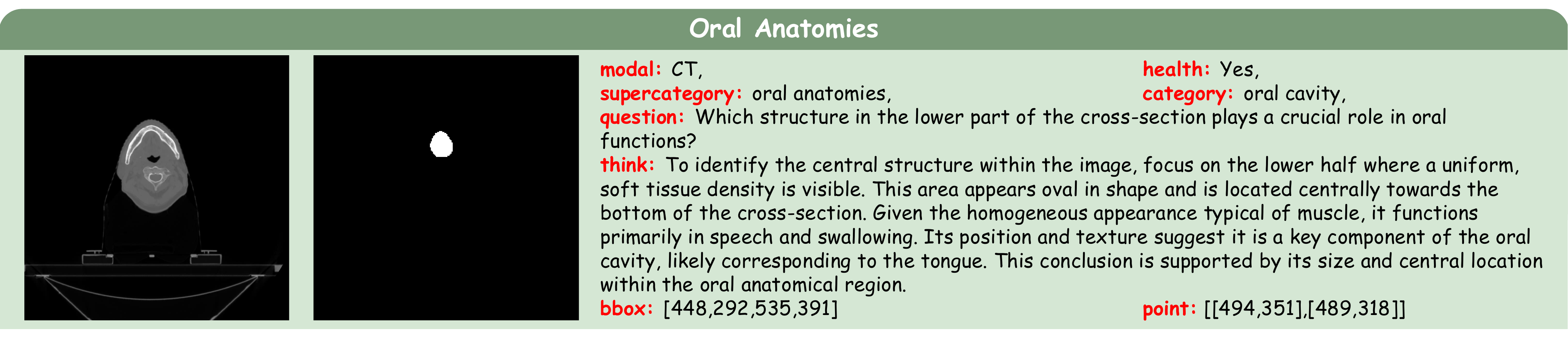}
        \label{fig:qa_pairs_organ9}
    \end{subfigure}

    \caption{Examples of the QA pairs from three anatomical super-categories. From top to bottom: \textbf{Skeletal Anatomies}, \textbf{Vessel}, \textbf{Oral Anatomies}. Each case presents the image along with its: \textit{modal}, \textit{health}, \textit{super-category}, \textit{category}, \textit{question}, \textit{think}, and \textit{answer (bbox, point)}.}

    \label{fig:qa_pairs_7_to_9}
\end{figure*}

\begin{figure*}[t]
    \centering

    \begin{subfigure}[b]{\textwidth}
        \centering
        \includegraphics[width=\linewidth]{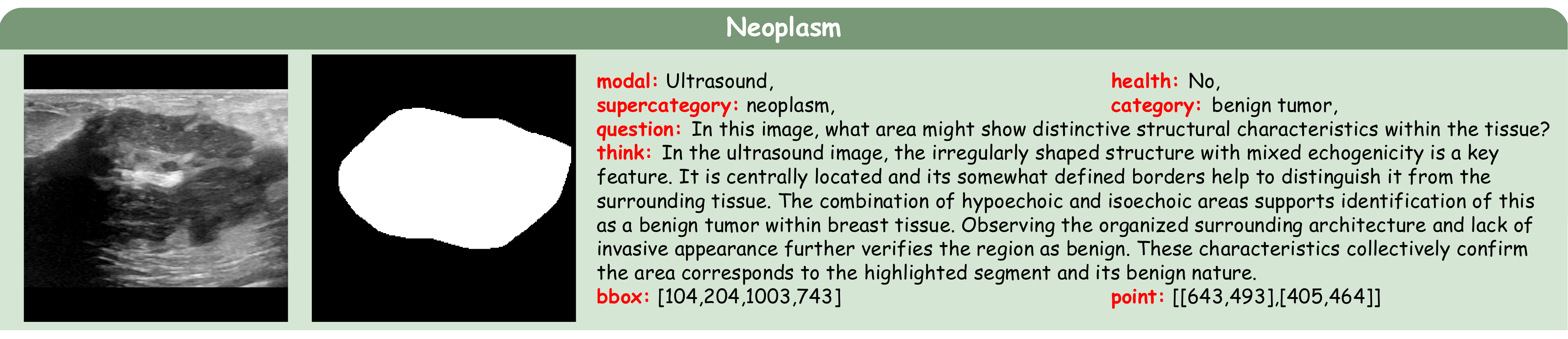}
        \label{fig:qa_pairs_abnorm1}
    \end{subfigure}

    \begin{subfigure}[b]{\textwidth}
        \centering
        \includegraphics[width=\linewidth]{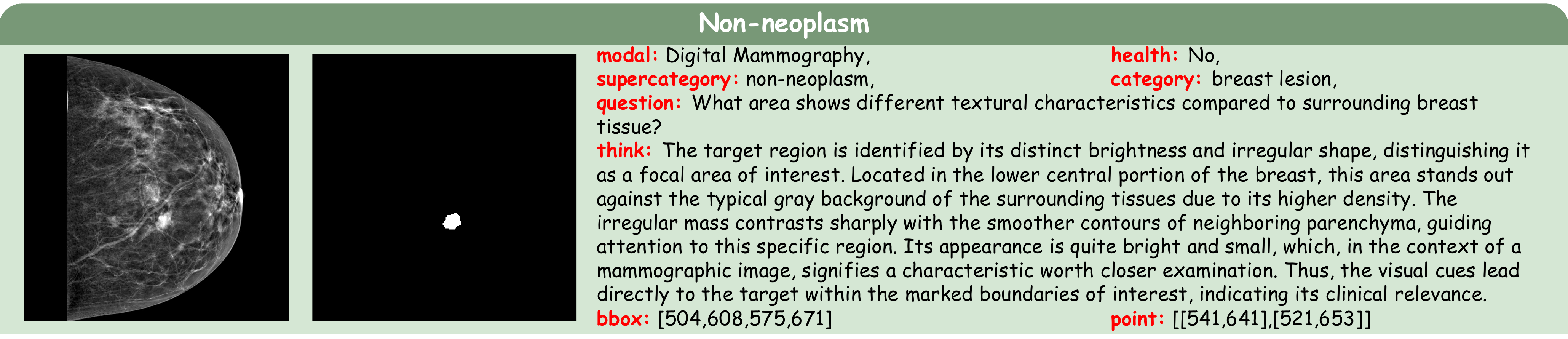}
        \label{fig:qa_pairs_abnorm2}
    \end{subfigure}
    
    \begin{subfigure}[b]{\textwidth}
        \centering
        \includegraphics[width=\linewidth]{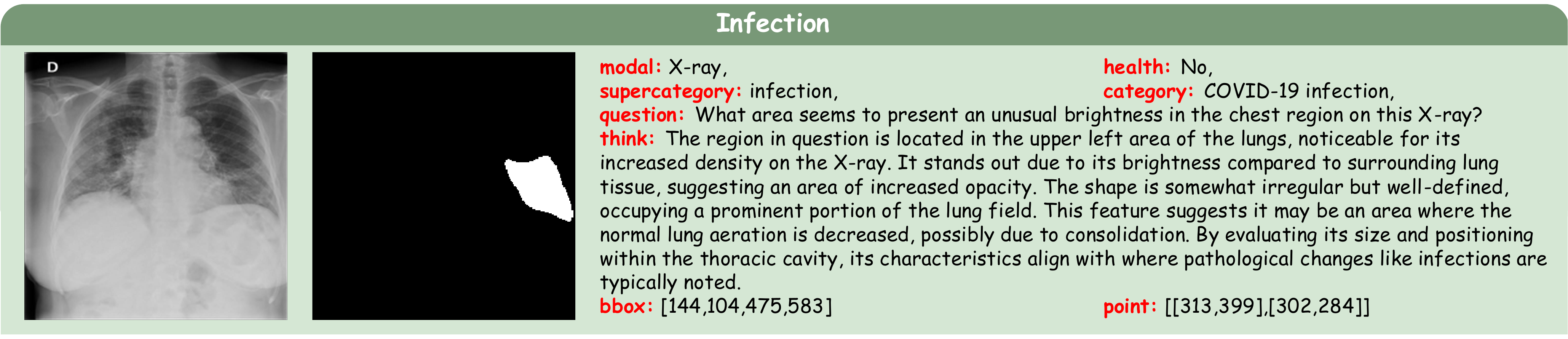}
        \label{fig:qa_pairs_abnorm3}
    \end{subfigure}
    
    \begin{subfigure}[b]{\textwidth}
        \centering
        \includegraphics[width=\linewidth]{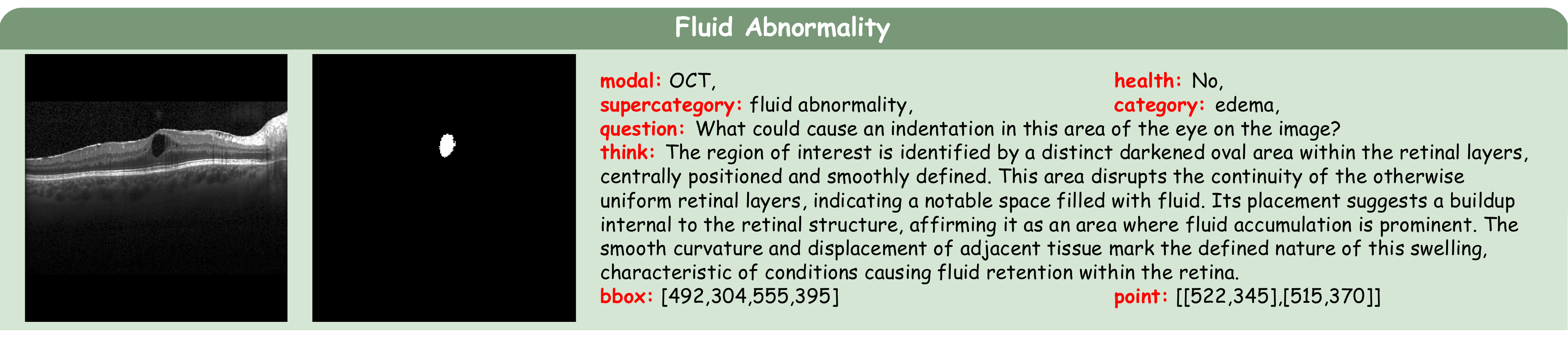}
        \label{fig:qa_pairs_abnorm4}
    \end{subfigure}

    \caption{Examples of the QA pairs from four lesions super-categories. From top to bottom: \textbf{Neoplasm},  \textbf{Non-Neoplasm}, \textbf{Infection}, \textbf{Fluid Abnormality}. Each case presents the image along with its: \textit{modal}, \textit{health}, \textit{super-category}, \textit{category}, \textit{question}, \textit{think}, and \textit{answer (bbox, point)}.}

    \label{fig:qa_pairs_10_to_13}
\end{figure*}

\begin{figure*}[t]
    \centering

    \vspace{50pt}
    
    \begin{subfigure}[b]{\textwidth}
        \centering
        \includegraphics[width=\linewidth]{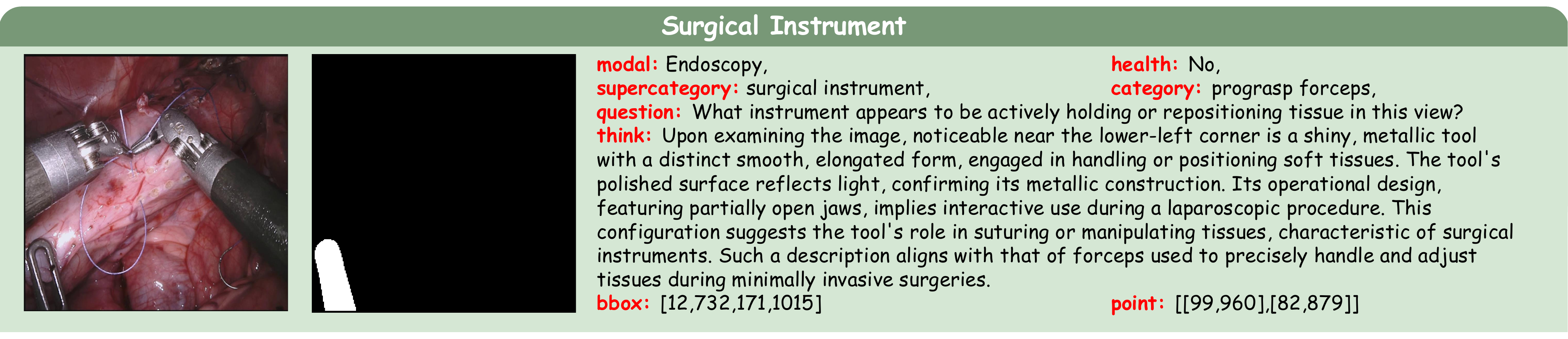}
        \label{fig:qa_pairs_other1}
    \end{subfigure}

    \begin{subfigure}[b]{\textwidth}
        \centering
        \includegraphics[width=\linewidth]{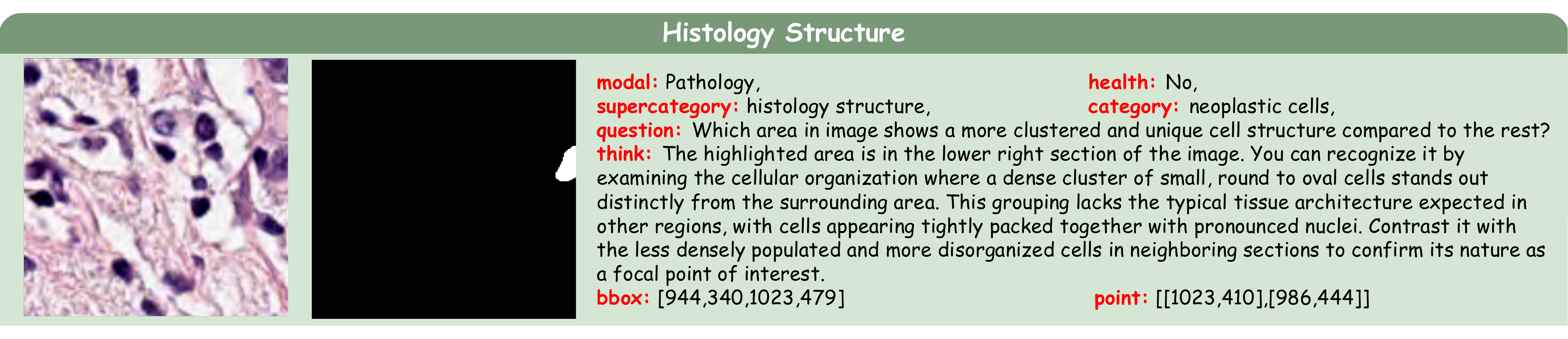}
        \label{fig:qa_pairs_other2}
    \end{subfigure}

    \vspace{50pt}

    \caption{Examples of the QA pairs from two other super-categories. From top to bottom: \textbf{Surgical Instrument},  \textbf{Histology Structure}. Each case presents the image along with its: \textit{modal}, \textit{health}, \textit{super-category}, \textit{category}, \textit{question}, \textit{think}, and \textit{answer (bbox, point)}.}

    \label{fig:qa_pairs_14_to_15}
\end{figure*}

We illustrate the \textbf{QA pairs} design using 15 representative cases, with one example selected from each super-category and arranged across Fig.~\ref{fig:qa_pairs_1_to_3} to Fig.~\ref{fig:qa_pairs_14_to_15}.
For each image in the dataset, we construct a set of information that includes several key attributes, such as \textbf{imaging modality} and \textbf{subject health status}.
In addition, we provide a set of carefully constructed QA pairs for each case. The \textbf{question} is formulated with implicit clinical reasoning and manually filtered to align with the UMRG task requirements. The \textbf{think} field captures a step-by-step reasoning path generated by GPT-4o, simulating a clinician's thought process and enabling analysis of the model’s understanding. The \textbf{answer} contains accurate spatial grounding derived from the annotated mask, including a bounding box and two key points.



\end{document}